\documentclass{article}
\usepackage{arxiv}
\usepackage[utf8]{inputenc} 
\usepackage[T1]{fontenc}    
\usepackage{hyperref}       
\usepackage{url}            
\usepackage{booktabs}       
\usepackage{amsfonts}       
\usepackage{nicefrac}       
\usepackage{microtype}      
\usepackage{lipsum}		
\usepackage{graphicx}
\usepackage{natbib}
\usepackage{doi}
\usepackage{multirow}
\usepackage{pdflscape}
\usepackage{amsmath}
\usepackage[dvipsnames]{xcolor}

\title{Green Learning: Introduction, Examples and Outlook}

\author{
C.-C. Jay Kuo \\
University of Southern California \\
Los Angeles, California, USA \\
\And 
Azad M. Madni \\
University of Southern California \\
Los Angeles, California, USA \\
}

\begin{document}
\maketitle

\begin{abstract}

Rapid advances in artificial intelligence (AI) in the last decade have
largely been built upon the wide applications of deep learning (DL).
However, the high carbon footprint yielded by larger and larger DL
networks becomes a concern for sustainability. Furthermore, DL decision
mechanism is somewhat obscure and can only be verified by test data.
Green learning (GL) has been proposed as an alternative paradigm to
address these concerns. GL is characterized by low carbon footprints,
small model sizes, low computational complexity, and logical
transparency.  It offers energy-effective solutions in cloud centers as
well as mobile/edge devices. GL also provides a clear and logical
decision-making process to gain people's trust.  Several statistical
tools have been developed to achieve this goal in recent years. They
include unsupervised representation learning, supervised feature
learning, and supervised decision learning.  We have seen a few
successful GL examples with performance comparable with state-of-the-art
DL solutions. This paper offers an introduction to GL, its demonstrated
applications, and future outlook. 

\end{abstract}


\section{Introduction}\label{sec:introduction}

There have been rapid advances in artificial intelligence (AI) and
machine learning (ML) in the last decade. The breakthrough has largely
been built upon the construction of larger datasets and the design of
complex neural networks.  Representative neural networks include the
convolutional neural network (CNN) \citep{gu2018recent}, the recurrent
neural network (RNN), \citep{pascanu2013difficulty,
salehinejad2017recent, su2022recurrent} the long short-term memory
network (LSTM) \citep{greff2016lstm, hochreiter1997long, su2019extended},
etc. Deep neural networks (DNNs) have attracted a lot of attention from
academia and industry since its resurgence in 2012
\citep{NIPS2012_AlexNet}. As networks become larger and deeper, this
discipline is named deep learning (DL) \citep{Nature2015}.  DL has made
great impacts in various application domains, include computer vision,
natural language processing, autonomous driving, robotics navigation,
etc. 

DL is characterized by two design choices: the network architecture and
the loss function. Once these two choices are specified, model
parameters can be automatically determined via an end-to-end
optimization algorithm called backpropagation.  When the number of
training samples is less than the number of model parameters, it is
common to adopt pre-trained networks to build larger networks for better
performance; e.g., ResNet \citep{he2015delving} or DenseNet
\citep{huang2017densely} pretrained by ImageNet \citep{deng2009imagenet}.
Another emerging trend is the adoption of the transformer architecture
\citep{han2022survey, jaderberg2015spatial, khan2021transformers,
vaswani2017attention, wolf2020transformers}. 

Despite its rapid advance, the DL paradigm faces several challenges.  DL
networks are mathematically intractable, vulnerable to adversarial
attacks \citep{akhtar2018threat}, and demand heavy supervision.  Efforts
have been made in explaining the behavior of DL networks with certain
success, e.g., \citep{chan2022redunet, damian2022neural,
ma2022principles, oymak2021generalization,
soltanolkotabi2018theoretical, wright2022high}. Adversarial training has
been developed to provide a tradeoff between robustness and accuracy
\citep{zhang2019theoretically}.  Self-supervised \citep{misra2020self} and
semi-supervised learning \citep{sohn2020fixmatch, van2020survey} have
been explored to reduce the supervision burden. 

Two further concerns over DL technologies are less addressed. The first
one is about its high carbon footprint \citep{lannelongue2021green,
schwartz2020green, wu2022sustainable, xu2021survey}.  The training of DL
networks is computationally intensive. The training of larger complex
networks on huge datasets imposes a threat on sustainability
\citep{sanh2019distilbert, sharir2020cost, strubell2019energy}.  The
second one is related to its trusworthiness.  The application of
blackbox DL models to high stakes decisions is questioned
\citep{arrieta2020explainable, poursabzi2021manipulating, rudin2019stop}.
Conclusions drawn from a set of input-output relationships could be
misleading and counter intuition.  It is essential to justify an ML
prediction procedure with logical reasoning to gain people's trust. 

To tackle the first problem, one may optimize DL systems by taking
performance and complexity into account jointly.  An alternative
solution is to build a new learning paradigm of low carbon footprint
from scratch. For the latter, since it targets at green ML systems by
design, it is called green learning (GL).  The early development of GL
was initiated by an effort to understand the operation of computational
neurons of CNNs in \citep{kuo2017cnn, kuo2016understanding, kuo2018data,
kuo2019interpretable}. Through a sequence of investigations, building
blocks of GL have been gradually developed, and more applications have
been demonstrated in recent years.  As to the second problem, a clear
and logical description of the decision-making process is emphasized in
the development of GL.  GL adopts a modularized design. Each module is
statistically rooted with local optimization. GL avoids end-to-end
global optimization for logical transparency and computational
efficiency.  On the other hand, GL exploits ensembles heavily in its
learning system to boost the overall decision performance. GL yields
probabilistic ML models that allow trust and risk assessment with
certain performance guarantees. 

GL attempts to address the following problems to make the learning process 
efficient and effective:
\begin{enumerate}
\item How to remove redundancy among source image pixels for concise 
representations?
\item How to generate more expressive representations?
\item How to select discriminant/relevant features based on labels?
\item How to achieve feature and decision combinations in the design of
powerful classifiers/regressors? 
\item How to design an architecture that enables rich ensembles for 
performance boosting? 
\end{enumerate}
New and powerful tools have been developed to address each of them in
the last several years, e.g., the Saak \citep{kuo2018data} and Saab
transforms \citep{kuo2019interpretable} for Problem 1, the PixelHop
\citep{chen2020pixelhop}, PixelHop++ \citep{chen2020pixelhop++} and
IPHop \citep{yang2022design} learning systems for Problem 2, the
discriminant and relevant feature tests \citep{yang2022supervised} for
Problem 3, the subspace learning machine \citep{fu2022subspace} for
Problem 4. The original ideas scattered around in different papers. 
They will be systematically introduced here.

In this overview paper, we intend to elaborate on GL's development,
building modules and demonstrated applications. We will also provide an
outlook for future R\&D opportunities.  The rest of this paper is
organized as follows. The genesis of GL is reviewed in Sec.
\ref{sec:overview}.  A high-level sketch of GL is presented in Sec.
\ref{sec:sketch}. GL's methodology and its building tools are detailed
in Sec.  \ref{sec:fundamentals}.  Illustrative application examples of
GL are shown in Sec.  \ref{sec:applications}.  Future technological
outlook is discussed in Sec.  \ref{sec:outlook}.  Finally, concluding
remarks are given in Sec.  \ref{sec:conclusion}. 

\section{Genesis of Green Learning}\label{sec:overview}

The proven success of DL in a wide range of applications gives a clear
indication of its power although it appears to be a mystery.  Research
on GL was initiated by providing a high-level understanding of the
superior performance of DL \citep{kuo2016understanding, kuo2017cnn,
xu2017understanding}.  There was no attempt in giving a rigorous
treatment but obtaining insights into a set of basic questions such as:
\begin{itemize}
\item What is the role of nonlinear activation \citep{kuo2016understanding}?
\item What are individual roles played by the convolutional layers and
the fully-connected (FC) layers \citep{kuo2019interpretable}?
\item Is there a guideline in the network architecture design
\citep{kuo2019interpretable}?
\item Is it possible to avoid the expensive backpropagation optimization
process in filter weight determination \citep{kuo2018data, 
kuo2019interpretable, lin2022geometrical}?
\end{itemize}
As the understanding increases, it becomes apparent that one can develop
a new learning pipeline without nonlinear activation and backpropagation. 

\begin{figure}
\centering
\includegraphics[width=10cm]{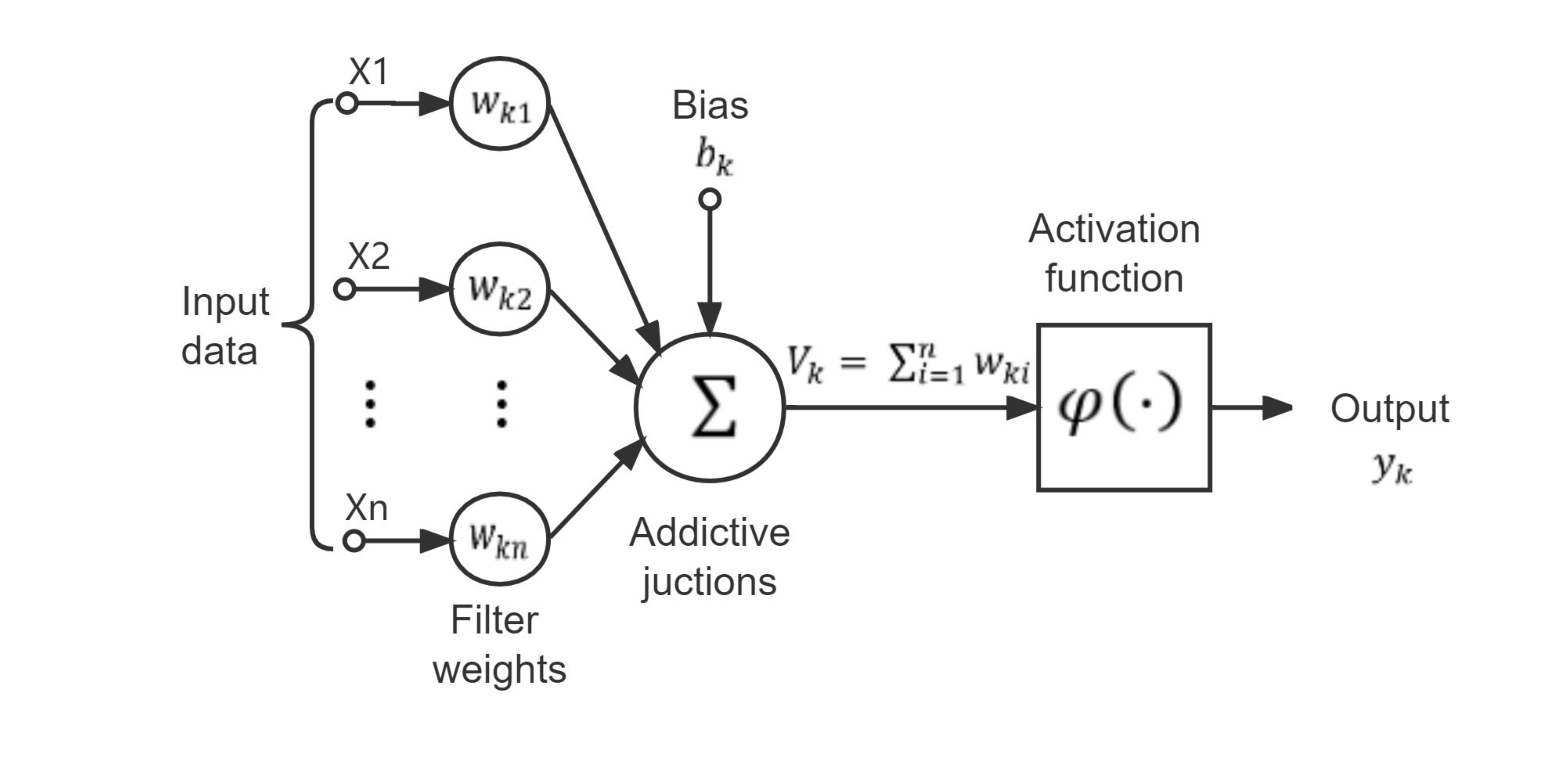}
\caption{Illustration of a computational neuron.}\label{fig:neuron}
\end{figure}

\subsection{Anatomy of Computational Neurons}\label{subsec:neurons}

As shown in Fig. \ref{fig:neuron}, a computational neuron consists of
two stages in cascade: 1) an affine transformation that maps an
$n$-dimensional input vector to a scalar, and 2) a nonlinear activation
operator. The affine transformation is relatively easy to understand.
Yet, nonlinear activation obscures the function of a computational
neuron. If the function of nonlinear activation is well understood, one
may remove it and compensate its function with another mechanism.  As
explained in Sec. \ref{subsec:networks}, one neural network achieves two
objectives simultaneously \citep{kuo2019interpretable}: 1) finding an
expressive embedding of the input data, and 2) making decision (i.e.,
classification or regression) based on data embedding. 

An affine transformation contains $n$ weights and bias $b$ for an input
vector of dimension $n$. According to the analysis in
\citep{fu2022subspace, lin2022geometrical}, weights and biases play
different roles. To determine proper weights of an affine transformation
is equivalent to finding a discriminant 1D projection for partitioning.
For example, one can use the linear discriminant analysis (LDA) to find
a hyper-plane that partitions the whole space into two halves.  The
weight vector is the surface normal of the hyper-plane. In other words,
the affine transformation defines a projection onto a line through the
inner product of an input vector and the weight vector. Then bias term
defines the split point in the projected 1D space (i.e., greater, equal
and less than zero).  The role of nonlinear activation was investigated
for CNNs in \citep{kuo2016understanding} and multi-layer perceptrons
(MLPs) in \citep{lin2022geometrical}.  It is used to avoid the sign
confusion problem caused by two neurons in cascade. 

\begin{figure}
\centering
\includegraphics[width=14cm,height=6cm]{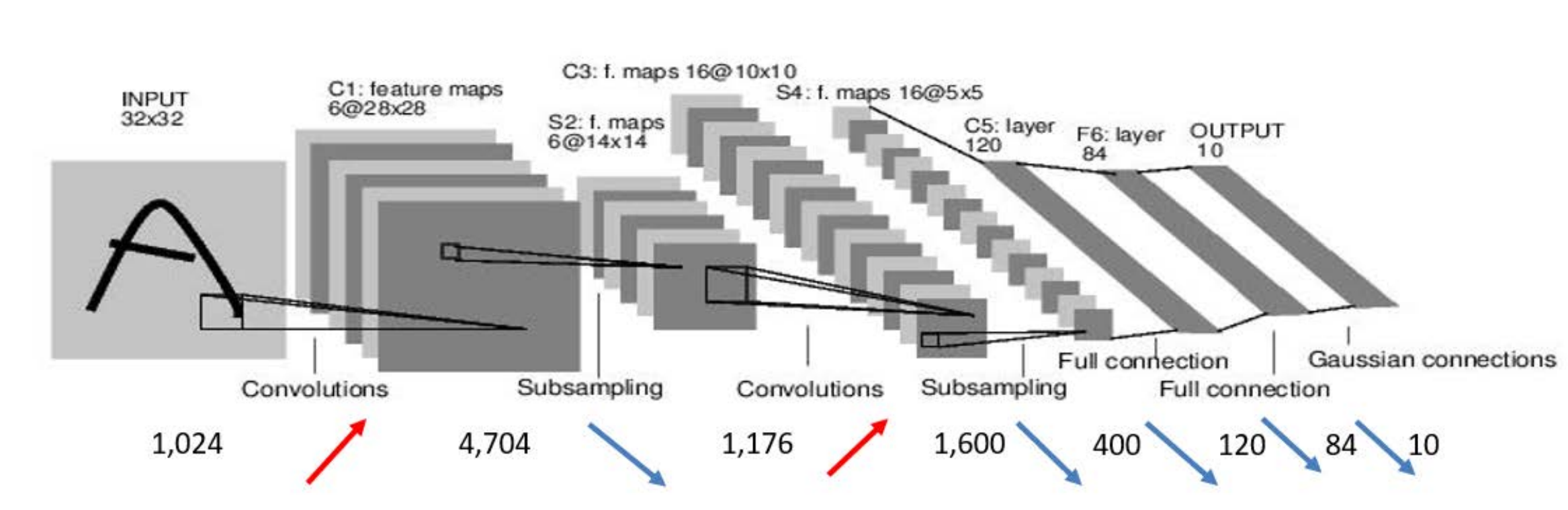}
\caption{Illustration of the LeNet-5 architecture \citep{LeNet1998},
where the dimensions of intermediate layers are shown under
the network.}\label{fig:LeNet-Dim}
\end{figure}

\subsection{Anatomy of Neural Networks}\label{subsec:networks}

MLPs have been commonly used as a classifier. The convolutional neural
networks (CNNs) can be viewed as the cascade of two sub-networks: the
convolutional layers and the FC layers. One simple example called the
LeNet-5 \citep{LeNet1998} is illustrated in Fig. \ref{fig:LeNet-Dim}.
It is convenient to have a rough breakdown of their functions as done in
\citep{kuo2019interpretable}. That is, the first sub-network is used to
find powerful embeddings while the second sub-network is used for
decision making.  Admittedly, this anatomy is too simplistic for more
complicated networks such as ResNet \citep{he2015delving}, DenseNet
\citep{huang2017densely} and Transformers \citep{vaswani2017attention}.
Yet, this viewpoint is helpful in deriving the GL framework.  We will
elaborate on the first and the second sub-networks in Secs.
\ref{subsubsec:embedding} and \ref{subsubsec:decision}

\subsubsection{Feature Subnet}\label{subsubsec:embedding}

Modern DL networks have inputs, outputs, and intermediate embeddings.
Their dimensions for image/video data are generally large. Consider two
examples below. 

\noindent
{\em Example 1.} An image in the MNIST dataset has gray-scale pixels of
spatial resolution $32\times 32$. Its raw dimension is 1,024. LeNet-5
\citep{LeNet1998} was designed for the MNIST dataset. It has two cascaded
convolutional-pooling layers. At the output of the 2nd
convolutional-pooling layers, one obtains an embedding space of
dimension $5 \times 5 \times 16 =400$. 

\noindent
{\em Example 2.} After image size normalization, an image in the
ImageNet dataset has color pixels of spatial resolution $224 \times
224$.  It has a raw dimension of $224\times 224 \times 3=150,528$.
AlexNet \citep{NIPS2012_AlexNet} and VGG-16 \citep{simonyan2014very} were
proposed to solve the object classification problem in the ImageNet
dataset. The last convolutional layers of AlexNet and VGG have an
embedding space of $13 \times 13 \times 256=43,264$ and $7 \times 7
\times 512=25,088$ dimensions, respectively. 

The embedding dimension of the last convolutional layer is significantly
smaller than the input dimension from the above two examples. Dimension
reduction is essential to the simplification of a decision pipeline. It
is well known that there are spatial correlations between neighboring
pixels of images, which can be exploited for dimension reduction.
However, dimension reduction is only one of two main functions of the
feature subnet. The feature subnet needs to find discriminant dimensions
at the same time.

The dimension variation at different stages of LeNet-5 is shown in Fig.
\ref{fig:LeNet-Dim}, where a red upward arrow denotes dimension
expansion and a blue downward arrow denotes dimension reduction.
Dimensions increase with convolution operations in the first two layers,
and dimension decreases with the (2x2)-to-(1x1) pooling operations.  The
increase in the dimension of an intermediate layer is determined by the
network architecture.  By adding more neurons and reducing the stride
number, the intermediate layer will have more dimensions. More embedding
variables allow the generation of more expressive embeddings.  The
design of network architecture highly depends on specific applications
(or datasets). The search of an optimal network architecture
\citep{elsken2019neural} and the associated loss function for new
problems is costly. 

\subsubsection{Decision Subnet}\label{subsubsec:decision}

MLPs have the universal approximation capability to an arbitrary
function \citep{cybenko1989approximation, hornik1989multilayer}.  Without
loss of generality, consider the mapping from an $n$-dimensional input
to an arbitrary 1D function. The latter can be approximated by a union of
piece-wise low-order polynomials. The basic unit of a piece-wise
constant (or linear) approximation is a box (or a triangular-shaped)
function of finite support. The form of activation determines the shape,
e.g., two step activations can be used to synthesize a box function.
Users are referred to \citep{lin2021relationship} for more detail.  For each
partitioned interval, the mean value of observed output samples (or the
labels of the majority class) can be assigned in a regression problem
(or a classification problem).  Decision making with space partitioning
is essential to all ML classifiers/regressors, e.g., the support vector
machine/regression (SVM/R), decision trees, random forests, gradient
boosting classifiers/regressors. Yet, no nonlinear activation is
required by them but neural networks. This difference will be explained
in Sec. \ref{subsec:decision}. 

\subsubsection{Filter Weight Determination}\label{subsubsec:weight}

One characteristics of DL systems is that filter weights can be
automatically adjusted by propagating decision errors backwards (i.e.,
backpropagation). As long as there exist negative gradients to lower the
loss function, the training loss will keep going down. Many advanced
network architectures are devised to allow more paths to avoid the
vanishing gradient problem.  Weight adjustment has different
implications in the feature subnet and the decision subnet.  For the
feature subnet, weights are adjusted to find the most expressive
embeddings under the architecture constraint. For the decision subnet,
weights are fine-tuned to find discriminant 1D projections for space
partitioning.  Sometimes, there may not be sufficient labeled data to
train DL networks.  To boost the overall performance, one may build a
modularized DL system, where filter weights of some modules are
pre-trained by other datasets. One example is ResNet/DenseNet pretrained
by ImageNet.  These pre-trained modules offer embeddings to interact
with other modules of the DL system. 

\subsection{Interpretable Feedforward-Designed CNN}\label{sec:ff-cnn}

The feedforward-designed convolutional neural network (FF-CNN)
\citep{kuo2019interpretable} plays a transitional role from DL to GL.
It follows the standard CNN architecture but determines the filter
weights in a feedforward one-pass manner. Filter weights in the
convolutional layers of FF-CNN are determined by the Saab transform
without supervision. 

Although a bias term is adopted by the Saab transform to address the
sign confusion problem in \citep{kuo2019interpretable}, this term is
removed in the later Saab transform implementation for the following
reason.  The sign confusion problem actually only exists in neural
network training because of the use of backpropagation optimization.  In
this context, one determines both the embeddings from the previous layer
and the filter weights of the current layer simultaneously based on the
desired output. In contrast, the input to the current layer is already
given in the feedforward design, one need to determine filter weights
first based on the statistics of the current input and then compute the
filter responses. As a result, there is no sign confusion problem in the
FF-CNN. 

Filter weights in the fully-connected (FC) layers of FF-CNN are
determined by linear least-squared regression (LAG). That is, in FF-CNN
training, one clusters training samples of the same class into multiple
sub-clusters and create a pseudo-label for each sub-cluster.  For
example, there are 10 labels (i.e. 0, 1, $\cdots$ , 9) for each image.
One can have 12 sub-clusters for each digit and create 120
pseudo-labels. The first FC layer maps from 400 latent variables in the
last convolution layer to 120 pseudo-labels. The determination of filter
weights can be formulated as a least-squared regression problem.  FF-CNN
has been used in the design of a privacy preservation framework in
\citep{hu2020slfb, wang2022privacy} and integrated with ensemble learning
\citep{chen2019ensembles} and semi-supervised learning \citep{chen2019semi}. 

\section{High-Level Sketch of GL}\label{sec:sketch}

\subsection{Overview}

As mentioned in Sec. \ref{sec:introduction}, GL tools have been devised to achieve
the following objectives:
\begin{enumerate}
\itemsep -1ex
\item Remove redundancy among source samples for concise representations.
\item Generate expressive representations.
\item Select discriminant/relevant features based on supervision (i.e., training labels).
\item Allow feature and decision combinations in classifier/regressor design.
\item Propose a system architecture that enables ensembles for performance boosting.
\end{enumerate}
These techniques are summarized in Table \ref{table:overview}.

\begin{table}[!ht]
\centering
\caption{A set of developed GL techniques.}\label{table:overview}
    \begin{tabular}{|l|l|l|l|} \hline
        Learning      & Need of     & Linear & Examples \\ 
        Techniques    & Supervision &  Operation &  \\ \hline
        Subspace      & No          & Yes  &  Saak Transform \citep{kuo2018data}, \\
        Approximation &    &        &  Saab Transform \citep{kuo2019interpretable} \\ \hline
        Expressive    & Maybe &  Maybe  &  Attention, Multi- \\ 
        Representation Generation   &       &        &  Stage Transform \citep{chen2020pixelhop++} \\ \hline
        Ensemble-enabled & Maybe & Maybe & PixelHop \citep{chen2020pixelhop} \\
        Architecture  &       &       & PixelHop++ \citep{chen2020pixelhop++} \\ \hline
        Discriminant & Yes & No     &  Discriminant  \\ 
        Features Selection   &  &             &  Feature Test \citep{yang2022supervised} \\ \hline
        Feature Space & Yes & No        &  Subspace Learning \\ 
        Partitioning &     &           &  Machine \citep{fu2022subspace} \\ \hline
    \end{tabular}
\end{table}

It is worthwhile to comment on differences and relationship between GL,
DL and classical ML.  Traditional ML consists of two building blocks:
feature design and classification.  Feature design is typically based on
human intuition and domain knowledge. Feature extraction and decision
are integrated without a clear boundary in DL.  Once the parameters of
DL networks are determined by end-to-end optimization, feature design
becomes a byproduct. Techniques No. 1-3 in Table \ref{table:overview}
correspond to the feature design in GL.  Techniques 1 and 2 can be
automated with little human involvement. Only hyper-parameters are
provided by humans, which is similar to the network architecture design
in DL. Technique 3 provides a feedback from labels to the learned
representations so as to zoom into the most powerful subset. Unlike
traditional ML, it does not demand human intuition or domain knowledge.
Finally, the last module in GL is the same as the classifier in
traditional ML. 

The reason to decompose the feature design module in classical ML into
three individual steps in GL is for the purpose of automation.  For
simplicity and yet without loss of generality, we use an illustrative
example to explain the four modules of GL below.  Consider the ML
problem of recognizing 10 handwritten digits (i.e. 0, 1, $\cdots$, 9)
with the MNIST dataset. It has 60,000 training images and 10,000 test
images.  Each image has a spatial resolution of $32\times 32$ pixels,
where each pixel has 256 gray scales.  Typically, we treat each pixel as
one dimension so that the input space has $32 \times 32 = 1024$
dimensions.  Since there are 10 output classes, we assign 10 dimensions
to the output space, denoted by $(p_0, \, p_1, \cdots, p_9)^T$, where
$p_i$ is the probability of the test image in class $i$.  Note that this
problem can be well solved by a neural network solution called LeNet-5
\citep{LeNet1998}. Although LeNet-5 is a shallow network, its
architecture can be generalized to deeper networks such as AlexNet
\citep{NIPS2012_AlexNet} and VGG-16 \citep{simonyan2014very} in a
straightforward manner. 

\subsection{Subspace Approximation}\label{subsec:approx}

Each pixel in an image is too weak to be a discriminant for any image
content.  Thus, we group a set of neighboring pixels to form a basic
unit. Typical units are squared blocks of size $3\times 3$ or $5 \times
5$. There exists correlation among pixels in PixelHop units. The
correlation can be reduced or removed by signal transforms.  Transform
kernels can be either predefined or derived from the input data.  The
discrete cosine transform (DCT) and the wavelet transform use predefined
kernels while the Karhunen-Lo$\acute{e}$ve transform (KLT) adopts the
data-driven transform kernel. Two new data-driven transforms have been
introduced to tackle Subproblem 1 in the GL setting. They are the Saab
(Subspace approximation via adjusted bias) transform
\citep{kuo2019interpretable} and the Saak (Subspace approximation via
augmented kernels) transform \citep{kuo2018data}. These new transforms
will be reviewed in Sec.  \ref{subsec:representation}. No supervision is
needed in learning the approximating subspace. 

\subsection{Generation of Expressive Representations}\label{subsec:expressive} 

There are many ways to conduct the 2D transform for images in the MNIST
dataset. We compare the following two designs. 
\begin{itemize}
\item[A] One-stage transform. \\
The transform has a kernel of size $32 \times 32=1024$. The transform
input and output have the same dimension; namely, 1024.  Since the
element-wise energy is not uniformly distributed for the 1024D output,
we drop those elements of extremely low energy to form a representation 
vector. 
\item[B] Two-stage transform. \\
The first-stage transform has a kernel size of $5 \times 5$. It is
applied to $28 \times 28$ interior pixels with stride equal to one.
Then, we take the absolute value of each response and conduct the
maximum pooling over non-overlapping $2\times2$ blocks. This leads to a
tensor output of dimension $14 \times 14 \times 25$, where $14 \times
14$ indicate the number of spatial locations and $25$ is the channel
response of the first-stage transform. Finally, we conduct the
second-stage transform to $14 \times 14$ locations for each channel. The
output of the second-stage transform still have a dimension of $14
\times 14 \times 25$. Again, we can drop those elements of extremely low
energy to form a representation vector. 
\end{itemize}

Computationally, the one-stage transform is less efficient than the
two-stage transform due to its large kernel size.  Here, we pay
attention to another issue - the expressiveness of derived
representations.  For fair comparison, we select the same element number
(e.g., 512) of the highest energy from designs A and B as the
representation vector.  We may ask which representation set is more
useful in the classification problem. For design A, only a few
low-frequency transform coefficients are helpful since a vast majority
of high-frequency transform coefficients are not stable for images of
the same label.  For design B, the first-stage transform captures the
local variation while the second-stage transform characterizes the
global variation. It has more useful representations since we can obtain
more stable associations between transform coefficients and the image
class. We use expressiveness to differentiate these two representation
sets. 

In this example, no supervision is used to derive expressive
representations.  However, it is sometimes useful to exploit image
labels to derive the expressive representation set. One example is
attention detection.  Consider the classification of dog and cat images.
Both dog and cat images have background, which is irrelevant to the
classification task.  It is desired to remove the background effect and
focus on the main object, which is called the attention. To obtain
attention, we can decompose images into smaller blocks, extract
block-level representations and use image labels to train the
probability of a block to be a dog or a cat. For the foreground object,
a test cat (or dog) block is expected to have a higher probability for
the cat (or dog) decision.  For a background block, if has the same
likelihood to be with cat and dog images, its test probability will be
about equal (i.e., 50
it has a higher likelihood to be with cat (or dog) images, then its test
probability will be higher for the cat (or dog) class. As a result, we
can use blocks with more skewed probability distribution as attention.
Clearly, attention offers expressive representations that have to be
trained by image labels. 

Subspace approximation is used to reduce redundancy in representation.
However, it is still desired to find more expressive representations
that are valuable in solving the target learning problem. The former is
straightforward while the latter is not.  This makes GL challenging yet
interesting. Since they are closely related to each other in finding an
effective representation of a data source, we treat them as a whole.
Details are presented in Sec.  \ref{subsec:representation}. 

\subsection{Ensemble-Enabled Architectures}\label{subsec:ensemble-enabled}  

Filter responses of DL architectures are latent variables.  Their values
keep evolving during the end-to-end optimization process since the
filter weights are adjusted through backpropagation. The exact values of
these individual latent variables are not important since they are
affected by many factors such as filter weight initialization and the
detailed implementation of the stochastic gradient descent optimization
(e.g., the order of mini-batches and dropout). People are mainly concerned
with the converged neural networks as a whole.

In contrast, filter weights in GL are determined in a feedforward
one-pass manner. Once computed, they stay the same and do not change any
longer.  The same GL architecture will have the same filter weights and
responses. GL needs other mechanisms to boost the learning performance.
One idea is the "ensemble-enabled architectures" as demonstrated in
PixelHop, PixelHop++, PointHop and PointHop++.  Filter responses at
various stages define joint spatial-spectral representations in form of 3D
tensors. The 3D tensors in earlier (or later) stages have higher (or
lower) spatial dimensions and smaller (or higher) spectral dimensions.
They have different physical meanings. GL can conduct feature ensembles
and/or decision ensembles to achieve better performance.

\subsection{Discriminant Features Selection}\label{subsec:DFT}  

Feature selection methods can be categorized into unsupervised
\citep{solorio2020review}, semi-supervised \citep{sheikhpour2017survey},
and supervised \citep{huang2015supervised} three types.  Unsupervised
methods focus on the statistics of each feature dimension while ignoring
the target class or value. However, their power is limited and less
effective than supervised methods.  Recently, Yang {\em et al.}
\citep{yang2022supervised} proposed a supervised learning tool in
selecting discriminant features and call it the discriminant feature
test (DFT).  DFT can rank the discriminant power of each dimension from
the highest to the lowest and determine a threshold that separates
discriminant dimensions automatically.  It offers an immensely powerful tool
for discriminant feature selection and will be reviewed in Sec.
\ref{subsec:features}. 

\subsection{Feature Space Partitioning}\label{subsec:partitioning}   

After discriminant feature selection, the last step is classification
that maps a sample in the feature space to a label. There are two main
ideas to implement the classification task:
\begin{itemize}
\item[A] {\em Change of variables from input features to output
label-probability vector.} Examples include the logistic regression
classifier \citep{dreiseitl2002logistic}, the linear least-squares
regression (LLSR) classifier \citep{loh2011classification}, and the
multilayer perceptron (MLP) \citep{rosenblatt1958perceptron}. 
\item[B] {\em Partitioning of the input feature space into purer cells
whose samples are mostly from the same class.} There are two major
families: 1) the support vector machine (SVM) \citep{svm} and 2) the
decision tree (DT) \citep{CART}, and its variants such as random forests
\citep{RF} and XGBoost \citep{chen2016xgboost, chen2015xgboost}. 
\end{itemize}
Recently, Lin {\em et al.} \citep{lin2022geometrical} provides a geometrical
interpretation to MLP that builds a bridge between the above two idea.
Furthermore, Fu {\em et al.} \citep{fu2022subspace} proposed an enhanced
version of DT known as the subspace learning machine (SLM) tree as well
as SLM Forest and SLM Boost. They will be described in Sec.
\ref{subsec:decision}. 

\section{GL Methodologies}\label{sec:fundamentals}

\subsection{Overview of GL Systems}\label{subsec:system}

\begin{figure*}[tb]
\centering
\includegraphics[width=0.95\linewidth]{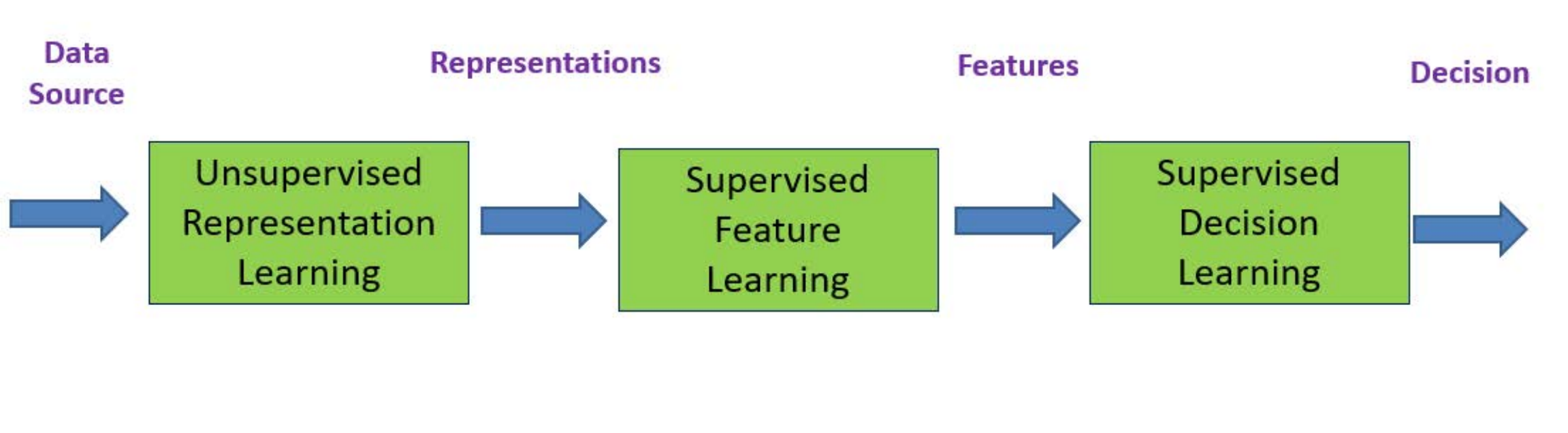}
\caption{An overview of a generic GL system, which consists of three
modules in cascade: 1) unsupervised representation learning, 2)
supervised feature learning and 3) supervised decision learning.}
\label{fig:GL_system}
\end{figure*}

The architecture of a GL system departs from that of neural networks
completely.  A GL system does not have the computational neuron as its
basic unit.  Its global architecture is not a network, either.  As
illustrated in Fig. \ref{fig:GL_system}, a generic GL system consists
of three learning modules in cascade: 1) unsupervised representation
learning, 2) supervised feature learning and 3) supervised decision
learning. The input and the output of the system are data sources and
decisions, respectively. The two intermediate results are
representations and features. They are not latent but explicit and
observable.  The entire system is characterized by three traits:
feedforward design, individually optimized module, and statistical
ensembles. 

\begin{figure*}[tb]
\centering
\includegraphics[width=0.95\linewidth]{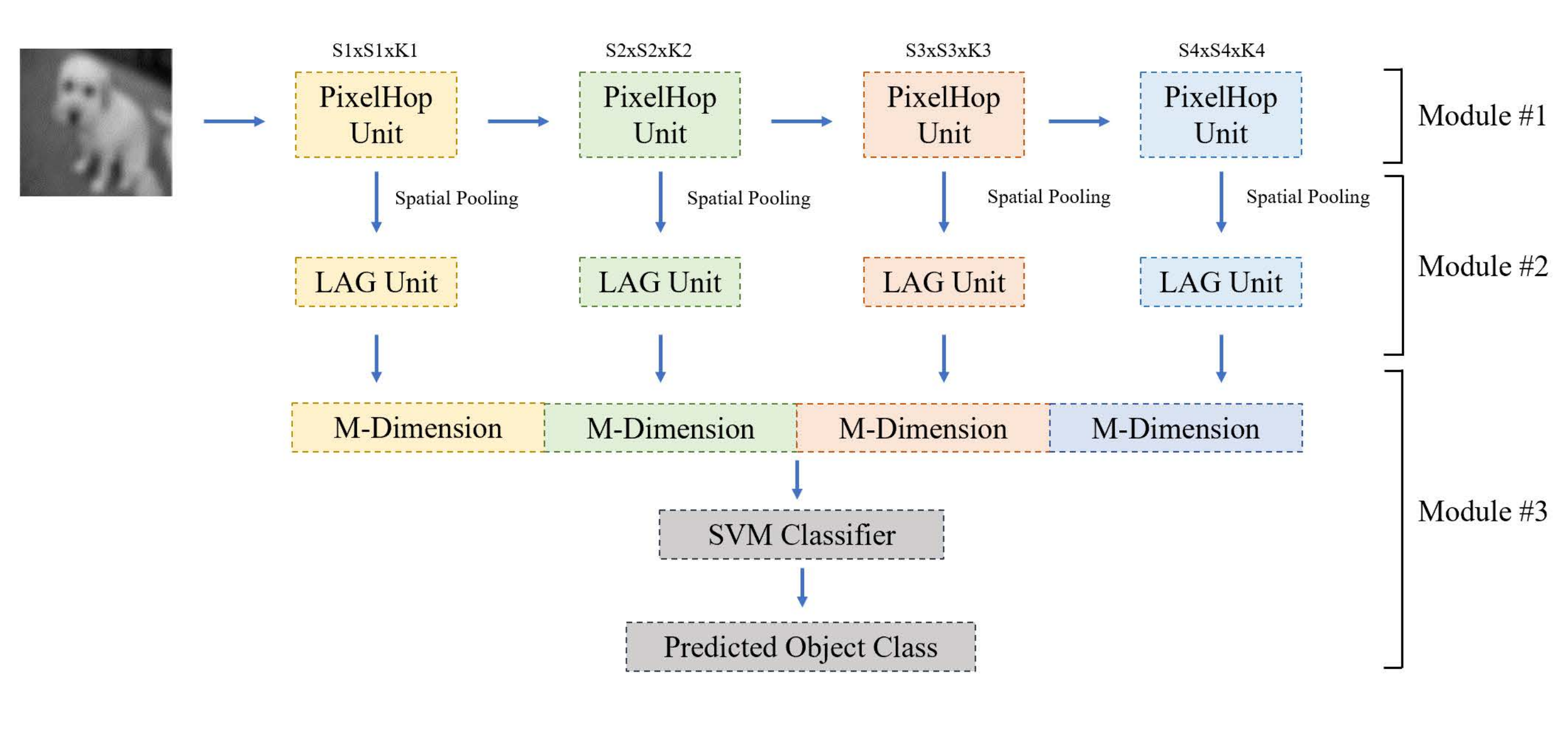}
\caption{The block diagram of the PixelHop learning system 
\citep{chen2020pixelhop}.}\label{fig:pixelhop}
\end{figure*}

A concrete example of the GL system, called PixelHop
\citep{chen2020pixelhop}, is shown in Fig.  \ref{fig:pixelhop}. Its
three modules are elaborated below. 
\begin{itemize}
\item Module \#1: Derive a rich set of representations from image sources 
through a sequence of PixelHop units in cascade.  \\
The main purpose of this module is to derive attributes from near-to-far
pixel neighborhoods of source images without supervision (i.e. labels).
Each PixelHop unit takes the neighborhood of any pixel as the input and
learns its spectral representation. The Saab transform is used for
dimension reduction.  Furthermore, one can use spatial pooling to reduce
the dimension of the whole image. The pooling process helps enlarge the
neighboring size in the following PixelHop unit. 
\item Module \#2: Select useful features from a large set of representations. \\
Each PixelHop unit yields new spatial-spectral representations of the
input. Their dimensions are still high. The main purpose of this module
is to find a smaller set of discriminant features from a larger set of
representations based on the label information for the desired task. The
Label-Assisted reGressor (LAG) unit was proposed in PixelHop to achieve
this objective. A more powerful tool was recently developed to replace
the LAG unit in \citep{yang2022supervised}, which will be discussed in
Sec.  \ref{subsec:features}. 
\item Module \#3: Conduct feature ensembles and the final classification task. \\
All features across multiple PixelHop units are concatenated to form an
ensemble feature set. Then, they are fed into an ML classifier for the
final classification task. 
\end{itemize}

\subsection{Unsupervised Representation Learning}\label{subsec:representation}

\subsubsection{Single-Stage Data-Driven Transforms}\label{subsubsec:one-stage}

Signal transforms are widely used in signal processing to reveal
spectral properties of underlying signals. Many signal transforms are
data independent such as the Fourier transform, the discrete Cosine
transform (DCT), and the wavelet transform. However, there are
transforms that are data-driven. One famous example is KLT, which is
also known as the principal component analysis (PCA). KLT can
decorrelate the correlations among elements of an input signal vector
and generate a compact description.  In the context of computer vision,
the eigenface for face recognition \citep{turk1991eigenfaces} was
developed based on the 2D KLT. 

Saak and Saab transforms were proposed to define filter weights of a
computational neuron in convolutional layers of CNNs.  Let us use the
neuron in Fig. \ref{fig:neuron} as an example.  The dimension of its
input vector ${\bf x}$ is $n$. Its KLT kernel is in form of ${\bf h}_k =
(h_{k1}, \cdots, h_{k,n})^T$, $k=1, \cdots, n$. However, we may choose a
smaller kernel number for lossy approximation. Typically, eigenvalue
$\lambda_k$ is ordered in a decreasing order of $k$. If $k=n$, this
leads to the lossless Saab transform.  If the kernel number is reduced
from $k$ to $k' \le k$, this leads to the lossy Saab transform.  The
Saak transform demands that filter weights appear in pairs; namely, both
${\bf h}_k$ and $-{\bf h}_k$ co-exist so that it demands $2k'$ neurons.
It is desired to reduce the neuron number from $2k'$ back to $k'$. 

Also, input ${\bf x}$ has to be a mean-removed vector in PCA.  Although
one can remove the mean of the input in the training, there is no
guarantee that the training and testing data have the same mean values.
To address the mean mismatch problem, the Saab transform introduce a DC
kernel whose elements have the same value. The operation of the DC
kernel on the input yields a local patch mean. For images that satisfy
the ergodic property, the local mean can provide a good approximation to
the ensemble mean. Then, one can remove the local-mean of the input and
conduct PCA to generate AC kernels. For the example in Fig.
\ref{fig:neuron}, one DC kernel and $(k'-1)$ AC kernels serve as the
filter weights of $k'$ neurons. To remove the effect of nonlinear
activation, all filters at the same layer add the same bias such that
all filter responses are positive. This filter weight design methodology
is computationally efficient. Besides, it allows a simple interpretation
of the role of convolutional layers. 

\subsubsection{Multi-Stage Transforms in Cascade}\label{subsubsec:multi-stage}

There exist long-, mid-, and short-distance correlations in images.
Such correlations are handled by multiple convolutional layers in
cascade in CNNs. To enlarge the receptive field of a deeper layer in
CNNs, there are two choices: 1) adopting a larger stride number, and 2)
applying the max-pooling operation at the output of each neuron. The
same idea is followed by Module \#1 in the PixelHop system. 

\begin{figure*}[tb]
\centering
\includegraphics[width=0.75\linewidth]{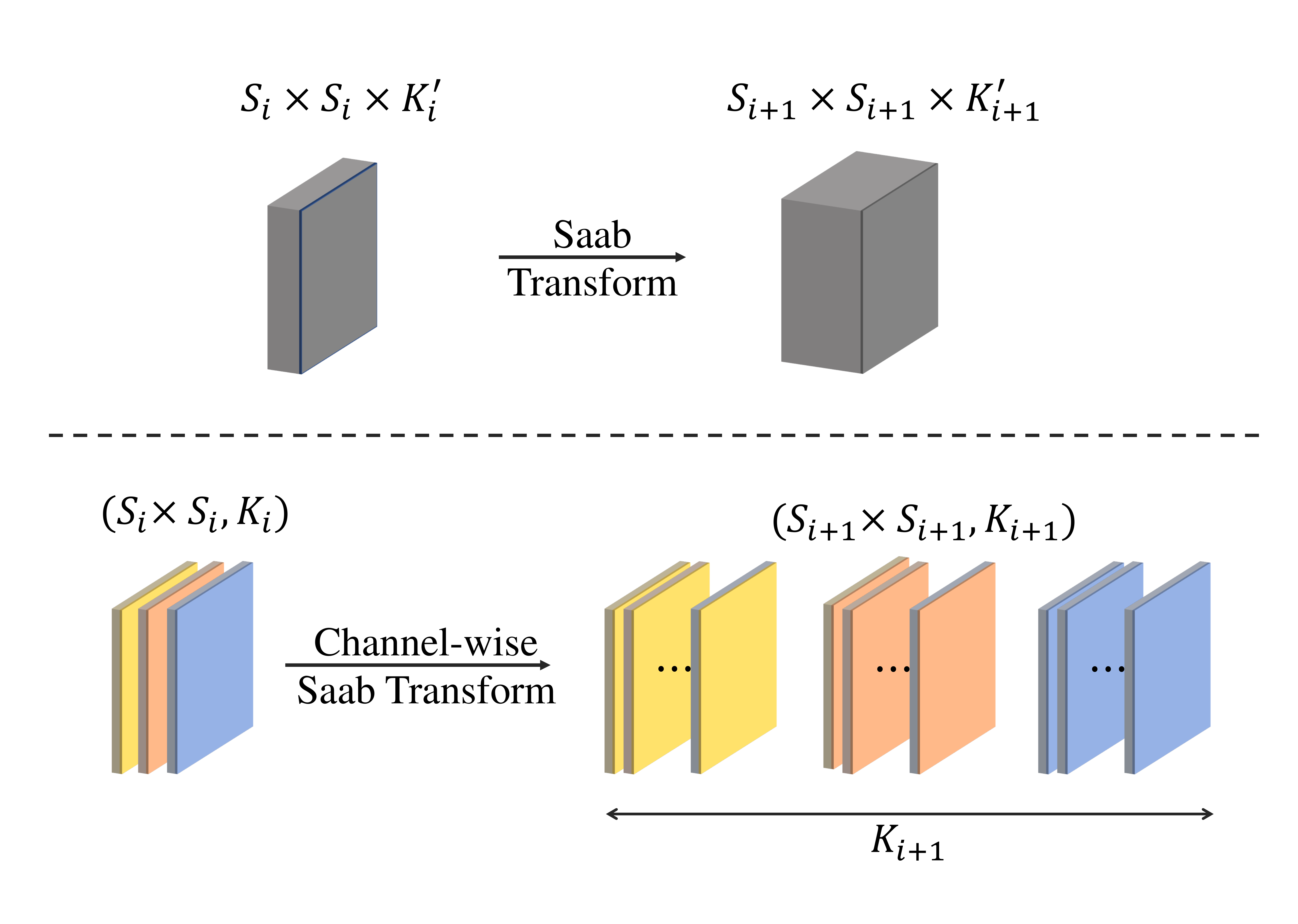}
\caption{Illustration of the one-stage channel-wise (c/w) Saab
transform, which decomposes a 3D tensor input into multiple 2D tensor
inputs and multiple 2D Saab transforms (denoted by different colors
individually) are conducted accordingly \citep{chen2020pixelhop++}.}
\label{fig:cwsaab_1}
\end{figure*}

Saab coefficients are weakly correlated in the spectral domain due to
the use of PCA in their derivation.  As show in Fig.
\ref{fig:cwsaab_1}, the weak spectral correlation enables the
decomposition of a 3D (i.e., 2D spatial and 1D spectral) input tensor of
dimension $S_i \times S_i \times K_i$ into $K_i$ spatial tensors of
neighborhood size $S_i \times S_i$ (i.e., one for each spectral
component) for the $i$th PixelHop unit, respectively. Then, instead of
performing the Standard 3D Saab transform as given in the top subfigure,
one applies $K_i$ channel-wise (c/w) Saab transforms to each of the
spatial tensors as shown in the bottom subfigure. There are two
advantages for c/w Saab transform. First, for the lossless Saab
transform, the model size of the c/w Saab transform is smaller than that
of the standard 3D Saab transform. Second, for the lossy Saab transform,
the model size saving of the c/w Saab transform can be even more
substantial. 

\begin{figure*}[tb]
\centering
\includegraphics[width=0.95\linewidth]{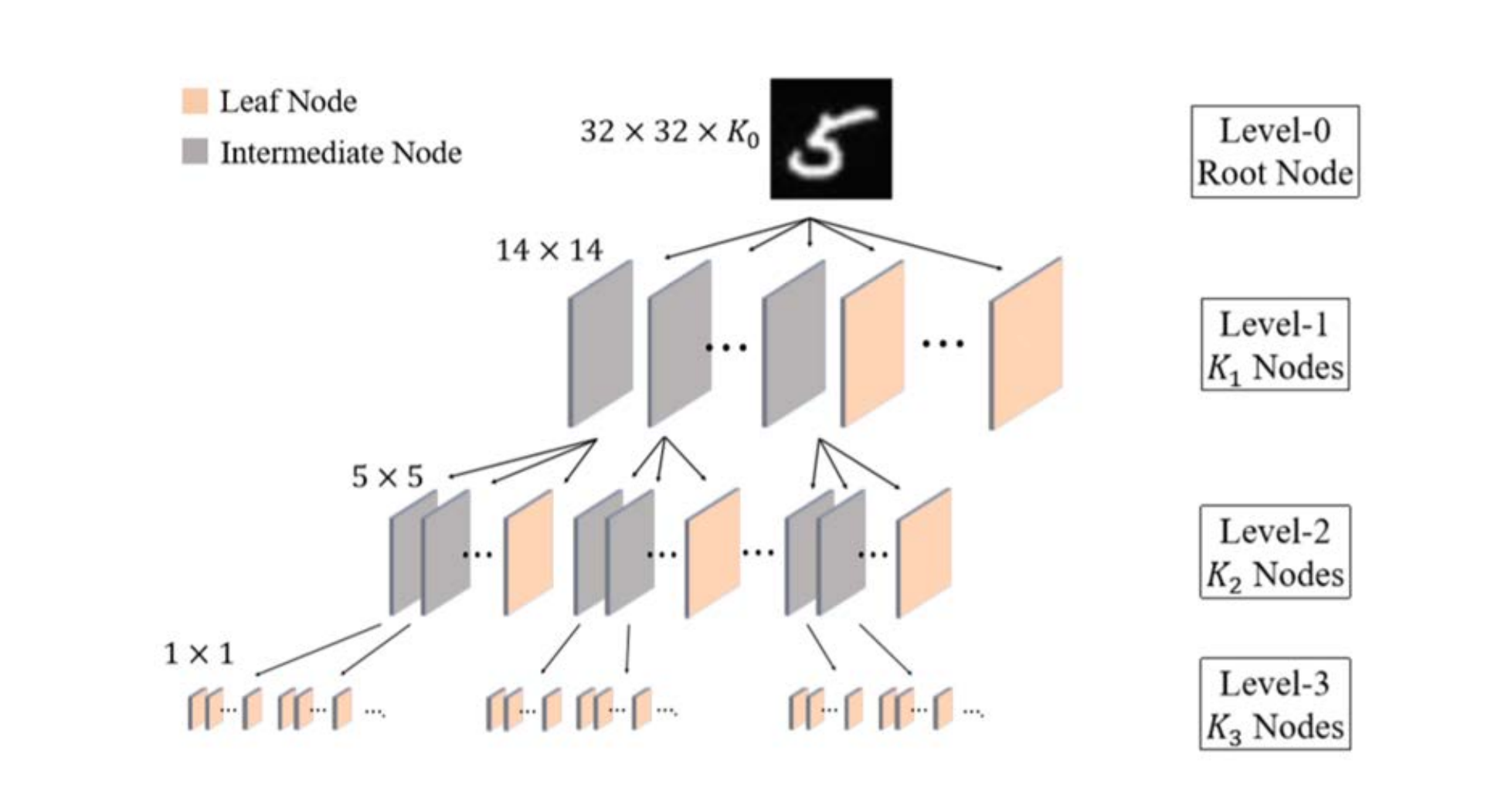}
\caption{Illustration of multi-stage lossy c/w Saab transforms, where
2D Saab transforms are only conducted at selected low frequency
channels \citep{chen2020pixelhop++}.}\label{fig:cwsaab_2}
\end{figure*}

To check the first point, let $n_{3D}=S_i^2 K_i$ and $n_{C/W}=S_i^2$
denote the dimensions of the inputs to the standard and c/w Saab
transforms. Then, their model sizes are:
\begin{eqnarray}
\mbox{Standard Saab:} & n_{3D}^2  = & S_i^4 K_i^2, \\
\mbox{C/W Saab:}      & K_i n_{C/W}^2  = & S_i^4 K_i,
\end{eqnarray}
To see the second point, we show the multi-stage lossy c/w Saab
transform for the whole image of size $32\times 32$ as an example in
Fig.  \ref{fig:cwsaab_2}. For high frequency components, the spatial
correlation of pixels in a local patch is so weak that its covariance
matrix is a nearly diagonal matrix, and there is no need to conduct the
Saab transform afterwards. These channels are colored in pink and called
leaf nodes. On the other hand, for low frequency components, the spatial
correlation of pixels in a local patch is still strong so that the c/w
Saab transform is conducted. They are colored in gray and called
intermediate nodes. 

\subsubsection{Enrichment of Expressive Representations}\label{subsubsec:enrichment}

\begin{figure*}[tb]
\centering 
\includegraphics[width=0.95\linewidth]{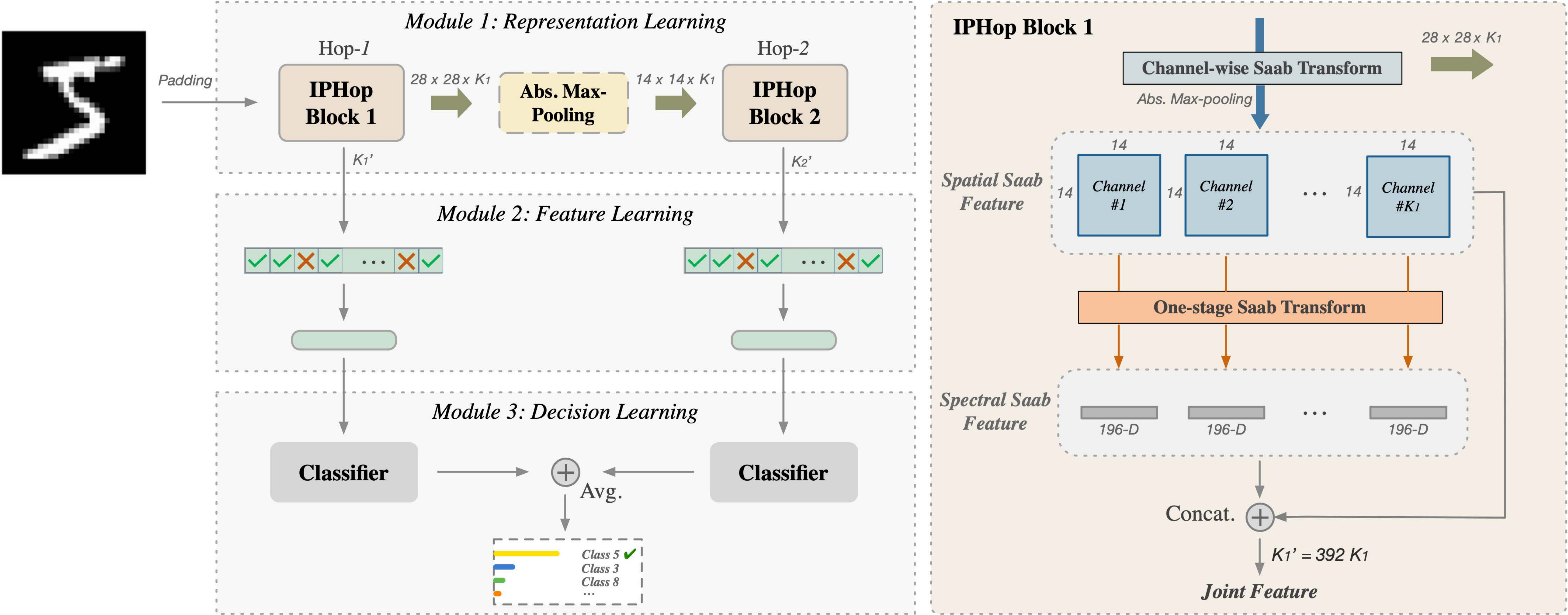}
\caption{The block diagram of the IPHop learning system
\citep{yang2022design}, where IPHop is the acronym of the improved
Pixelhop.}\label{fig:iphop}
\end{figure*}

There are more ways to generate expressive representations. Two ideas
were proposed in IPHOP \citep{yang2022design}. IPHop is an improved
PixelHop system. Its system diagram is shown in Fig.~\ref{fig:iphop},
which has two hops. 

The first module of IPHop is similar to that of PixelHop++ except
that max-pooling in PixelHop++ is changed to absolute-max-pooling. That
is, the responses from the previous stage can be either positive or
negative (without adding the bias term). Instead of clipping negative
values to zero as done in the ReLU unit, we take the absolute value of
all responses followed by the maximum pooling operation. The
absolute-max-pooling operation gives a small gain over the
straightforward max-pooling since the former preserves the low frequency
components of the spatial responses at the output of any single Saab
filter. 

The filter responses extracted at Hop-1 and Hop-2 only have a local view
on a larger object due to the limited receptive field size. Thus, they
are not discriminant enough.  For a given Saab filter, its responses are
spatially correlated. To decorrelate them, one can conduct another Saab
transform across them at each channel. This operation provides
channel-wise spectral Saab representations at Hop-1 and Hop-2 as shown
in the right subfigure of Fig.~\ref{fig:iphop}.  As compared to
representations learned by enlarging the neighborhood range gradually,
the channel-wise spectral Saab representations can capture the long
range correlation at a fine scale.  The spatial and spectral Saab
representations are concatenated at Hop-1 and Hop-2 to form
joint-spatial-spectral Saab representations. 

\subsection{Supervised Feature Learning}\label{subsec:features}

The leverage of labels (i.e., supervision) effectively to boost the
performance of a learning system is a key question in ML. Classical ML
only exploits labels in the classifier design. DL uses labels to adjust
filter weights in the feature subnet and the decision subnet and
achieves better performance. The linear least-squared regression adopted
by FF-CNN is also an example of supervised feature learning.  We
discuss an effective supervised learning tool in this subsection. 

Existing semi-supervised and supervised feature selection methods can be
classified into wrapper, filter and embedded three classes
\citep{sheikhpour2017survey}. Wrapper methods \citep{kohavi1997wrappers}
create multiple models with different subsets of input features and
select the model containing the features that yield the best
performance. One example is recursive feature elimination (RFE)
\citep{rfe}. This process can be computationally expensive.  Filter
methods involve evaluating the relationship between input and target
variables using statistics and selecting those variables that have the
strongest relation with the target ones.  One example is the analysis of
variance (ANOVA) \citep{anova}. This approach is computationally
efficient with robust performance. Embedded methods perform feature
selection in the process of training and are usually specific to a
single learner. One example is ``feature importance" (FI) obtained from
the training process of the XGBoost classifier/regressor
\citep{chen2016xgboost, chen2015xgboost}, which is also known as
``feature selection from model" (SFM).  Recently, the discriminant
feature test (DFT) and the relevant feature test (RFT) were proposed in
\citep{yang2022supervised} for the classification and the regression
problems, respectively. They belong to the filter class. 

\begin{figure*}[tb]
\centering
\includegraphics[width=0.95\linewidth]{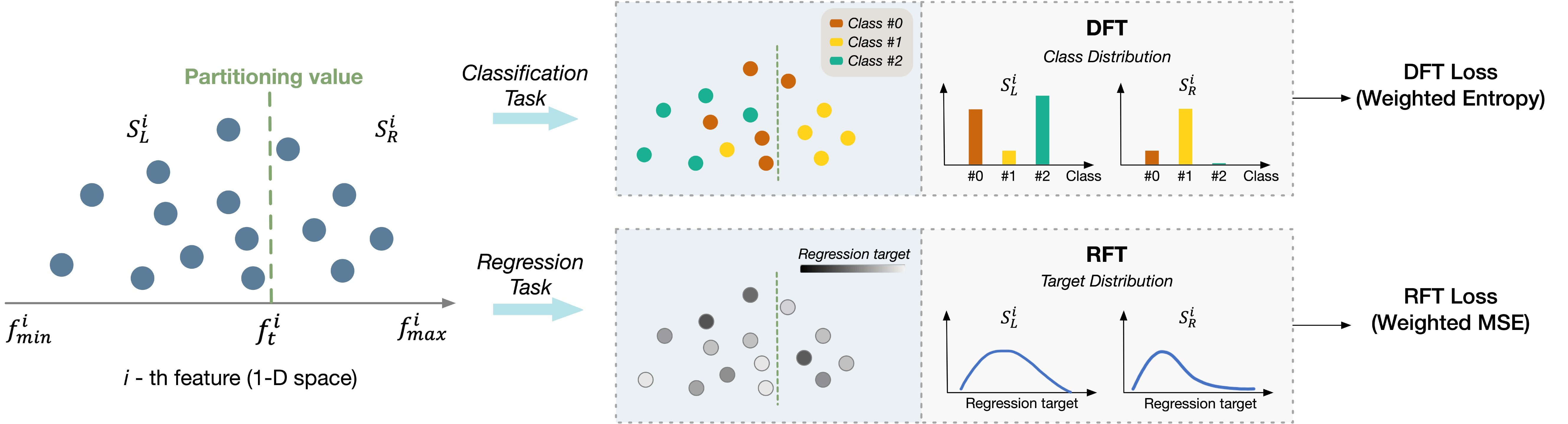}
\caption{The block diagram of two supervised feature selection methods:
the discriminant feature test (DFT) and the relevant feature test (RFT).
For the $i$th feature, DFT measures the class distribution in $S^i_L$
and $S^i_R$ to compute the weighted entropy as the DFT loss while RFT
measures the weighted estimated regression MSE in both sets as the RFT
loss. They are used to select discriminant/relevant features from a
large set of representations learned from the source without labels
\citep{yang2022supervised}.}\label{fig:dft}
\end{figure*}

Consider a classification problem with $N$ data samples, $P$ features
and $C$ classes. Let $f^i$, $1 \le i \le P$, be a feature dimension and
its minimum and maximum are $f_{\min}^i$ and $f_{\max}^{i}$,
respectively.  DFT is used to measure the discriminant power of each
feature dimension out of a $P$-dimensional feature space independently.
If feature $f^i$ is a discriminant one, one expects that data samples
projected to it should be classified more easily. To check it, an idea
is to partition $[f_{\min}^i, f_{\max}^{i}]$ into $M$ non-overlapping
subintervals and adopt the maximum likelihood rule to assign the class
label to samples inside each sub-interval. Then, one can compute the
percentage of correct predictions. The higher the prediction accuracy,
the higher the discriminant power. Although prediction accuracy may
serve as an indicator for purity, it does not tell the distribution of
the remaining $C-1$ classes if $C>2$. Since the weighted entropy measure
provides more valuable information, it was adopted in
\citep{yang2022supervised}. 

\begin{figure*}[tb]
\centering
\includegraphics[width=0.55\linewidth]{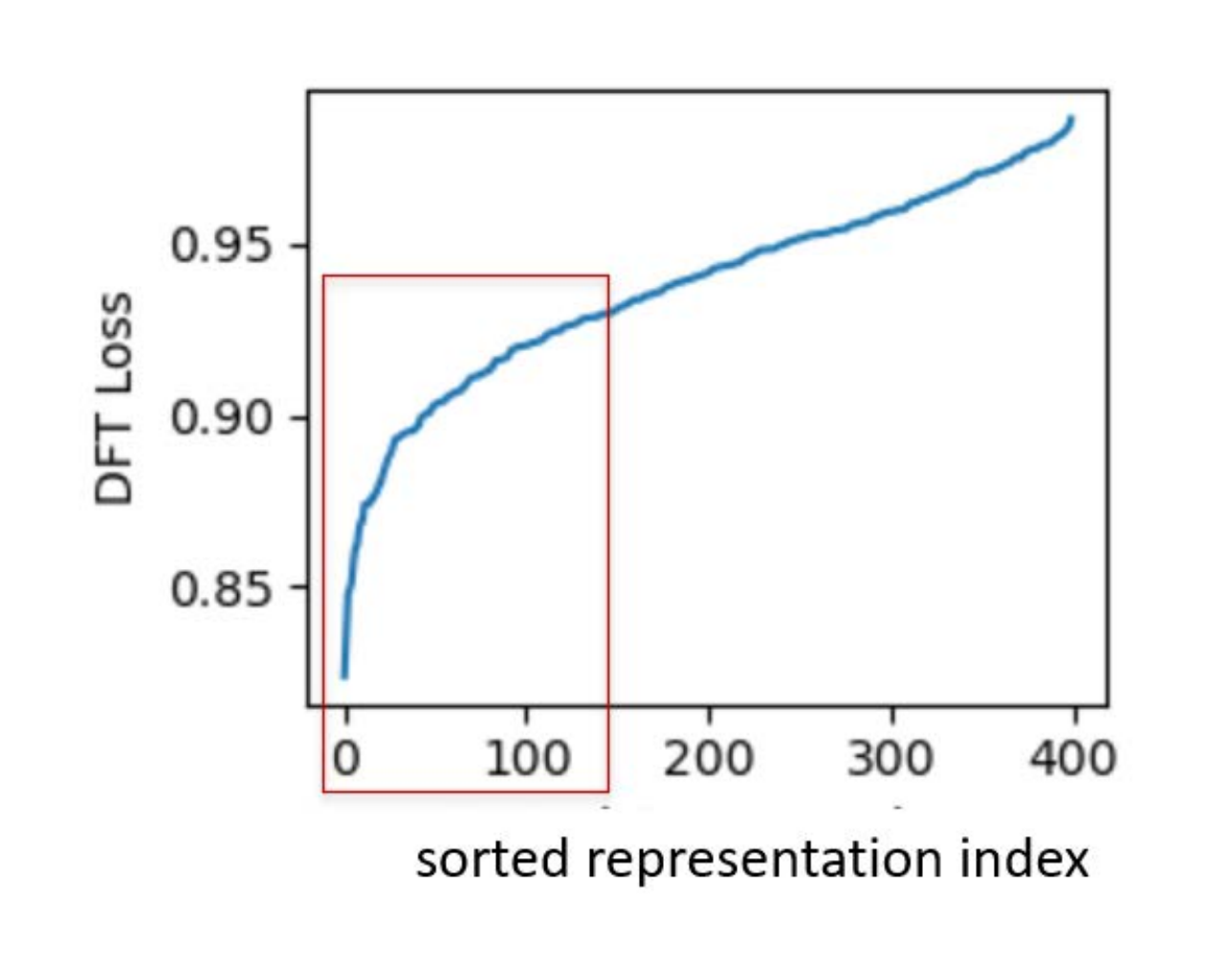}
\caption{The plot of a representative DFT curve, where the $x$-axis
denotes sorted representations, and the $y$-axis shows the optimal cost
value of an individual representation. The red box encloses
representations that are selected as features.}\label{fig:dft_curve}
\end{figure*}

By following the practice of a binary decision tree, the work in
\citep{yang2022supervised} chose $M=2$ as shown in the left subfigure of
Fig. \ref{fig:dft}, where $f_t^{i}$ denotes the threshold position of
two sub-intervals.  For a sample with its $i$th dimension, $x^{i}_n <
f_t^{i}$, it goes to the subset associated with the left subinterval.
Otherwise, it will go to the subset associated with the right
subinterval. DFT and RFT uses the weight entropy and the weighted
mean-squared-error (MSE) as the cost functions, respectively.  The
optimal cost values are computed for each feature dimension. A
representation is more discriminant (or relevant) if its optimal cost
value is lower.  To select a set of discriminant features, one can sort
representations based on their optimal cost values from the lowest to
the highest and plot the curve accordingly. A representative DFT curve
is shown in Fig.  \ref{fig:dft_curve}. An elbow point can be identified
in most applications. Both early and late elbow points were considered
in \citep{yang2022supervised}. A late elbow point was illustrated in
Fig.  \ref{fig:dft_curve}. The set of representations with optimal cost
values lower than the elbow point are selected as features. They
are enclosed by a red box in Fig.  \ref{fig:dft_curve}.

\subsection{Supervised Decision Learning}\label{subsec:decision}

Consider a classification problem with $N$-dimensional feature vector,
denoted by ${\bf f}=(f_1, \cdots, f_N)^T$, and $M$-dimensional decision
vector, denoted by ${\bf d}=(d_1, \cdots, d_M)^T$. The latter means that
there are $M$ classes and $d_m$, $m=1,\cdots,M$, is the probability of
choosing class $m$. To optimize the performance of a classifier, it is 
often to consider two types of manipulation:
\begin{itemize}
\item Feature Combination:
\begin{equation}\label{eq:feature_processing}
F: {\bf f} \rightarrow {\bf w}^T {\bf f} + w_0
\end{equation}
where ${\bf w}=(w_1, \cdots, w_N)^T$ is an $N$-dimensional weight vector.
\item Decision Combination:
\begin{equation}\label{eq:decision_processing}
D: ({\bf d}_1, \cdots, {\bf d}_P)^T \rightarrow \sum_{p=1}^P \alpha_p {\bf d}_p
\end{equation}
where ${\bf d}_1, \cdots, {\bf d}_P$ are $P$ decision vectors.
\end{itemize}
The decision learning problem is to find optimal feature weight vector
${\bf w}$ and/or decision weight vector ${\bf \alpha}= (\alpha_1 \cdots
\alpha_P)^T$ to optimize the classification performance.  Table
\ref{table:classifiers} lists several representative classifiers. They
are categorized based on whether a feature processing operation (along
columns) or a decision processing operation (along rows) is exploited in
the classification procedure. 

\begin{table}[!ht]
\centering
\caption{Categorization of classifiers based on whether a
feature combination operation (along columns) or a decision combination
operation (along rows) is conducted in the classification procedure.} 
\label{table:classifiers}
{\small
\begin{tabular}{|c||c|c|c|}    \hline
\multirow{2}{*}{Decision Combination}          & \multicolumn{3}{c|}{Feature Combination} \\
\cline{2-4}
                    & None           & One-Stage   & Multi-Stage  \\ \hline
None                & Decision Tree  & SVM   & MLP, SLM Tree \\ \hline
Ensemble            & Random Forest  &  & SLM Forest     \\ \hline
Boosting            & XG Boost       &   &  SLM Boost   \\ \hline
\end{tabular}}
\end{table}

\subsubsection{Classifiers Built Upon Feature Combinations}

The use of feature combinations to optimize classification performance
is examined in this section. Only SVM, SLM and MLP in Table
\ref{table:classifiers} exploit feature combinations. 

To introduce the SVM classifier, one can re-write Eq.
(\ref{eq:feature_processing}) and set it to zero:
\begin{equation}\label{eq:hyperplane}
{\bf w}^T {\bf f} - \phi = 0,
\end{equation}
where $\phi = - w_0$ is a bias.  There is a physical meaning associated
with Eq. (\ref{eq:feature_processing}). It corresponds to an $N$-dimensional
hyper-plane that partitions the $R^N$ space into two nonoverlapping half-spaces:
\begin{eqnarray}
R_+ & = & \left\{{\bf f} \mid {\bf w}^T {\bf f} - \phi \ge 0 \right\}, \label{eq:partition+} \\
R_- & = & \left\{{\bf f} \mid {\bf w}^T {\bf f} - \phi < 0 \right\}, \label{eq:partition-}
\end{eqnarray}
For a simple binary classification problem, SVM chooses parameters ${\bf
w}^T$ and $\phi$ to maximize the separation, or margin, between the two
classes.  Such a hyper-plane is called the maximum-margin hyper-plane.
The location of the hyper-plane is determined by a couple of specific
data points called support vectors. Generally, a set of hyper-planes is 
determined to partition the feature space. 

SVM and a single SLM tree belong to the same category in Table
\ref{table:classifiers}.  They use different criteria to search for the
optimal parameters. The optimization of an SLM tree consists of two steps.
The first step is to find the most discriminant 1D direction, denoted by
${\bf w}_opt$, that minimizes the weighted entropy of a decision under
an optimal split point $-w_0$ (or $\phi$). The second step is to project
$N$-dimensional feature vectors onto 1D feature vectors and then find
the optimal split (bias) in the 1D space.  A probabilistic search and an
adaptive particle swarm optimization algorithm were proposed in
\citep{fu2022subspace} and \citep{fu2022acceleration} to implement the
SLM, respectively.  Actually, the stochastic gradient descent (SGD)
algorithm can also be used to speed up the search of optimal combination
parameters. 

It is important to emphasize that the feature combination operations in
SVM and SLM only yield partition hyper-planes. They do not change the
feature vectors of input samples. One can evaluate the sign of ${\bf
w}^T {\bf f} - \phi$ to decide the location of a sample against a
specific partitioned hyper-plane. When there are multiple partitioning
planes, the information can be binary coded against each of the planes
as explained in \citep{lin2022geometrical}. Two samples belong to the
same region if they share the same binary code. However, this is not 
the case in MLP. 

The feature combination in Eq. (\ref{eq:feature_processing}) transforms
the embeddings of all samples from their original space into a new
scalar variable, denoted by $g$, at a single neuron. If there are $Q$
neurons, the embeddings of all samples evolve in the following manner:
\begin{equation}\label{eq:fevolve}
{\bf f}=(f_1, \cdots, f_N)^T \rightarrow {\bf g}=(g_1, \cdots, g_Q)^T,
\end{equation}
where
\begin{equation}
g_q=\sum_{i=1}^N w_{q,i} f_i + w{q,0}, \quad q=1, \cdots, Q.
\end{equation}
The embedding combination occurs at least at two layers (e.g., one
hidden layer and the output layer). A nonlinear activation unit is
needed after each embedding combination step to resolve the sign
confusion \citep{kuo2016understanding, lin2022geometrical}. 

\subsubsection{Classifiers Built Upon Decision Processing}

For traditional DTs, RF is the most popular bagging ensemble algorithm.
It consists of a set of tree predictors, where each tree is built based
on the values of a random vector sampled independently and with the same
distribution for all trees in the forest.  With the Strong Law of Large
Numbers, the performance of RF converges as the tree number increases.
As compared to the individual DTs, significant performance improvement
is achieved with the combination of many weak decision trees.  

Motivated by RF, SLM Forest is developed by learning a series of single
SLM tree models to enhance the predictive performance of each individual
SLM tree. SLM is a predictive model stronger than DT. Besides, the
probabilistic projection provides diversity between different SLM
models.  Following RF, SLM Forest takes the majority vote of the
individual SLM trees as the ensemble result for classification tasks,
and it adopts the mean of each SLM tree prediction as the ensemble
result for regression tasks. 

With standard DTs as weak learners, GBDT \citep{GBDT} and XGBoost
\citep{chen2015xgboost, chen2016xgboost} can deal with a large amount of
data efficiently and achieve the state-of-the-art performance in many
machine learning problems. They take the ensemble of standard DTs with
boosting, i.e. by defining an objective function and optimizing it with
learning a sequence of DTs.  By following the gradient boosting process,
SLM Boost was proposed to ensemble a sequence of SLM trees in
\citep{fu2022subspace}.  All the above discussion can be extended to the
design of the subspace learning regressor (SLR). 

\subsection{Historical Notes}\label{subsec:history}

\begin{table}[!ht]
\caption{The history on the development of various GL techniques}\label{table:history}
\centering
{\footnotesize
\begin{tabular}{l l l l}\hline
Year & Technology & Description & Ref. \\ \hline
2016 & Sign Confusion  & Identified  the sign confusion problem and explained & 
     \citep{kuo2016understanding} \\ 
     & Resolution      & the use of nonlinear activation to resolve it in CNNs & \\ 
2017 & RECOS  & Attempted to give a physical meaning to filter weights &
     \citep{kuo2017cnn} \\ 
     & Transform        & in CNNs    & \\ 
2018 & Saak  & Determined filter weights in CNNs with Saak transform &
     \citep{kuo2018data} \\
     & Transform      & which is a variant of principal component analysis \\ 
2019 & Saab & Determined filter weights in CNNs with Saab transform & 
     \citep{kuo2019interpretable} \\
     & Transform      & which is another variant of principal component analysis & \\ 
2019 & Label-Assisted & Attempted to link unsupervised representations with labels & \citep{kuo2019interpretable} \\
     & Regression     &                                                     &  \\  
2020 & PixelHop       & Proposed Successive Subspace Learning (SSL), & 
     \citep{chen2020pixelhop} \\
     &                & which deviates from CNN architecture completely &  \\ 
2020 & PointHop       & Applied Saab transform to octant representations &
     \citep{zhang2020pointhop} \\
     &                &  targeting at point cloud data & \\ 
2020 & Channel-Wise   & Proposed an effective multi-stage Saab transform & 
     \citep{chen2020pixelhop++,zhang2020pointhop++} \\
     & Saab Trans. &  applicable to PixelHop and PointHop                        & \\ 
2022 & DFT/RFT        & Proposed effective supervised feature selection methods: &
     \citep{yang2022supervised} \\
     &                & Discriminant/Relevant Feature Tests &  \\ 
2022 & SLM/SLR        & Proposed a new classifier/regressor that allows  & 
     \citep{fu2022acceleration,fu2022subspace} \\
     &                & both feature and decision combinations &  \\ 
2022 & IPHop          & Proposed an improved PixelHop system that has new & \citep{yang2022design} \\
     &                & unsupervised features and replaces LAG with DFT&  \\ \hline
\end{tabular}}
\end{table}

\begin{enumerate}
\item Early Exploration (2015-2017) \\
In the early exploration stage, the efforts focused on the understanding
of the roles played by the filter weights and the nonlinear activation
unit in a computational neuron of CNNs. The main objective was to shed
light on the superior performance from the signal processing viewpoint
since it was difficult to get engineering insights from the end-to-end
non-convex optimization problem formulation.  The necessity of nonlinear
activation was explained in \citep{kuo2016understanding} as a mechanism
to resolve the sign confusion problem in neural network training. 
The filter weights were explained as the memory of an CNN learning
system \citep{kuo2017cnn} and used to capture local, mid-level and
global intrinsic representations of objects \citep{xu2017understanding}.
Although the explanation of the role of filter weights was preliminary,
it did inspire the later work on unsupervised representation learning. 

\item FF-CNN, Saab Transforms and channel-wise Saab Transforms (2018-2019) \\
Based on the foundation laid by \citep{kuo2017cnn, xu2017understanding},
novel ways to determine filter weights of the convolutional layers in
CNNs in a feedforward manner were proposed in
\citep{kuo2019interpretable}. The design is called the feedforward CNN
(FF-CNN).  The filter weights in convolutional layers of FF-CNN are
determined by the Saab (or Saak) transforms.  The transform operations
are unsupervised since no class labels are used. The filter weights in
FC layers of FF-CNN are determined by the label-guided least-squared
regression (LAG). The main objective was to lower the training
complexity with an interpretable design. 

\item PixelHop, PixelHop++, PointHop, PointHop++, and GL Applications 
(2020-2021) \\
PixelHop \citep{chen2020pixelhop} and PixelHop++
\citep{chen2020pixelhop++} were two first-generation GL systems.  They
designs depart from the neural networks completely.  Instead of
demanding end-to-end connectivity of neurons, GL systems adopt
feedforward and ensemble learning principle.  Both PixelHop and
PixelHop++ have three modules.  They use multi-stage Saab transforms
and pooling operation to yield joint spatial-spectral representations with no
supervision in the first module.  The LAG operation is exploited to
reduce the representation dimension before being fed into classifiers in the
second stage. A traditional ML classifier is used in the third stage.
The main difference between PixelHop and PixelHop++ lies in the
multi-stage ``standard" and ``channel-wise" Saab transforms.  While
PixelHop and PixelHop++ target at the image classification application,
PointHop \citep{zhang2020pointhop} and PointHop++
\citep{zhang2020pointhop++} were developed for the point cloud
classification. Due to the irregular spatial structure of a 3D point
cloud scan, a novel octant representation was introduced before the Saab
transform in PointHop. 

We have witnessed many successful applications of GL since 2020.  Their
performance is comparable with that of state-of-the-art DL solutions
with significantly less resource.  Examples include image classification
\citep{yang2021pixelhop}, image enhancement \citep{azizi2020noise},
image quality assessment \citep{mei2022greenbiqa, zhang2020data},
deepfake image/video detection \citep{chen2022defakehop++,
chen2021defakehop, chen2021geo, zhu2022pixelhop}, point cloud
classification, segmentation, registration \citep{kadam2021r,
kadam2020unsupervised, kadam2022pcrp, liu20213d, zhang2021gsip,
zhang2020unsupervised}, face biometrics \citep{rouhsedaghat2021facehop,
rouhsedaghat2021low}, texture analysis and synthesis
\citep{lei2020nites, lei2021tghop, zhang2019texture}, graph node
classification \citep{xie2022graphhop++, xie2022graphhop}, data
compression \citep{chen2020point, zhang2020image}, joint image
compression and classification \citep{tseng2020interpretable}, 3D
medical image analysis \citep{ding2020saak, liu2021voxelhop}, etc.  We
will present several successful examples to provide insights for future
development of GL in Sec. \ref{sec:applications}. 

\item Advanced GL Tools and Broader Applications (2022- ) \\
Two advanced GL tools were developed in 2022: 1) DFT/RFT for supervised
discriminant/relevant feature selection \citep{yang2022supervised} (see
Sec. \ref{subsec:features}) and 2) SLM/SLR as a new classifier/regressor
that exploits both feature combinations and decision combinations (see
Sec. \ref{subsec:decision}). 

Furthermore, a more advanced GL system called IPHop was proposed in
\citep{yang2022design}. It replaces the traditional max-pooling
operation with the absolute-max-pooling operation and yields more
unsupervised representations by introducing the spatial Saab transform at each
channel in the first module of PixelHop.  Furthermore, it replaces the
LAG operation with DFT in the second module of PixelHop. 
\end{enumerate}

\section{Demonstrated Examples}\label{sec:applications}

GL has been successfully applied to quite a few application problems. In this
section, we choose five representative examples to illustrate the power of GL.

\subsection{Deepfake Detection}\label{subsec:deepfake}

\begin{table}[ht]
\caption{Comparison of the detection performance of several deepfake
detectors on the second-generation datasets under cross-domain training
and with AUC as the performance metric.  The AUC results of DefakeHop
and DefakeHop++ in both frame-level and video-level are given. The best
and the second-best results are shown in boldface and with underbar,
respectively. Furthermore, we include results of DefakeHop and
DefakeHop++ under the same-domain training in the last 4 rows.  The AUC
results of benchmarking methods are taken from \citep{li2020celeb} and
the number of parameters are from \url{https://keras.io/api/applications}.  
Also, $^{a}$ and  $^{b}$ denote DL and GL methods, respectively
\citep{chen2022defakehop++}.}\label{tab:compare2}
{\small
\begin{center}
\begin{tabular}{c|c|cc|c} \hline\hline
                                                &                                               & \multicolumn{2}{c|}{2nd Generation}  &   \\ \hline
Methods                                        & Model                                         & Celeb-DF v1     & Celeb-DF v2 & Param No. \\ \hline
Two-stream \citep{zhou2017two}                  & InceptionV3$^{a}$& 55.7\%    & 53.8\%            & 23.9M \\
Meso4 \citep{afchar2018mesonet}                 & Designed CNN$^{a}$                            & 53.6\%    & 54.8\%            & 28.0K \\
MesoInception4 \citep{afchar2018mesonet}        & Designed CNN$^{a}$                            & 49.6\%    & 53.6\%            & 28.6K \\
HeadPose \citep{yang2019exposing}               & SVM$^{b}$                                     & 54.8\%    & 54.6\%            & - \\
FWA \citep{li2018exposing}                      & ResNet-50$^{a}$             & 53.8\%    & 56.9\%            & 25.6M \\
VA-MLP \citep{matern2019exploiting}             & Designed CNN$^{a}$                            & 48.8\%    & 55.0\%            & - \\
VA-LogReg \citep{matern2019exploiting}          & Logistic Regression$^{b}$                     & 46.9\%    & 55.1\%            & - \\
Xception-raw \citep{rossler2019faceforensics++} & XceptionNet$^{a}$  & 38.7\%    & 48.2\%            & 22.9M \\
Xception-c23 \citep{rossler2019faceforensics++} & XceptionNet$^{a}$  & -    & 65.3\%            & 22.9M \\
Xception-c40 \citep{rossler2019faceforensics++} & XceptionNet$^{a}$  & -    & 65.5\%            & 22.9M \\
Multi-task \citep{nguyen2019multi}              & Designed CNN$^{a}$                            & 36.5\%    & 54.3\%            & - \\
Capsule \citep{nguyen2019use}                   & CapsuleNet$^{a}$     & -    & 57.5\%            & 3.9M \\
DSP-FWA \citep{li2019exposing}                  & SPPNet$^{a}$            & -    & \underbar{64.6\%}      & - \\
Multi-attentional \citep{zhao2021multi}         & Efficient-B4$^{a}$      & -    & {\bf 67.4\%}           & 19.5M \\

DefakeHop++ \citep{chen2022defakehop++} (Frame)              & DefakeHop++$^{b}$    & \underbar{56.30\%}     & 60.5\%            & 238K \\
DefakeHop++ \citep{chen2022defakehop++} (Video)              & DefakeHop++$^{b}$    & {\bf 58.15\%}     & 62.4\%            & 238K \\\hline

DefakeHop \citep{chen2021defakehop} (Trained on Celeb-DF, Frame)             & DefakeHop$^{b}$      & 93.1\%    & 87.7\% & 42.8K \\
DefakeHop \citep{chen2021defakehop} (Trained on Celeb-DF, Video)             & DefakeHop$^{b}$      & 95.0\%    & 90.6\% & 42.8K \\
DefakeHop++ \citep{chen2022defakehop++} (Trained on Celeb-DF, Frame)             & DefakeHop++$^{b}$    & \underbar{95.4\%} & \underbar{94.3\%}  & 238K \\
DefakeHop++ \citep{chen2022defakehop++} (Trained on Celeb-DF, Video)             & DefakeHop++$^{b}$    & {\bf 97.5\%}    & {\bf 96.7\%}         & 238K \\\hline
\end{tabular}
\end{center}}
\end{table}

The fast-growing Generative Adversarial Network (GAN) technology has
been applied to image forgery in recent years. They are effective in
reducing manipulation traces detectable by human eyes.  It is
challenging for humans to distinguish Deepfake images from real ones
against advanced deepfake methods.  

There are three generations of deepfake video datasets, which shows the
evolution of deepfake techniques. The UADFV dataset
\citep{li2018exposing} belongs to the first-generation. It has 50 video
clips generated by one deepfake method. Its real and fake videos can be
easily detected by human eyes.  FaceForensics++ (FF++)
\citep{rossler2019faceforensics++}, Celeb-DF-v1 and Celeb-DF-v2
\citep{li2020celeb} belong to the second generation.  They have more
video clips with more subjects. It is difficult for humans to
distinguish real and fake faces.  DFDC \citep{dolhansky2020deepfake} is
an example of the third-generation dataset. It contains more than 100K
fake videos generated by 8 deepfake techniques and perturbed by 19
distortions types.  

State-of-the-art deepfake detectors are based on DNNs. Although they
offer high detection performance, their model sizes are exceptionally large so
that they cannot be deployed on mobile phones. For example, the winning
team of the DFDC contest \citep{seferbekov2020dfdc} used seven
pre-trained EfficientNets that contain 432 million parameters. In
contrast, two lightweight deepfake detectors called DefakeHop
\citep{chen2021defakehop} and DefakeHop++ \citep{chen2022defakehop++} were
developed based on the GL principle.  Their model sizes are
significantly smaller as shown in the last column of Table
\ref{tab:compare2}. 

The detection performance of several deepfake detectors on the
second-generation datasets under cross-domain training was compared in
Table \ref{tab:compare2}, where AUC is used as the performance metric.
For the cross-domain setting, deepfake detectors were trained on FF++
and tested on Celeb-DF-v1 and Celeb-DF-v2. As shown in the table,
video-level DefakeHop++ gives the best AUC score while frame-level
DefakeHop++ gives the second best for Celeb-DF-v1. As to Celeb-DF-v2,
Multi-attentional yield the best AUC score while Xception-raw and
Xception-c23 offer the second best scores.  DefakeHop++ is slightly
inferior to them. Furthermore, the performance of DefakeHop and
DefakeHop++ under the same domain training is shown in the last four
rows of Table \ref{tab:compare2}. Their performance has improved
significantly. Video-level DefakeHop++ outperforms video-level DefakeHop
by 2.5\% in Celeb-DF v1 and 6.1\% in Celeb-DF v2. 

\begin{figure*}[!t]
\centering
\includegraphics[width=0.8\linewidth]{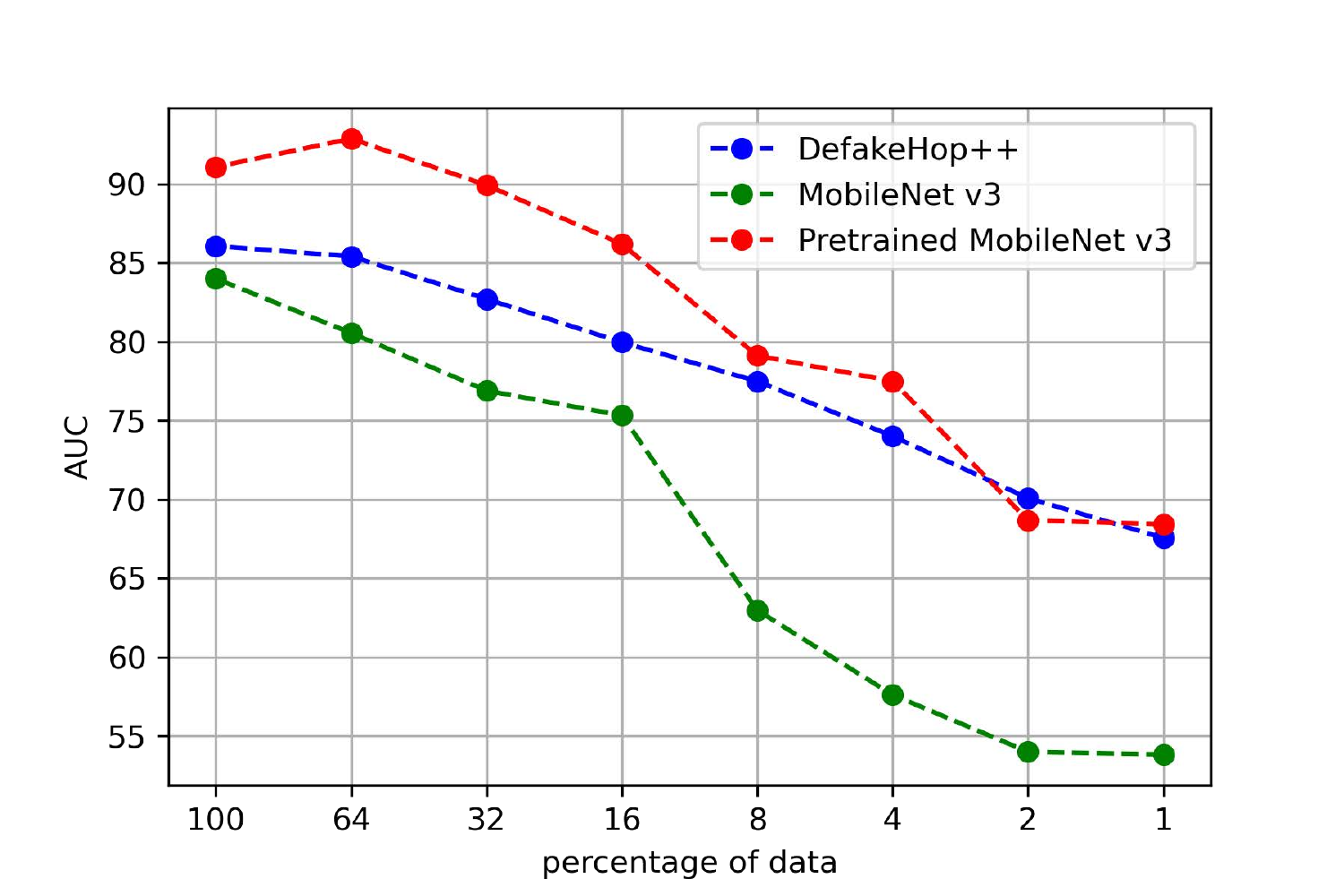}
\caption{Comparison of the detection AUC performance as a function of
training data percentages of the DFDC dataset: 1) DefakeHop++, 2) MobileNet
v3 with pre-training by ImageNet, and 3) MobileNet v3 without pre-training
\citep{chen2022defakehop++}.}\label{fig_weak}
\end{figure*}

Fig. \ref{fig_weak} compares the detection AUC performance of MobileNet
v3 and DefakeHop++ on the third-generation dataset, DFDC.  MobileNet v3
is a lightweight CNN model of 1.5M parameters targeting at mobile
applications, DefakeHop++ has an even smaller model size of 238K
parameters (i.e., 16\% of MibileNet v3).  We see from Fig.
\ref{fig_weak} that DefakeHop++ outperforms MobileNet v3 without data
augmentation and leverage of pre-trained models. DefakeHop++ and pretrained
MobileNet v3 have similar detection performance under weak supervision.

\subsection{Blind Image/Video Quality Assessment}\label{subsec:biqa}

Image/video quality assessment (QA) aims at evaluating image/video
quality to facilitate image streaming.  Based on the availability of
undistorted reference images/videos, objective QA can be classified into
three categories \citep{lin2011perceptual}: full-reference (FR),
reduced-referenced (RR) and no-reference (NR). The last one is also
known as blind QA (BQA).  FR-QA metrics have achieved high consistency
with human subjective evaluation. Many FR-QA methods have been well
developed in the last two decades such as SSIM \citep{wang2004image},
FSIM \citep{zhang2011fsim} and VMAF \citep{li2018vmaf}.  RR-QA metrics
only utilize features of reference images/videos for quality evaluation.
In some application scenarios (e.g., image receivers), users cannot
access reference images/videos so that NR-QA is the only choice. BQA
methods attract growing attention in recent years. 

The GL technology has been applied to BIQA (blind image quality
assessment) as summarized in this subsection.  Conventional BIQA methods
has two steps: 1) extraction of quality-aware features and 2) adoption
of a regression model for quality score prediction.  Recently, as the
amount of user generated images grows rapidly, DNN-based BIQA methods
have been proposed with significant performance improvement
\citep{yang2019survey}. DNN-based solutions often rely on a huge
pre-trained network which is trained by a large dataset. The large model
size demands high computational complexity and memory requirement. In
\citep{mei2022greenbiqa}, a lightweight BIQA method, called GreenBIQA,
was proposed to achieve high performance that is competitive with
DNN-based solutions yet demands much less computing power and memory.
To this end, for a video source, GreenBIQA can predict perceptual
quality scores frame by frame in real time. 

The performance of GreenBIQA is evaluated on four IQA datasets.  CSIQ
\citep{larson2010most} and KADID-10K \citep{lin2019kadid} are two
synthetic-distortion datasets. They were created by applying multiple
distortions of various levels to a set of reference images.  LIVE-C
\citep{ghadiyaram2015massive} and KonIQ-10K \citep{hosu2020koniq} are two
authentic-distortion datasets. They contain a wide range of distorted
images captured by cameras. The performance metrics are the Pearson
Linear Correlation Coefficient (PLCC) and the Spearman Rank Order
Correlation Coefficient (SROCC). PLCC is used to measure the linear
correlation between predicted scores and subjective quality scores.
SROCC is adopted to measure the monotonicity between predicted and
subjective quality scores. Each dataset is split into 80\% for training
and 20\% for testing.  Besides, 10\% of training data is used for
validation. 

The performance of GreenBIQA is compared with four conventional and five
DL-based methods in Table \ref{table:biqa}, where the experiments run 10
times and the median PLCC and SROCC values are reported.  The
benchmarking methods can be classified into four categories. 
\begin{itemize}
\item NSS-based handcrafted features: NIQE \citep{mittal2012making} and 
BRISQUE \citep{mittal2012no}
\item Codebook-learning methods: CORNIA \citep{ye2012unsupervised} and 
HOSA \citep{xu2016blind}
\item DL methods without pre-training: BIECON \citep{kim2016fully} and 
WaDIQaM \citep{bosse2017deep}
\item DL methods with pre-training: PQR \citep{zeng2018blind}, DBCNN
\citep{zhang2018blind}, and HyperIQA \citep{su2020blindly}
\end{itemize}
Methods in the first two categories are called conventional BIQA methods.

As shown in Table \ref{table:biqa}, GreenBIQA outperforms conventional
BIQA methods in all four datasets.  It also outperforms two DL methods
without pre-training in both authentic-distortion datasets.  As compared
with DL methods with pretraining, GreenBIQA achieves competitive or even
better performance in synthetic-distortion datasets (i.e., CSIQ and
KADID-10K).  For authentic-distortion datasets, GreenBIQA performs
better than DL methods without pretraining, which demonstrates that it
can generalize well to diverse distortions.  There is performance gap
between GreenBIQA and DL models with pre-training.  Yet, these
pre-trained models have much larger model sizes as a tradeoff. 

\begin{table}[!ht]
\caption{Comparison of the SROCC/PLCC performance and model sizes of various
blind image quality assessment methods on several IQA databases
\citep{mei2022greenbiqa}.}\label{table:biqa}
\centering
{\small
\begin{tabular}{ l  cc cc cc cc  c }
\toprule
\multirow{2}{*}{BIQA Method} & \multicolumn{2}{c}{CSIQ} & \multicolumn{2}{c}{LIVE-C} & \multicolumn{2}{c}{KADID-10K} & \multicolumn{2}{c }{KonIQ-10K} & \multirow{2}{*}{Model Size (MB)} \\
\cmidrule(l){2-3} \cmidrule(l){4-5} \cmidrule(l){6-7} \cmidrule(l){8-9}
& SROCC & PLCC & SROCC & PLCC & SROCC & PLCC & SROCC & PLCC \\ \midrule
NIQE \citep{mittal2012making}     &0.627 &0.712 &0.455 &0.483 &0.374 &0.428 &0.531 &0.538 &-\\
BRISQUE \citep{mittal2012no}     &0.746 &0.829 &0.608 &0.629 &0.528 &0.567 &0.665 &0.681 &-\\ \midrule
CORNIA \citep{ye2012unsupervised}&0.678 &0.776 &0.632 &0.661 &0.516 &0.558 &0.780 &0.795 &7.4\\
HOSA \citep{xu2016blind}         &0.741 &0.823 &0.661 &0.675 &0.618 &0.653 &0.805 &0.813 &0.23\\ \midrule
BIECON \citep{kim2016fully}      &0.815 &0.823 &0.595 &0.613 &-     &-     &0.618 &0.651 &35.2\\
WaDIQaM \citep{bosse2017deep}    &\textbf{0.955} &\textbf{0.973} &0.671 &0.680 &-     &-     &0.797 &0.805 &25.2\\ \midrule
PQR \citep{zeng2018blind}        &0.872 &0.901 &0.857 &0.882 &-     &-     &0.880 &0.884 &235.9\\
DBCNN \citep{zhang2018blind}     &0.946 &0.959 &0.851 &0.869 &0.851 &\textbf{0.856} &0.875 &0.884 &54.6\\
HyperIQA \citep{su2020blindly}   &0.923 &0.942 &\textbf{0.859} &\textbf{0.882} &\textbf{0.852} &0.845 &\textbf{0.906} &\textbf{0.917} &104.7\\ \midrule
GreenBIQA \citep{mei2022greenbiqa} &0.925 &0.936 &0.673 &0.689 &0.847 &0.848 &0.812 &0.834 &1.9\\ \bottomrule
\end{tabular}}
\end{table}

\begin{figure*}[!htbp]
\centering
\includegraphics[width=1.0\linewidth]{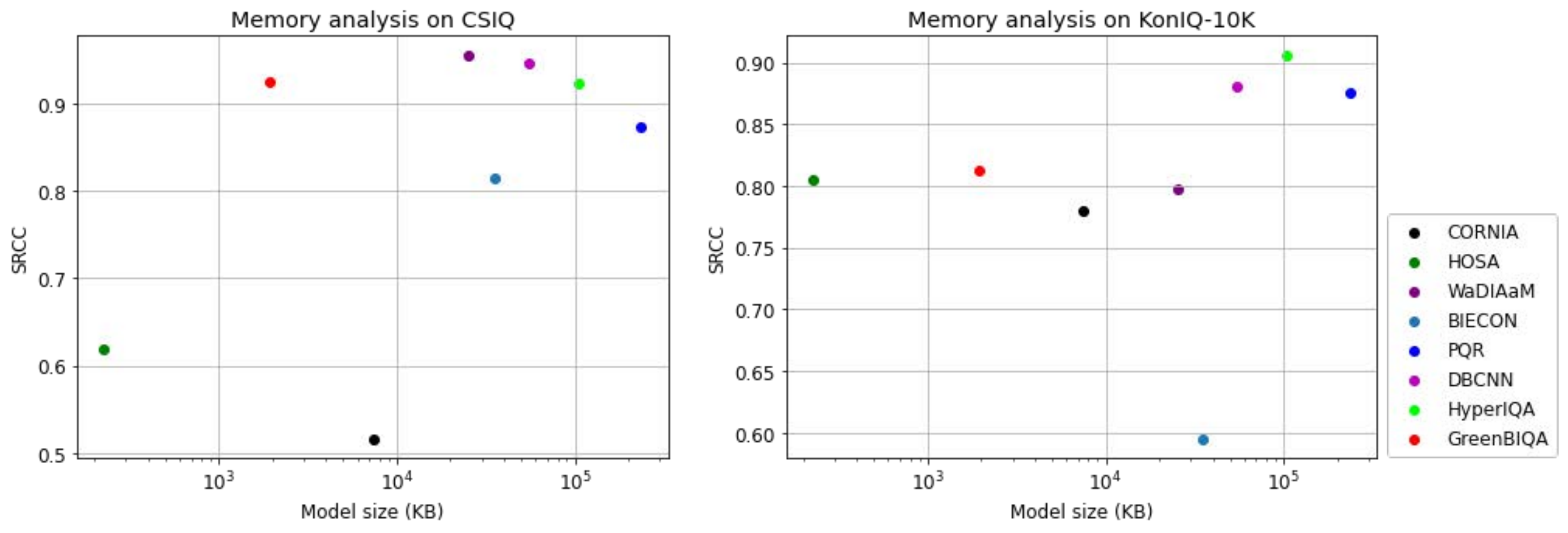}\\
(a) SRCC versus model sizes  \\
\includegraphics[width=1.0\linewidth]{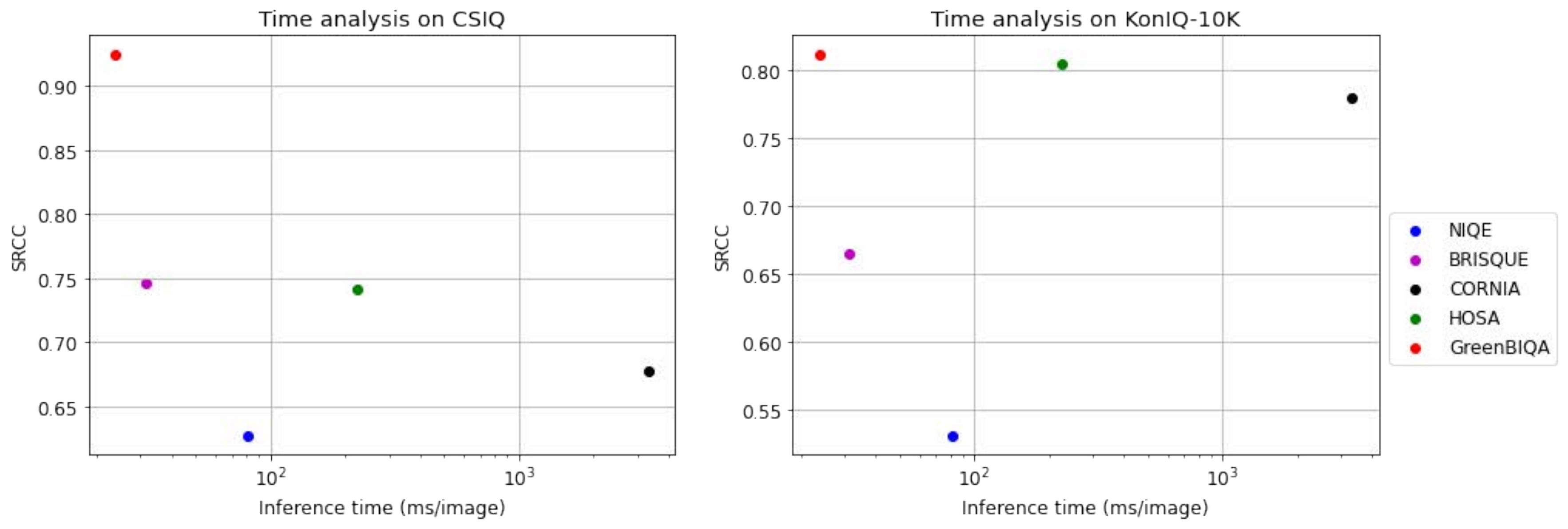}\\ 
(b) SRCC versus running time 
\caption{Comparison of (a) SRCC vs. model sizes and (b) SRCC vs. 
running time of several BIQA methods on the CSIQ and KonIQ-10K
datasets \citep{mei2022greenbiqa}.}\label{fig:analysis}
\end{figure*}

\subsection{Point Clouds Registration}\label{subsec:pc_registration}

GL technologies have been developed for point cloud classification,
segmentation and registration tasks. Point cloud registration is used as
an example to demonstrate GL techniques in this subsection.  Given a
pair of point cloud scans, registration attempts to find a rigid
transformation that optimally aligns them. The quality of registration
affects downstream tasks such as classification, segmentation, object
detection, pose estimation and odometry. They are commonly encountered
in autonomous driving, robotics, computer graphics, localization, AR/VR,
etc. Present focus of point cloud registration research is to develop
learning models that can handle challenges such as robustness to noise,
varying point densities, outliers, occlusion and partial views. 

To develop an accurate 3D correspondence solution, it is essential to
have a good representation for points. Earlier solutions, e.g.,
\citep{tombari2010unique}, \citep{johnson1997spin}, used 3D descriptors to
capture local geometric properties such as surface normals, tangents and
curvatures.  They are derived based on the first- or higher-order
statistics of neighboring points, histogram, angles, etc.  Recently, the
trend is to learn features from an end-to-end optimization setting with
DNNs. DL methods are often run on GPUs since they demand larger model
sizes and longer training/inference time.  DL-based point cloud
processing is no exception.  It is desired to look for a green solution
that demands much less power consumption. This implies a smaller model
size and less training/inference time. Yet, its performance should still
be on par with that of DNNs.  

\begin{figure*}[htbp]
\centering
\includegraphics[width=1.0\linewidth]{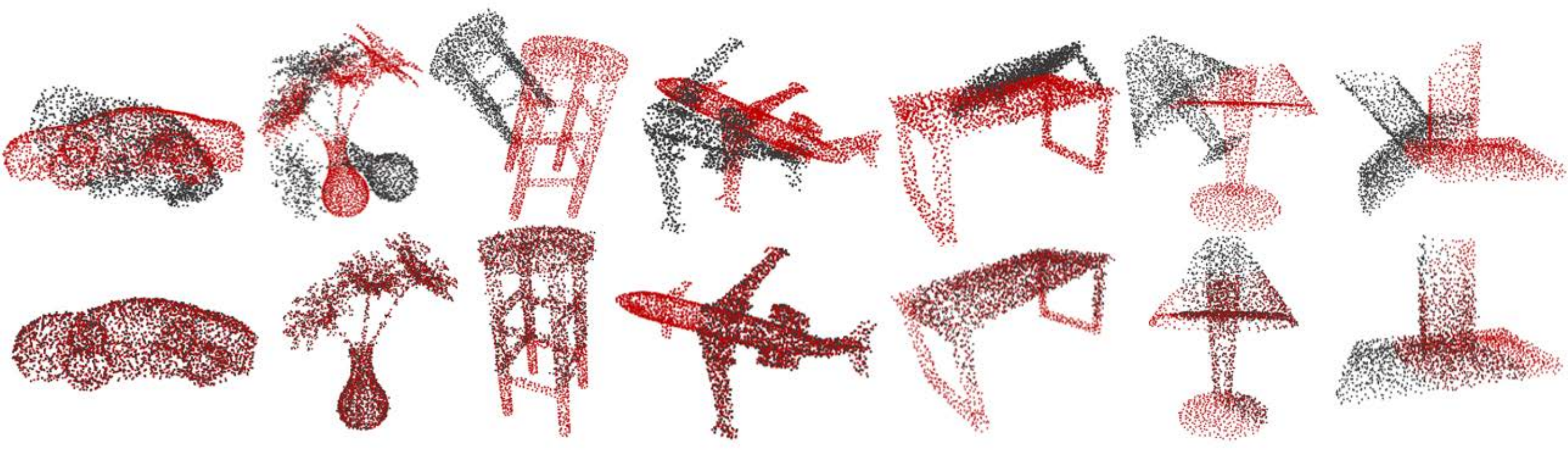}
\caption{Registration of seven point clouds from the ModelNet40 dataset
using R-PointHop. The first row shows source point clouds (in black) and
their target point clouds (in red), respectively. The second row shows
registration results.  Both the source and the target are complete point
clouds in columns \#1 and \#2.  The source in columns \#3-\#5 contains
only part of the target. Both the source and the target are partial in
columns \#6 and \#7 \citep{kadam2022r}.}\label{fig:registration}
\end{figure*}

A GL-based method, called R-PointHop, was proposed in \citep{kadam2022r}
to meet the need. It learns representations in an unsupervised manner
for point correspondence. Afterwards, the correspondences are used to
find the 3D transformation for registration.  R-PointHop provides a new
way to extract representations that are invariant to point cloud
rotation and translation.  Rotation invariance is achieved by adopting a
local reference frame (LRF) defined at each point. LRF enables
R-PointHop to find robust point correspondences for larger rotation
angles.  Seven source (in black) and target (in red) point clouds and
their registered results are illustrated in Fig. \ref{fig:registration}.
The first two columns give registration results of full point clouds.
columns 3, 4 and 5 provide results where only the source is partial.
columns 6 and 7 show results where both the source and target are
partial. 

Registration of partial point clouds occurs in practical scenarios,
where the source and the target have only a subset of points in common.
R-PointHop can handle partial-to-partial registration which is often
encountered in real world problems. A critical element in registering
partial point clouds is to find correspondences between overlapping
points. To test the capability of partial point clouds' registration,
R-PointHop is trained on the ModelNet40 dataset \citep{wu20153d} in the
reported experiment. ModelNet40 is a synthetic dataset consisting of 40
categories of CAD models of common objects such as car, chair, table,
airplane, person, etc.  In this experiment, the initial point cloud has
1,024 point, the partial point cloud has 768 points, and the overlapping
region between the source and target the source and target is random and
between 512 and 768.  The results of partial-to-partial registration are
given in Table \ref{tab:Result4} with two settings: 1) registration on
unseen point clouds and 2) registration on unseen classes.  R-PointHop
gives the best performance in both. 

In Table \ref{tab:Result4}, ICP \citep{besl1992method}, Go-ICP
\citep{yang2015go} and FGR \citep{zhou2016fast} are three model-free
methods, while PointNetLK \citep{aoki2019pointnetlk}, DCP
\citep{wang2019deep} and PR-Net \citep{wang2019prnet} are supervised
DL-based methods. The model size of R-PointHop is only 200kB as compared
with 630kB of PointNetLK and 21.3MB of DCP. The use of transformer makes
the model size of DCP significantly larger.  Although the model free
methods are most favorable in terms of model sizes and training time,
their registration performance is much worse. R-PointHop offers a good
balance when all factors are considered. 

\begin{table}[!ht]
{\footnotesize
\centering
\caption{Comparison of the registration performance for various methods on partial 
point clouds \citep{kadam2022r}.} \label{tab:Result4}
\begin{center}
\begin{tabular}{c| c c c c c c } \hline 
 &\multicolumn{6}{c}{\textbf{Registration errors on unseen objects}} \\ \hline
Method & MSE(R)  &  RMSE(R) & MAE(R) & MSE(t) & RMSE(t) & MAE(t) \\ \hline 
ICP \citep{besl1992method}  & 1134.55 & 33.68  & 25.05 & 0.0856 & 0.2930 & 0.2500  \\ 
Go-ICP \citep{yang2015go}  & 195.99 & 13.99  & 3.17 & 0.0011 & 0.0330 & 0.0120  \\ 
FGR \citep{zhou2016fast}  & 126.29 & 11.24  & 2.83 & 0.0009 & 0.0300 & 0.0080   \\ 
PointNetLK \citep{aoki2019pointnetlk}  & 280.04 & 16.74  & 7.55 & 0.0020 & 0.0450 & 0.0250  \\ 
DCP \citep{wang2019deep}   & 45.01 & 6.71  & 4.45 & 0.0007 & 0.0270 & 0.0200  \\ 
PR-Net \citep{wang2019prnet} & 10.24 & 3.12 & 1.45 & 0.0003 & 0.0160 & 0.0100  \\
R-PointHop  \citep{kadam2022r}  & \bf{2.75} & \bf{1.66} & \bf{0.35} & \bf{0.0002} & \bf{0.0149} & \bf{0.0008} \\ \hline 
 &\multicolumn{6}{c}{\textbf{Registration errors on unseen classes}} \\ \hline
ICP \citep{besl1992method} & 1217.62 & 34.89  & 25.46 & 0.0860 & 0.293 & 0.251 \\ 
Go-ICP \citep{yang2015go}  & 157.07 & 12.53  & 2.94 & 0.0009 & 0.031 & 0.010 \\ 
FGR \citep{zhou2016fast}   & 98.64 & 9.93  & 1.95 & 0.0014 & 0.038 & 0.007  \\ 
PointNetLK \citep{aoki2019pointnetlk} & 526.40 & 22.94  & 9.66 & 0.0037 & 0.061 & 0.033 \\ 
DCP \citep{wang2019deep}   & 95.43 & 9.77 & 6.95 & 0.0010 & 0.034 & 0.025 \\ 
PR-Net \citep{wang2019prnet} & 15.62 & 3.95 & 1.71 & 0.0003 & 0.017 & 0.011  \\
R-PointHop  \citep{kadam2022r} & \bf{2.53} & \bf{1.59} & \bf{0.37} & \bf{0.0002} & \bf{0.0148} & \bf{0.0008} \\ \hline
\end{tabular}
\end{center}}
\end{table}

\subsection{Graph Node Classification}\label{subsec:graph}

Semi-supervised learning that exploits labeled and unlabeled data is
highly in demand in real-world applications because of the expensive
data labeling cost and the availability of a large amount of unlabeled
samples. For graph problems with few labels, the geometric or manifold
structure of unlabeled data can be leveraged to improve the performance
of classification, regression, and clustering algorithms.  Graph
convolutional networks (GCNs) have been widely accepted as the
\textit{de facto} tool in addressing semi-supervised graph learning
problems \citep{kipf2016semi, hamilton2017inductive, song2022graph}. GCN
parameters are learned via label supervision through backpropagation
\citep{rumelhart1986learning}.  The joint attributes encoding and
propagation as smoothening regularization enable GCNs to yield superior
performance. 

There are however challenges in GCN-based solutions.  First, GCNs demand
a sufficient number of labeled samples for training.  Instead of
deriving label embeddings from graph regularization as done in
\citep{zhu2003semi, zhou2003learning}, GCNs learns a series of
projections from the embedding space to the label space. The
transformations rely on ample labeled samples for supervision.  Lack of
sufficient labeled samples often makes training unstable (or even
divergent). To address this problem, one may integrate other
semi-supervised learning techniques (e.g., self- and co-training
\citep{li2018deeper}) with GCN training or enhance the filter power to
strengthen regularization \citep{li2019label}. Second, GCNs usually
consist of two convolutional layers so that the information from a small
neighborhood of each node is exploited \citep{kipf2016semi,
velickovic2018graph, hamilton2017inductive, abu2019mixhop}. The learning
ability is handicapped by ignoring correlations from nodes of longer
distance. Although increasing the number of layers could be a remedy, it
leads to an oversmoothing problem and results in inseparable node
representations \citep{li2018deeper, zeng2021decoupling}.  Furthermore, a
deeper model needs to train more parameters, which in turn requires even
more labeled samples. 

Two enhanced label propagation methods, called GraphHop
\citep{xie2022graphhop} and GraphHop++ \citep{xie2022graphhop++}, was
recently proposed. They are GL solutions to graph learning problems.
They achieves state-of-the-art performance as compared with GCN-based
methods \citep{kipf2016semi, li2018deeper, velickovic2018graph} and other
classical propagation-based algorithms \citep{zhou2003learning,
wang2007label, nie2010general}. They perform particularly well at
extremely low label rates.  To give an example, the performance of
GraphHop and GraphHop++ on the Coauthor CS graph dataset is shown in
Table \ref{tab:results_cs}. Each column in the table presents the
classification accuracy (\%) of all benchmarking methods on test data
under a specific label rate. 

\begin{figure*}[!t]
{\centering
\includegraphics[width=0.3\linewidth]{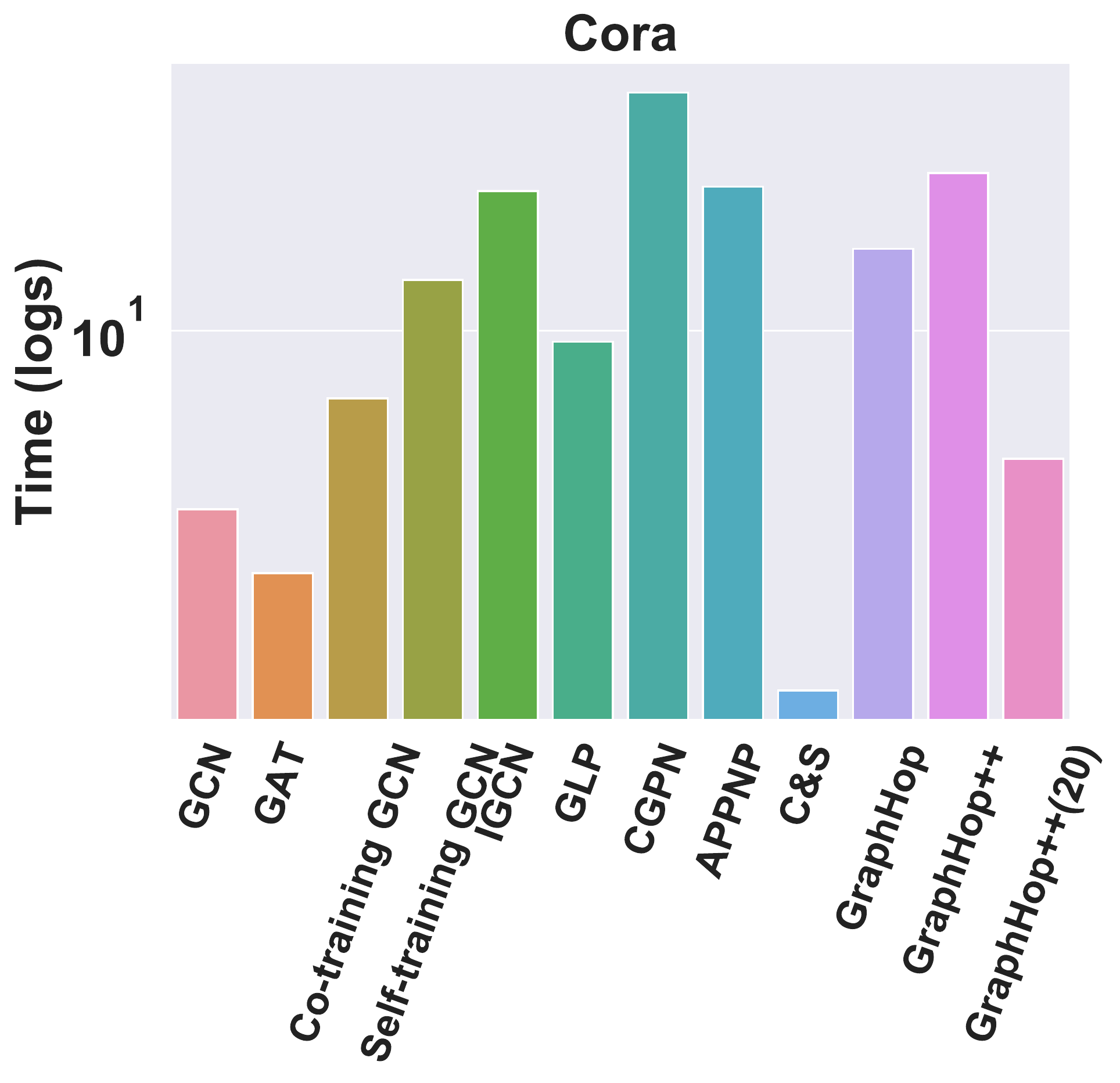}
\includegraphics[width=0.3\linewidth]{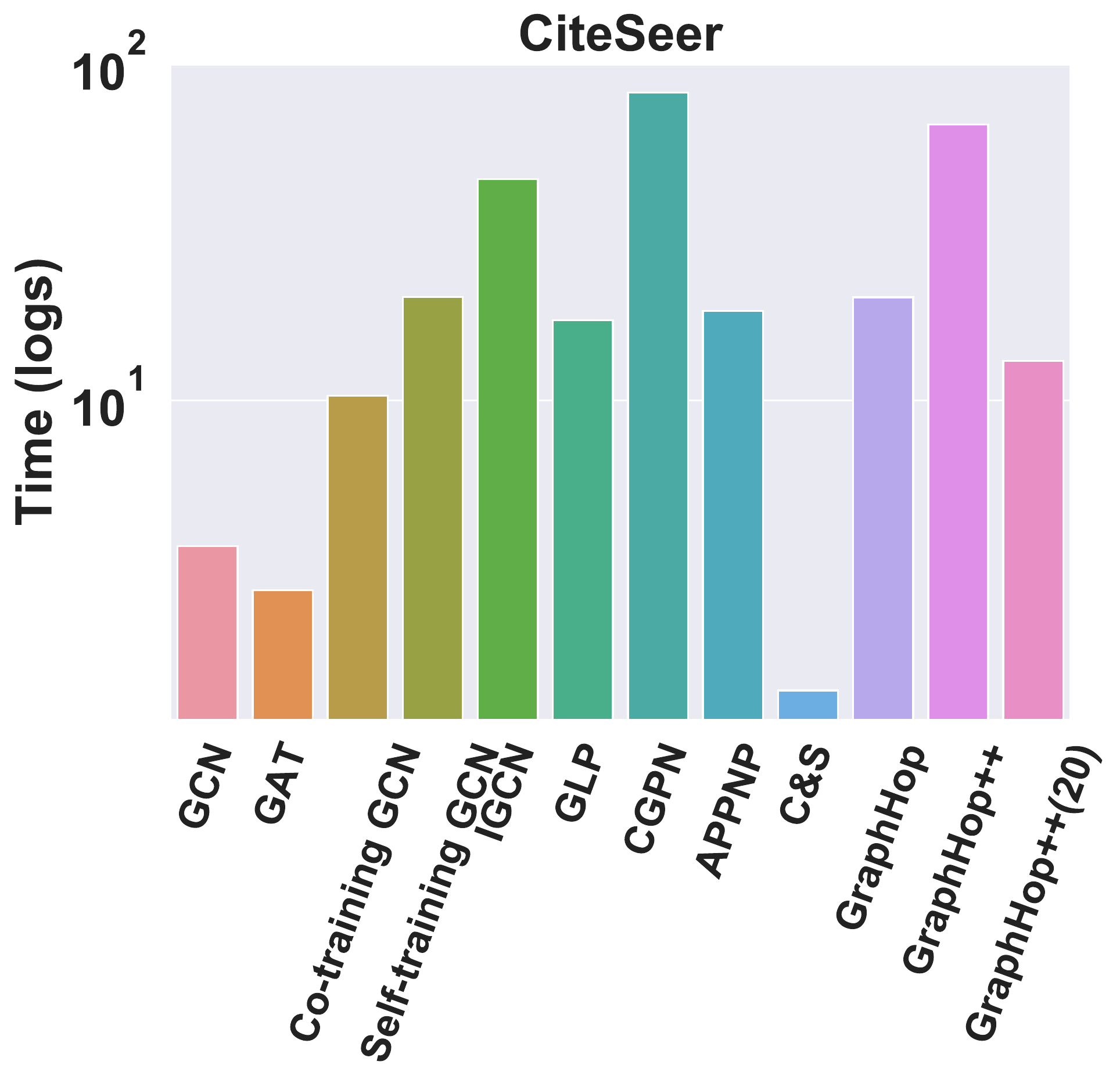}
\includegraphics[width=0.3\linewidth]{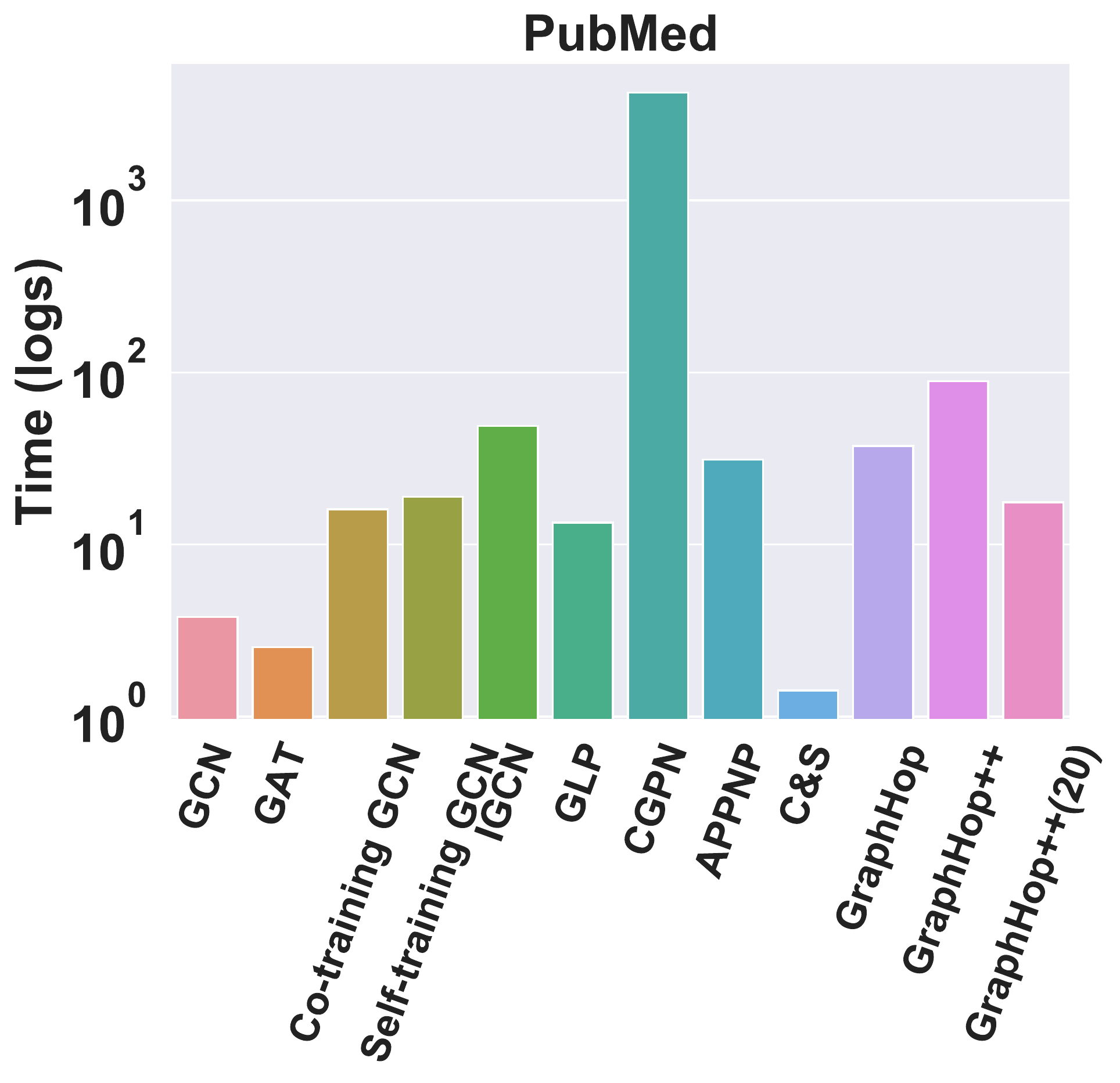}} \\
{\centering
\includegraphics[width=0.3\linewidth]{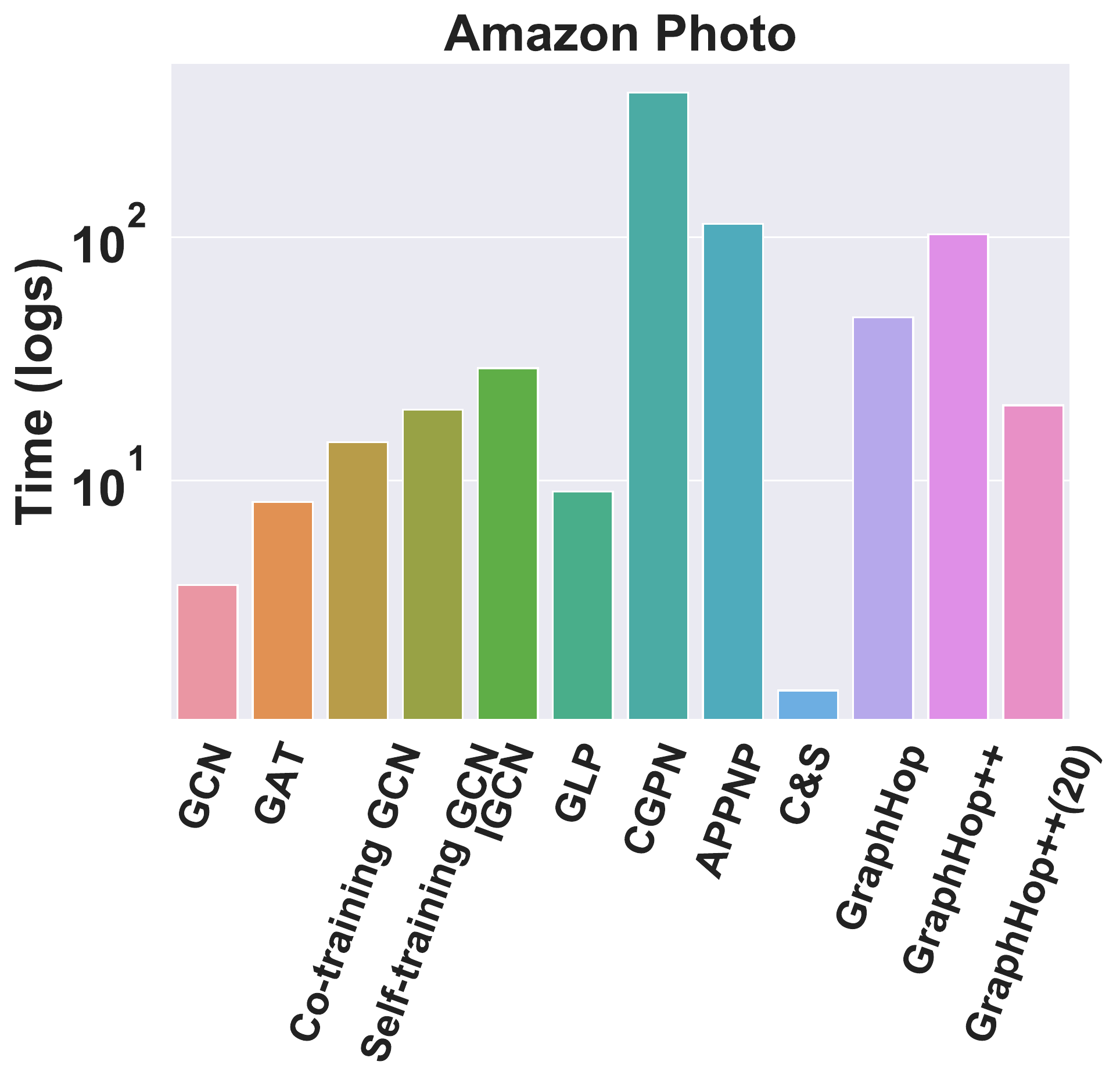}
\includegraphics[width=0.3\linewidth]{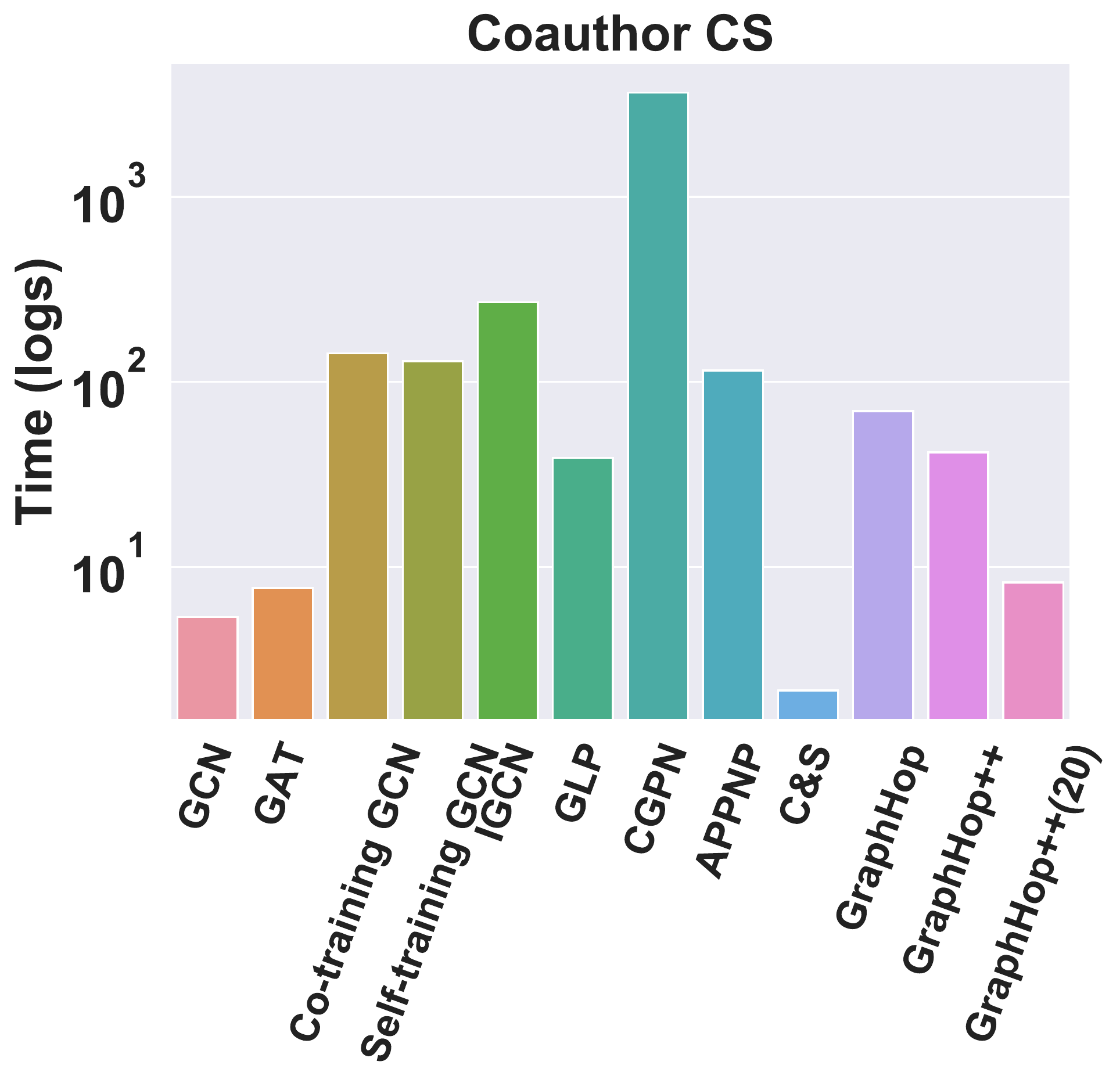}}
\caption{Comparison of computational efficiency of different methods
measured by log(second) for Cora, CiteSeer dataset, PubMed, Amazon Photo
and Coauthor C\&S datasets, where the label rate is 20 labeled samples per
class and GraphHop++(20) is the result of GraphHop++ with 20 alternate
optimization rounds.} \label{fig:time_model}
\end{figure*}

\begin{figure*}[!b]
{\centering
\includegraphics[width=0.3\linewidth]{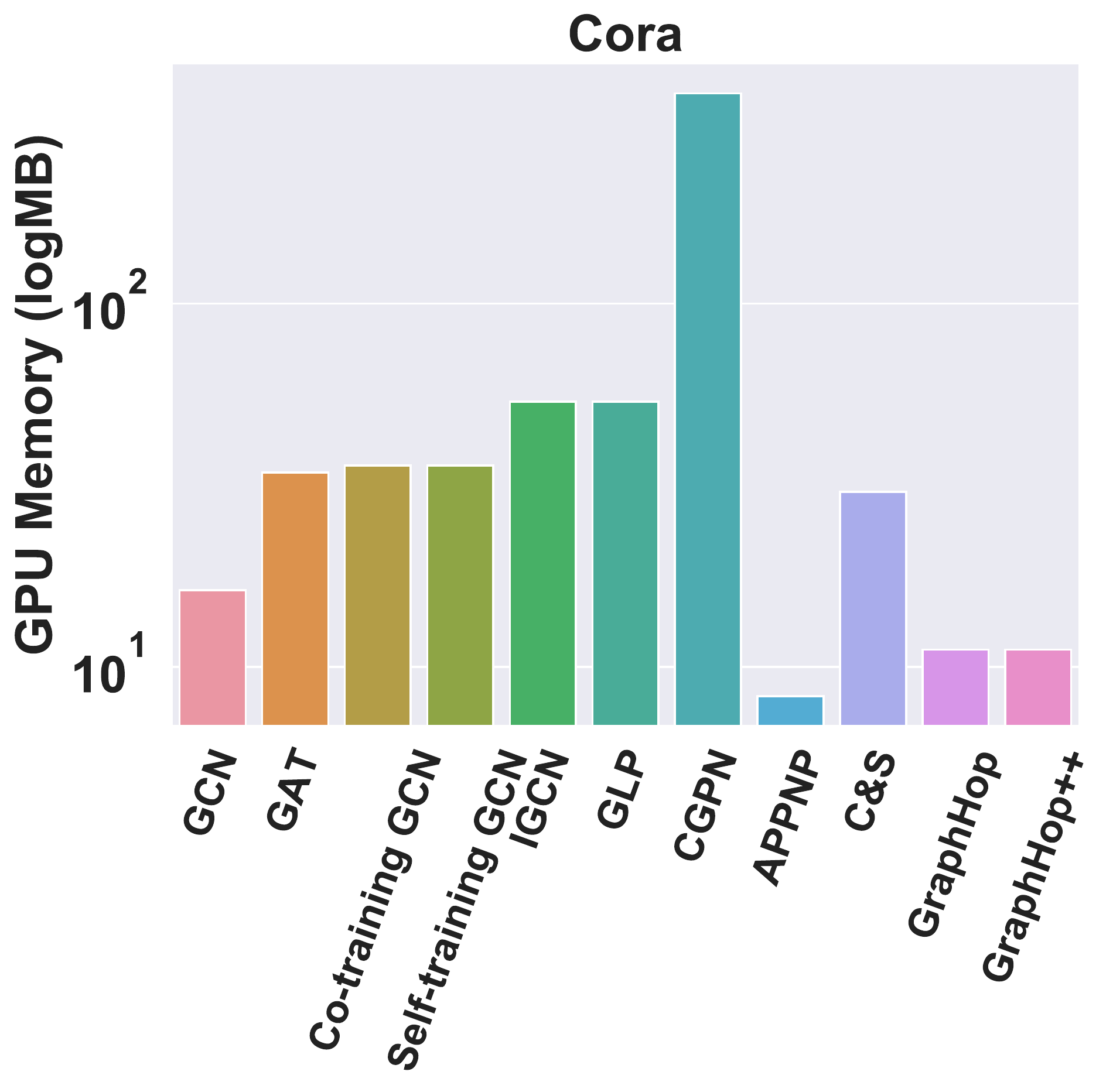}
\includegraphics[width=0.3\linewidth]{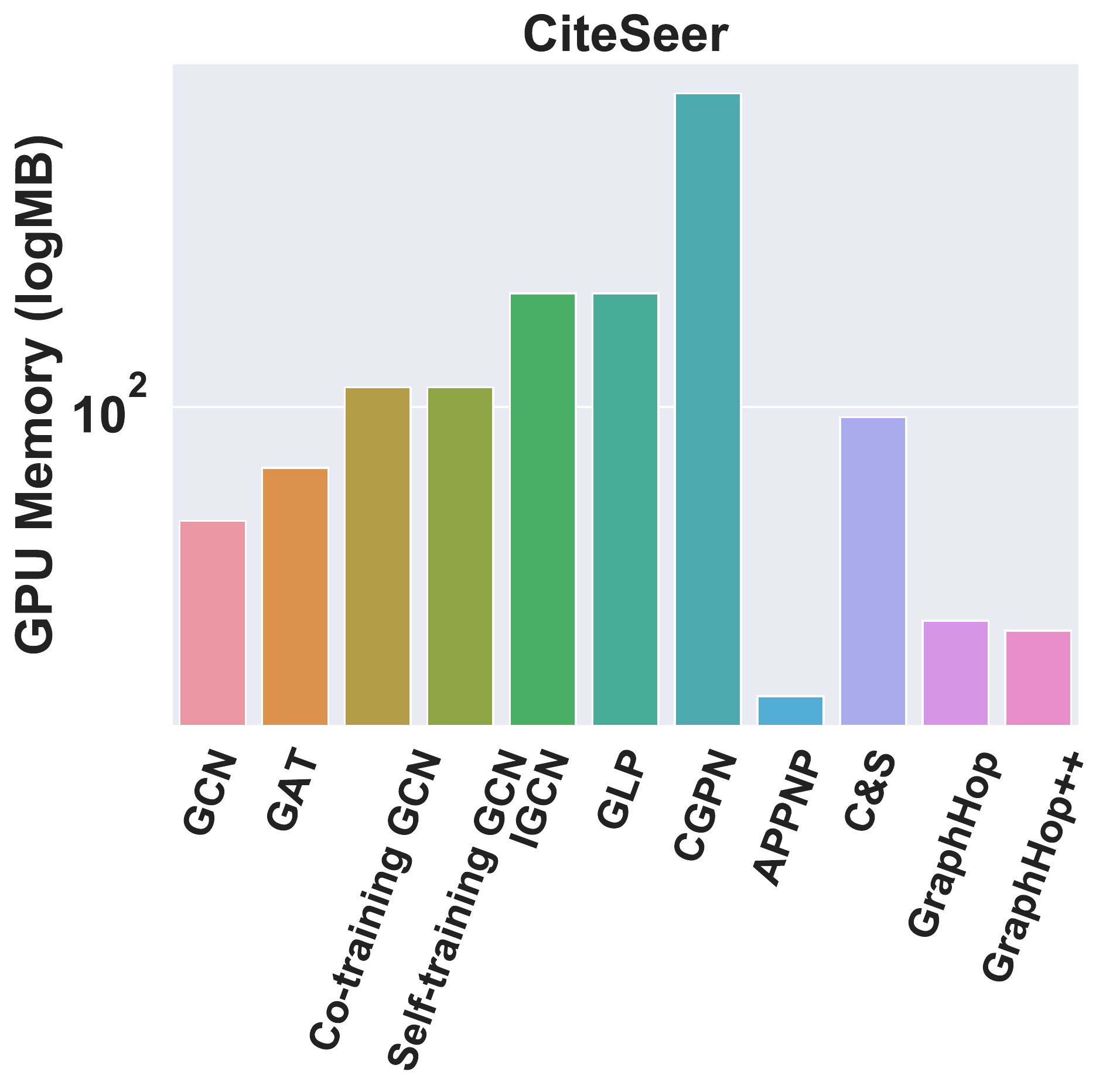}
\includegraphics[width=0.3\linewidth]{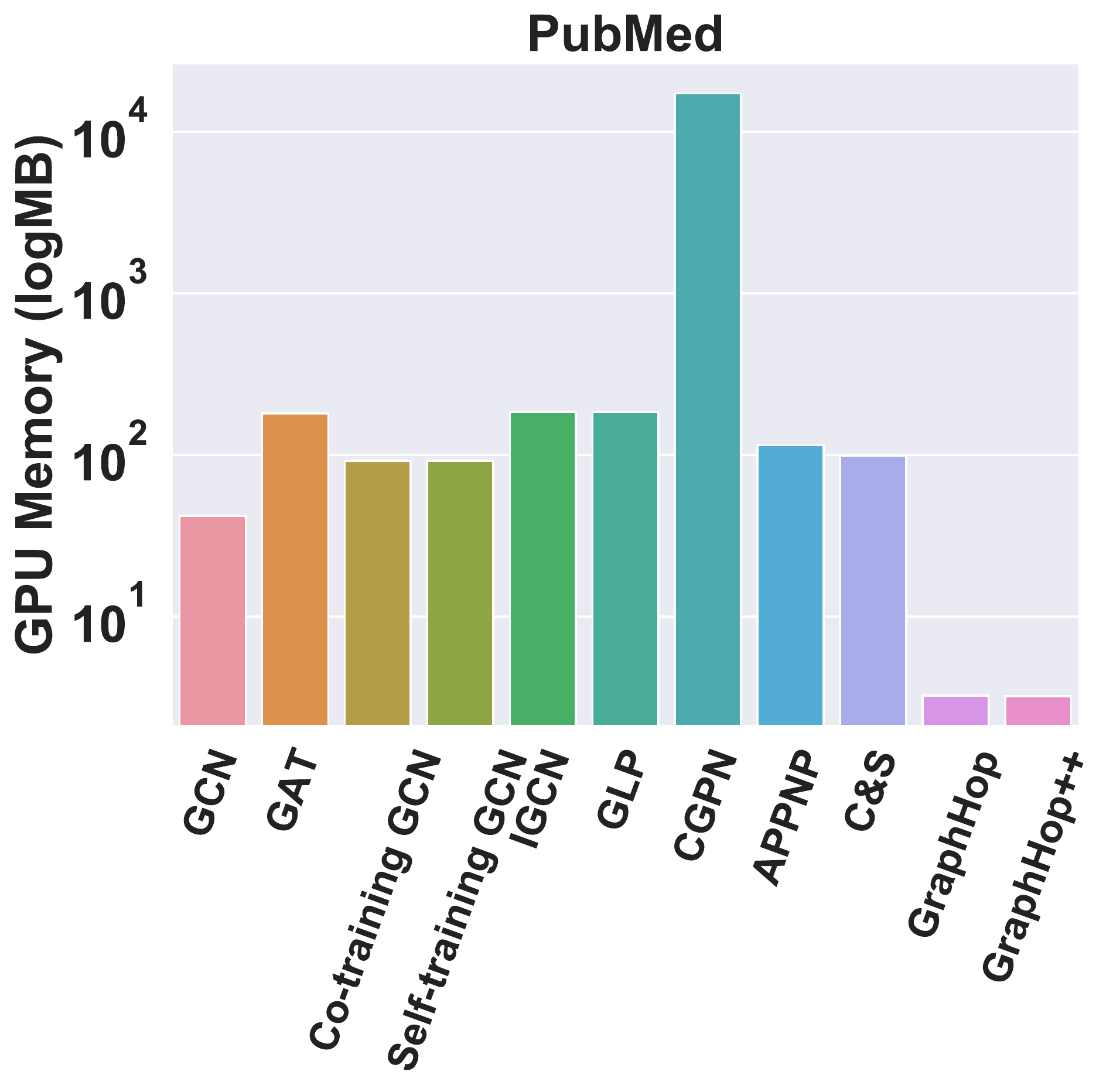}} \\
{\centering 
\includegraphics[width=0.3\linewidth]{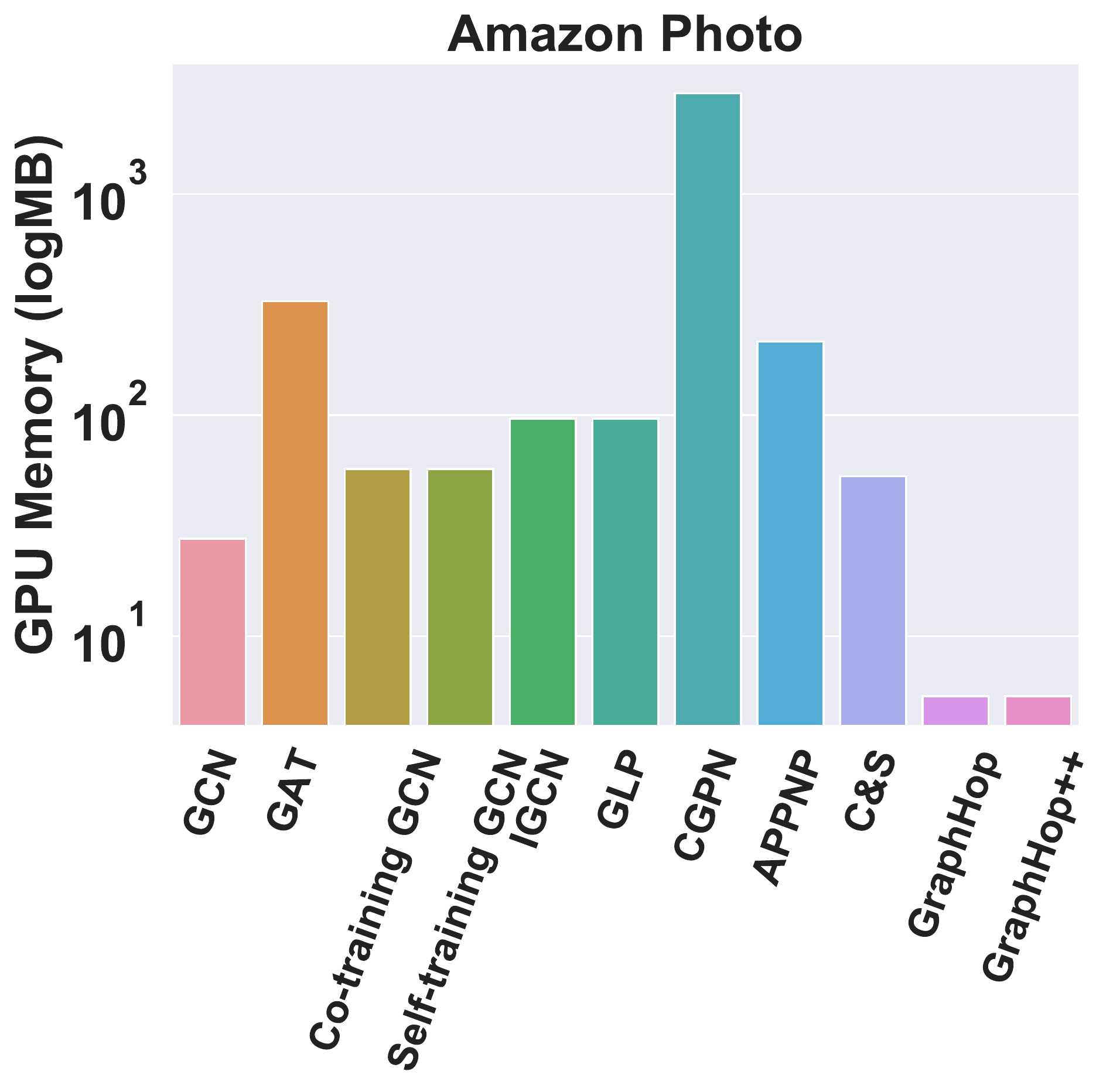}
\includegraphics[width=0.3\linewidth]{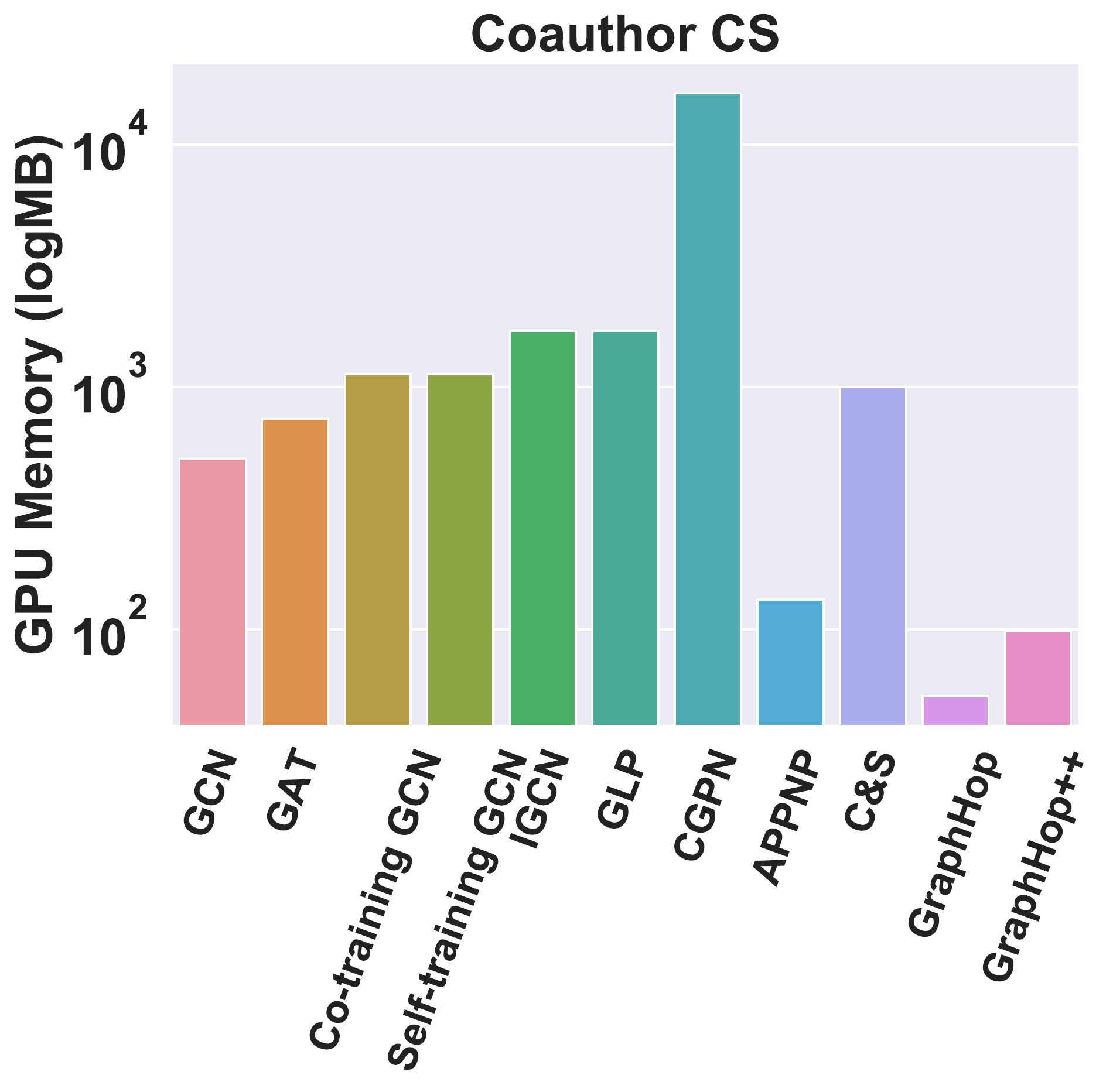}}
\caption{Comparison of GPU memory usages of different methods
measured by log(Mega Bytes) for Cora, CiteSeer dataset, PubMed, Amazon
Photo and Coauthor C\&S datasets, where the label rate is 20 labeled
samples per class.} \label{fig:memory_model}
\end{figure*}

Overall, GraphHop++ has the top performance in most cases.  In
particular, for cases with extremely limited labels, GraphHop++
outperforms other methods by a large margin.  This is because GCNs are
difficult to train with a small number of labels. The lack of
supervision prevents them from learning the transformation from the
embedding space to the label space with nonlinear activation at each
layer.  Instead, GraphHop++ applies regularization to the label space.
It is more effective since it relieves the burden of learning the
transformation. A small number of labeled nodes also restricts the
efficacy of message passing on graphs of GCN-based methods. 

\begin{table}[!ht]
\centering
\caption{Test accuracy for the Coauthor CS dataset with extremely low
label rates measured by ``the mean accuracy ($\%$) $\pm$ standard
deviation". The highest mean accuracy is in \textbf{bold} while the
second and third ones are underlined \citep{xie2022graphhop++}.}\label{tab:results_cs}
{\footnotesize
\centering
\resizebox{\textwidth}{!}
{\begin{tabular}{c|cccccc} \hline
    \multicolumn{7}{c}{\textbf{Coauthor CS}} \\ \hline
    \# of labels per class & 1 & 2 & 4 & 8 & 16 & 20 \\ \hline \hline
    GCN \citep{kipf2016semi} & $64.42\pm2.14$ & $74.07\pm1.09$ & $82.62\pm0.74$ & $87.88\pm0.55$ & \underline{$89.86\pm0.15$} & \underline{$90.19\pm0.17$} \\
    GAT \citep{velickovic2018graph} & \underline{$72.81\pm2.01$} & $80.76\pm1.25$ & $85.37\pm0.94$ & $88.06\pm0.61$ & $89.83\pm0.16$ & $89.82\pm0.17$ \\
    LP \citep{zhou2003learning} & $52.63\pm0.00$ & $59.34\pm0.00$ & $61.77\pm0.00$ & $68.60\pm0.00$ & $73.50\pm0.00$ & $74.65\pm0.00$ \\
    APPNP \citep{klicpera2018predict} & $71.06\pm17.48$ & \underline{$86.18\pm0.56$} & ${86.57\pm0.58}$ & \underline{$89.66\pm0.30$} & \underline{$90.65\pm0.21$} & \underline{$90.55\pm0.17$} \\
    C\&S \citep{huang2020combining} & $35.63\pm13.20$ & $67.11\pm3.50$ & $78.57\pm1.22$ & $83.92\pm1.15$ & $87.09\pm0.95$ & $87.97\pm0.56$ \\
    Co-training GCN \citep{li2018deeper} & \underline{$75.78\pm1.00$} & $\mathbf{86.94\pm0.70}$ & \underline{$87.40\pm0.78$} & $88.81\pm0.32$ & $89.22\pm0.42$ & $89.01\pm0.55$ \\
    Self-training GCN \citep{li2018deeper} & $69.69\pm3.27$ & $82.79\pm4.10$ & \underline{$87.62\pm1.49$} & \underline{$88.82\pm1.00$} & $89.53\pm0.54$ & $89.07\pm0.70$ \\
    IGCN \citep{li2019label} & $62.16\pm2.81$ & $59.56\pm2.91$ & $65.82\pm4.86$ & $86.57\pm0.78$ & $87.76\pm0.74$ & $88.10\pm0.52$ \\
    GLP \citep{li2019label} & $43.56\pm7.06$ & $50.74\pm7.55$ & $46.61\pm9.85$ & $76.61\pm3.39$ & $81.75\pm2.81$ & $82.43\pm3.31$ \\
    CGPN \citep{wan2021contrastive} & $67.66\pm0.00$ & $64.49\pm0.00$ & $71.00\pm0.00$ & $77.09\pm0.00$ & $78.75\pm0.00$ & $79.71\pm0.00$ \\
    GraphHop \citep{xie2022graphhop} & $65.03\pm0.01$ & $77.59\pm3.17$ & $83.79\pm1.13$ & $86.69\pm0.62$ & $89.47\pm0.38$ & $89.84\pm0.00$ \\
    GraphHop++ \citep{xie2022graphhop++} & $\mathbf{82.46\pm1.28}$ & \underline{$86.37\pm0.37$} & $\mathbf{88.45\pm0.35}$ & $\mathbf{89.87\pm0.48}$ & $\mathbf{90.69\pm0.13}$ & $\mathbf{90.87\pm0.06}$ \\
    \hline
\end{tabular}}}
\end{table}

In general, only two convolutional layers are adopted by GCN methods. It
means that only messages in the two-hop neighborhood of each labeled
node can be supervised.  Apparently, the two-hop neighborhood of a
limited number of labeled nodes cannot cover the whole network
effectively. Thus, a large number of nodes do not have supervised
training from labels. It explains the inferior performance of all GCN
methods.  Some label-efficient GCN methods try to alleviate this problem
by exploiting pseudo-labels as supervision (e.g., self-training GCN) or
improving message passing capability (e.g., IGCN and CGPN).  Yet, they
are still handicapped by deficient message passing in the graph
convolutional layers.  In contrast, label propagation methods (e.g.,
APPNP, GraphHop and GraphHop++) achieve better performance in most
datasets at low label rates since the messages from labeled nodes can
pass over a longer distance on graphs through an iterative process. It
is worthwhile to comment that, although classical LP and C\&S are also
propagation-based, their performance is poorer for the following
reasons. Classical LP fails to encode the rich node attribute
information in model learning.  C\&S was originally designed for
supervised learning and lack of labeled samples degrades its performance
significantly. 

The running time performance of GraphHop++ and its several benchmarking
methods is compared in Fig.  \ref{fig:time_model}.  It shows the results
at a label rate of 20 labeled samples per class with respect to five
datasets. For GraphHop++, results are reported with 100 rounds and 20
rounds, respectively, since it can converge within 20 rounds.  As shown
in Fig. \ref{fig:time_model}, GraphHop++(20) can achieve an average
running time. The most time-consuming part of GraphHop++ is the training
of the LR classifiers. It demands multiple loops of all data samples.
However, we find that incorporating classifiers is essential to the
superior performance of GraphHop++ under limited label rates. In other
words, GraphHop++ trades time for effectiveness in the case of extremely
low label rates. Among all datasets, GraphHop++ achieves the lowest
running time for Coauthor C\&S due to its uncomplicated design as LP. 

Next, the GPU memory usage of GraphHop++ and its several benchmarking
methods is compared in Fig. \ref{fig:memory_model}, where the values of
GraphHop++ are taken from the same experiment of running time comparison
as given in Fig.  \ref{fig:time_model}.  It is measured by
\textsf{{torch.cuda.max\_memory\_allocated()}} for PyTorch and
\textsf{{tf.contrib.memory\_stats.{MaxBytesInUse}()}} for TensorFlow.
Generally, GraphHop and GraphHop++ achieves the lowest GPU memory usage
among all benchmarking methods.  This is because GraphHop and GraphHop++
allow minibatch training. The only parameters to be stored in the GPU
are classifier parameters and one minibatch of data. In contrast, since
embeddings from different layers need to be stored for backpropagation,
GCN methods cannot simply conduct minibatch training.  Their GPU memory
consumption increases significantly. 

\subsection{Knowledge Graph Completion}\label{subsec:kgc}

Knowledge graph completion (KGC) aims to discover missing relationships
between entities in knowledge graphs (KGs).  Knowledge graph embedding
(KGE) techniques offer state-of-the-art solutions to this problem.  In
KGE models, embeddings for entities and relations are determined by
optimizing triple score functions. That is, scores among positive
triples are maximized while those among negative triples are minimized.
The entity and relation embeddings are stored as model parameters. The
number of parameters in a KGE model is linearly proportional to the
embedding dimension and the number of entities and relations in KGs,
i.e. $O((|E| + |R|)d)$, where $|E|$ is the number of entities, $|R|$ is
the number of relations, and $d$ is the embedding dimension. 

Most prior KGC methods focus on learning representations for entities
and relations.  A higher-dimensional embedding space is usually required
for better reasoning capability, which leads to a larger model size.  It
hinders the applicability of KGE methods to real-world problems (e.g.,
large-scale KGs or mobile/edge computing). It is critical to obtain a
KGC method that has good reasoning capability in low dimensions.  A
lightweight modularized KGC solution, called GreenKGC, was proposed in
\citep{wang2022greenkgc} to meet the requirement. 

GreenKGC consists of three modules: 1) representation learning, 2)
feature pruning, and 3) decision learning.  Each module is optimized by
itself. In Module 1, a KGE method, called the baseline method, is
leveraged to learn high-dimensional entity and relation representations.
In Module 2, a feature pruning process is applied to the
high-dimensional entity and relation representations to yield
discriminant low-dimensional features for triples. Since logical
patterns for different relations vary a lot, triples are partitioned
into multiple relation groups of homogeneous logical patterns.  As a
result, relations in the same group can share the same low-dimensional
features.  It helps improve the performance and reduces the model size.
In Module 3, a binary classifier is trained for each relation group so
that it predicts triple's score in inference. The predicted score is a
soft value between 0 and 1, indicating the likelihood of a certain
relation. 

We examine the performance of GreenKGC on three link prediction
datasets.  They are FB15k-237 \citep{bordes2013translating,
toutanova2015observed}, WN18RR \citep{bordes2013translating,
dettmers2018convolutional} and YAGO3-10
\citep{dettmers2018convolutional}. FB15k-237 is a subset of Freebase
\citep{bollacker2008freebase} on real-world relationships.  WN18RR is a
subset of WordNet \citep{miller1995wordnet} on lexical relationships
between word senses. YAGO3-10 is a subset of YAGO3
\citep{mahdisoltani2014yago3} on person's attributes.  For link
prediction, the goal is to predict the missing entity given a query
triple, i.e. $(h, r, ?)$ or $(?, r, t)$. The correct entity should be
ranked higher than other candidates. To this end, several common ranking
metrics are MRR (Mean Reciprocal Rank) and Hits@k (k=1, 3, 10).
Following the convention in \citep{bordes2013translating}, we adopt the
filtered setting, where all entities serve as candidates except for the
ones that have been seen in training, validation, or testing sets. 

One novel characteristic of GreenKGC is that it prunes a
high-dimensional KGE to low-dimensional triple features and make
prediction with a binary classifier as a powerful decoder.  The
capability of GreenKGC in saving the number of parameters and
maintaining the performance by pruning original 500D KGE to 100D triple
features is shown in Table \ref{tab:main}.  As shown in the table,
GreenKGC can achieve comparable or even better performance with around 4
times smaller model size.  Especially, its Hits@1 metric is improved in
most datasets except for CoDEx-L.  Furthermore, negative sampling
adopted in classifier training can correct some failed cases in original
KGE methods.  For all datasets, 100D GreenKGC could generate better
results than 100D KGE models. 

The performance of GreenKGC with KGE, classification-based, and
low-dimensional KGC methods of 32D is compared in Table \ref{tab:low}.
KGE methods, such as TransE and RotatE, cannot yield good performance in
low dimensions due to over-simplified score functions.
Classification-based methods outperform KGE methods as they adopt DNNs
as complex decoders. Low-dimensional KGC methods are specifically
designed for low dimensions. They provide state-of-the-art KGC solutions
in low dimensions. Yet, GreenKGC outperforms all of them in FB15k-237
and YAGO3-10.  For WN18RR, KGE methods perform poorly.  Since GreenKGC
relies on KGE as its baseline, its performance is affected. In spite of
this fact, GreenKGC still outperforms all KGE and classification-based
methods for WN18RR. 

\begin{table}[!ht]
\centering
{\footnotesize
\caption{Performance comparison on the link prediction task, where the
performance gain (or loss) in terms of percentages with an up (or down)
arrow and the ratio of the model size  are shown within the parentheses against
those of respective 500D models \citep{wang2022greenkgc}.} \label{tab:main}
\begin{tabular}{lc  ccc  ccc  ccc} \toprule
&& \multicolumn{3}{c}{\textbf{FB15k-237}} & \multicolumn{3}{c}{\textbf{WN18RR}} &
\multicolumn{3}{c}{\textbf{YAGO3-10}} \\
\midrule
Baseline & Dim. & MRR & H@1 & \#P (M) & MRR & H@1 & \#P(M) & MRR & H@1 & \#P (M) \\
\midrule \midrule
\multirow{5}{*}{TransE} &
500
& 0.325 & 0.228 & 7.40 & 0.223 & 0.013 & 20.50
& 0.416 & 0.271 & 61.60 \\
\cmidrule(lr){2-11}
&\multirow{2}{*}{100} 
& 0.274 & 0.186 & 1.48 & 0.200 & 0.009 & 4.10
& 0.377 & 0.269 & 12.32 \\
&& \textcolor{red}{$\downarrow$} 15.7\% & \textcolor{red}{$\downarrow$} 18.5\% & (0.2x) 
& \textcolor{red}{$\downarrow$} 10.3\% & \textcolor{red}{$\downarrow$} 30.8\% & (0.2x) 
& \textcolor{red}{$\downarrow$} 2.0\% & \textcolor{Green}{$\uparrow$} 0.7\% & (0.21x) \\
\cmidrule(lr){2-11}
& GreenKGC
& 0.316 & 0.232 & 2.04 & 0.407 & 0.361 & 4.30 
& 0.455 & 0.358 & 12.88 \\
& 100
& \textcolor{red}{$\downarrow$} 2.8\% & \textcolor{Green}{$\uparrow$} 1.8\% & (0.28x)
& \textcolor{Green}{$\uparrow$} 82.5\% & \textcolor{Green}{$\uparrow$} 176.9\% & (0.21x) 
& \textcolor{Green}{$\uparrow$} 9.4\% & \textcolor{Green}{$\uparrow$} 32.1\% & (0.21x) \\
\midrule \midrule
\multirow{5}{*}{RotatE} &
500
& 0.333 & 0.237 & 14.66 & 0.475 & 0.427 & 40.95
& 0.495 & 0.402 & 123.20 \\
\cmidrule(lr){2-11}
&\multirow{2}{*}{100} 
& 0.296 & 0.207 & 2.93 & 0.452 & 0.427 & 8.19
& 0.482 & 0.393 & 24.64 \\
&& \textcolor{red}{$\downarrow$} 11.1\% & \textcolor{red}{$\downarrow$} 12.7\% & (0.2x)
& \textcolor{red}{$\downarrow$} 3.8\% & - & (0.2x)
& \textcolor{red}{$\downarrow$} 2.0\% & \textcolor{Green}{$\uparrow$} 0.7\% & (0.21x) \\
\cmidrule(lr){2-11}
& GreenKGC
& 0.348 & 0.266 & 3.49 & 0.476 & 0.430 & 8.75 
& 0.485 & 0.405 & 25.20 \\
& 100
& \textcolor{Green}{$\uparrow$} 4.5\% & \textcolor{Green}{$\uparrow$} 12.2\% & (0.24x)
& \textcolor{Green}{$\uparrow$} 0.2\% & \textcolor{Green}{$\uparrow$} 0.7\% & (0.21x) 
& \textcolor{red}{$\downarrow$} 2.0\% & \textcolor{Green}{$\uparrow$} 0.7\% & (0.21x) \\
\bottomrule
\end{tabular}}
\end{table}

\begin{table}[!ht]
\centering
\caption{Performance comparison on link prediction in low dimensions
($d=32$), where the best and the second-best numbers are in bold and
with an underbar, respectively \citep{wang2022greenkgc}.}\label{tab:low}
{\scriptsize
\begin{tabular}{l c  c  c  c  c  c  c  c  c  c  c  c}
\toprule
& \multicolumn{4}{c}{\textbf{FB15k-237}} & \multicolumn{4}{c}{\textbf{WN18RR}} & \multicolumn{4}{c}{\textbf{YAGO3-10}} \\
\midrule
Model & MRR & H@1 & H@3 & H@10 & MRR & H@1 & H@3 & H@10 & MRR & H@1 & H@3 & H@10 \\
\midrule \midrule
\multicolumn{13}{c}{\textit{Knowledge Graph Embedding Methods}} \\
TransE \citep{bordes2013translating} & 0.270 & 0.177 & 0.303 & 0.457 
       & 0.221 & 0.020 & 0.388 & 0.517
       & 0.324 & 0.221 & 0.374 & 0.524 \\
RotatE \citep{sun2019rotate} & 0.290 & 0.208 & 0.316 & 0.458 
       & 0.387 & 0.330 & 0.417 & 0.491 
       & \underline{0.419} & \underline{0.321} & \underline{0.475} & \underline{0.607} \\
\midrule \midrule
\multicolumn{13}{c}{\textit{Classification-based KGC Methods}} \\
ConvKB \citep{nguyen2017novel} & 0.232 & 0.157 & 0.255 & 0.377
       & 0.346 & 0.300 & 0.374 & 0.422 
       & 0.311 & 0.194 & 0.368 & 0.526 \\
ConvE  \citep{dettmers2018convolutional} & 0.282 & 0.201 & 0.309 &  0.440 
       & 0.405 & 0.377 & 0.412 & 0.453
       & 0.361 & 0.260 & 0.396 & 0.559 \\
\midrule \midrule
\multicolumn{13}{c}{\textit{Low-dimensional KGE Methods}} \\
AttH   \citep{chami2020low} & \underline{0.324} & \underline{0.236} & \underline{0.354} & \underline{0.501} 
       & \underline{0.466} & \underline{0.419} & \underline{0.484} & \underline{0.551} 
       & 0.397 & 0.310 & 0.437 & 0.566 \\
DualDE \citep{zhu2022dualde} & 0.306 & 0.216 & 0.338 & 0.489 
       & \textbf{0.468} & \textbf{0.419} & \textbf{0.486} & \textbf{0.560} 
       & - & - & - & - \\
\midrule
GreenKGC \citep{wang2022greenkgc}  & \textbf{0.326} & \textbf{0.248} & \textbf{0.351} & \textbf{0.479}
       & 0.411 & 0.367 & 0.430 & 0.491
       & \textbf{0.453} & \textbf{0.361} & \textbf{0.509} & \textbf{0.629} \\
\bottomrule
\end{tabular}}
\end{table}

\section{Future Outlook}\label{sec:outlook}

\subsection{Robustness}\label{subsec:robustness}

Today's DL networks are dominated by feedforward neural networks, i.e.,
information moves in only one forward direction during the inference
(i.e., from input nodes, through hidden nodes, and to output nodes).
They do not have cycles or loops. Their filter weights are obtained by
end-to-end optimization using the backpropagation algorithm. One
drawback of feedforward neural networks is that it is easy to add small
perturbations to the input to fool DL. This is known as the adversarial
attack \citep{akhtar2018threat}. Adversarial training
\citep{madry2017towards} has been proposed to improve the robustness of
DL networks albeit at the expense of poorer performance against clean
data. 

There is little research on the robustness of GL systems. To attack an
arbitrary input to a GL system, one needs to modify its representation,
the feature learner (i.e., DFT) or the classifier.  The latter two are
related to statistics of training data. They cannot be controlled by a
single input. As to the first one, we argue that adversarial attacks
designed for DL systems would not harm GL systems.  This is because GL
systems adopt signal transforms to reduce representation's dimension,
where signal transforms are obtained by statistics of training samples.
Small perturbations to the input obtained by adversarial attacks to DL
systems will be filtered out by signal transforms.  Furthermore, the
ensemble learning technique can be leveraged by GL to enhance its
accuracy since each individual GL decision is lightweight. 

Apparently, if the distributions of sampled training and test data are
sufficiently different, the trained GL systems will not predict well.
Yet, this is a common issue to all ML systems.  Generally speaking, it
is an open research problem to conduct adversarial attacks that are
specifically effective against GL systems. To this end, it is also not
clear on the design of more resilient GL systems. 

Building large AI/ML datasets is a laborious task.  Once data are
collected, they are partitioned into training and test data by a certain
ratio.  However, the methodology in collecting fewer yet representative
data for certain applications is rarely addressed in the current AI/ML
literature. Most datasets have been built in a brute force yet ad hoc
fashion. For certain downstream tasks, how to build relevant datasets is
critical to all ML systems. 

\subsection{Trust and Risk Assessment}

The application of blackbox ML models to high stakes' decisions such as
those involved in healthcare, legal judgment, critical infrastructure
management has been challenged by researchers, e.g.,
\citep{arrieta2020explainable, poursabzi2021manipulating, rudin2019stop}.
Today's evaluations conducted on DL models are based on a certain split
of training and test data, which are collected using the same protocol.
Such an evaluation methodology could be too simple to be trustworthiness.
Conclusions drawn from a set of input-output relationships can be
misleading and counter intuition. It is unclear how to quantify the
prediction performance on unseen data. It is difficult to provide a
logical description of the decision process to experts in the field to
win their trust.  For GL to be well accepted by the society, especially
those with severe consequences, it should be an interpretable discipline
validated by scientific rigor. Since GL is statistically and
mathematically rooted, it will yield probabilistic models that have
performance guarantees and allow risk assessment. 

The risk of GL's decisions can be attributed to three factors: 1)
sampling bias and fairness \citep{mehrabi2021survey,
zadrozny2004learning}; 2) labeling errors \citep{northcutt2021pervasive};
3) analysis errors.  Some domains have tighter quality assurance and
error control such as new drug development and medical treatment
procedure.  Others may have lower standards. For example, autonomous
driving should belong to the high stakes decision category. Due to the
dominance of DL in the field of computer vision and the blackbox nature
of DL, we do not see much error analysis on the methodology. This in
turn affects the standards on the first two factors. Although GL is
still at its infancy, it is important to include risk analysis in this
emerging discipline. It should be tolerant to labeling errors in the
training process. It should be able to estimate errors against different
data types, say, inliers are more robust while outliers are less
reliable. 

\subsection{Generative Models}\label{subsec:generation}

Given a family of images of similar characteristics, one ML problem is
to generate more images that share the same characteristics. DL has
achieved a significant amount of progress in this area. DL generative
models can be categorized into two types: non-adversarial and
adversarial ones. The former includes variational auto-encoders
(VAEs)~\citep{kingma2013auto}, GLO~\citep{bojanowski2018optimizing},
IMLE~\citep{li2018implicit} and GLANN~\citep{hoshen2019non} while the
latter includes the original GAN~\citep{goodfellow2014generative},
LSGAN~\citep{mao2017least}, WGAN~\citep{arjovsky2017wasserstein}, etc.  DL
generative models take an arbitrary white Gaussian random vector as
their input and generate an image of the target family via a DL
generator (or decoder). All outputs at intermediate layers are latent
variables. It is difficult to give them a physical meaning. 

There are preliminary efforts in the development of GL generative
models, e.g., NITES \citep{lei2020nites}, TGHop \citep{lei2021tghop},
Pager \citep{azizi2022pager}. GL generative models contains two modules:
\begin{itemize}
\item Module 1: Fine-to-Coarse Image Analysis \\
The dimension of color images of $N \times N$ pixels is $3N^2$. As $N$
becomes large, the diversity is too high to model with a reasonable
mathematical model.  To reduce the diversity, one straightforward idea
is to partition one large image into non-overlapping smaller blocks,
each of which has a significantly lower dimension. However, these blocks
cannot be generated independently since they are correlated with each
other. To take the block correlation into account, one can downsample
fine-resolution images to coarse-resolution ones and compute the
residual image between interpolated coarse-resolution images and
fine-resolution images. Then, to synthesize the fine-resolution images,
the GL generative model only needs to model residual images as well as
the coarse-resolution images. This process can be done recursively to
get residual images at multiple resolutions as well as the
coarsest-resolution images. This pipeline is called the fine-to-coarse
image analysis.
\item Module 2: Coarse-to-Fine Image Synthesis \\
Usually, the space formed by coarsest-resolution images is called the
core subspace. A sample in the core subspace corresponds to a
coarsest-resolution image. The core subspace was modeled using the
independent component analysis (ICA) in \citep{lei2021tghop} and the
Gaussian mixture model (GMM) in \citep{azizi2022pager}.  After a sample
is generated in the core subspace, we need to add more details to it.
To achieve this objective, we partition coarse-resolution images into
smaller blocks and use their content as the condition for detail
generation. In principle, these conditional probabilities can be learned
from Module 1. Then, one can generate high-resolution images through the
core generation as well as a sequence of detail generations. 
\end{itemize}
Although the high-level idea of GL generation can be described by words
easily above, it demands good mathematical tools, mature programming
skills and strong engineering insights to implement a high-performance
GL generation system. A recent paper \citep{granot2022drop} points to a
green single-image generative solution and its applications in
retargeting, conditional inpainting, collage, etc. 

\subsection{Structures of High-Dimensional Representation Space}\label{subsec:high}

The representation space of GL is derived by source data statistics (e.g.,
image pixel correlations) and a set of features that minimizes a cost
function defined by labels (or tasks). The process from the source space
to the representation space is deterministic and the resulting representation is
unique for the same representation. This is different from the case of
DL. Even for the same network, DL's embeddings are not uniquely defined.
They depend on the initialization of the filter weights and the order of
input samples in the stochastic gradient descent optimization process.
Since the representation space of a GL system can be uniquely specified, its
geometrical analysis will facilitate the development of GL-based learning
algorithms. 

Geometrical analysis of low-dimensional feature spaces was exploited
in \citep{lin2022geometrical} to design the multilayer perceptrons for
cases. Analysis of high-dimensional representation spaces remain to be a
challenge. The advancement in this area will contribute to development
of GL technologies. For example, we need to model the distributions of
representations in the core and residual subspaces as discussed in Sec.
\ref{subsec:generation}. Although ICA and GMM have been used as generic
modeling tools, they could be fine-tuned to match distribution's geometry
to yield simpler yet more powerful generative models. 

Another application scenario is the separation of inliers and outliers
in the representation space. The number of inliers is significantly higher
than that of outliers. The learning of inliers can be done in the weakly
supervised fashion without data augmentation since the sample density
is higher. On the other hand, the learning of outliers should be done in
the heavily supervised fashion with data augmentation since the sample
density is lower. To this end, some preliminary research on this topic
has been conducted in \citep{yang2022design}. Generally speaking, inliers
are easy samples while outliers are hard samples. The performance of ML
algorithms is largely determined by the ratio of easy and hard samples.
It is difficult to separate inliers and outliers in DL networks in the
training pipeline. Both have to be included in the backpropagation
process. In contrast, by separating inliers and outliers, we can zoom in
outliers, conduct data augmentation, and design a tailored feature
extractor and classifier in GL systems. 

\subsection{Object Detection, Classification and Semantic Segmentation}
\label{subsec:object}

A few large-scale datasets have been built for object detection, 
classification, and semantic segmentation problems:
\begin{itemize}
\itemsep -1ex
\item Object Detection Datasets: Pascal VOC \citep{everingham2010pascal}, 
ImageNet \citep{deng2009imagenet}, MS CoCo \citep{lin2014microsoft}, etc.
\item Object Classification Datasets: ImageNet \citep{deng2009imagenet}, 
Places \citep{zhou2017places}, etc.
\item Semantic Segmentation Datasets: Pascal VOC \citep{everingham2010pascal}, 
Cityscapes \citep{cordts2016cityscapes}, etc. 
\end{itemize}
These problems have been well solved by DL methods. There are quite a
few survey papers on these topics, e.g., \citep{zeng2021deep} for scene
classification, \citep{zhao2019object} for object detection, and
\citep{minaee2021image} for semantic segmentation.  One common challenge
of the three closely related problems is the multi-scale-size and
multi-aspect-ratio of underlying objects in given input images. For
DL-based solutions, object proposals of multiple sizes and aspect ratios
are adopted, e.g. in the YOLO object detection system
\citep{redmon2016you}. 

The idea of flexible object proposals is challenging to implement in the
GL setting. Instead, we may revisit an old framework known as ``feature
pyramids + sliding windows". This methodology was intensively studied
before 2014 \citep{zou2019object}. To derive a GL solution, we may handle
objects of various sizes through feature pyramids so as to have a rough
idea about object's location, size and class based on detected salient
regions. Then, we can exploit prior statistics on the roughly estimated
object to finetune the object proposal and/or refine the classification
result. Once a rough location of an object is detected, its precise
segmentation (i.e., pixel-based classification) can be solved more
easily. 

There are several constraints in classical object detection,
classification and segmentation solutions before the DL era. For
example, it is often to use histograms of oriented gradients (HoG)
\citep{dalal2005histograms} as the local representation and the
deformable part model \citep{felzenszwalb2008discriminatively} as the
global representation.  Both of them are neither expressive nor general
enough in characterizing a large number of object classes and appearance
variations. Their performance does not scale well with the dataset size
due to the limited power of the associated representations. 

A possible GL solution is outlined below. First, we can apply a sliding
window of a fixed size to different levels of a pyramid, which is
created by the Lanczos down-sampling operation, and apply IPHop-II
\citep{yang2022design} to each block to learn a more effective local
representation.  The main purpose is to capture distinctive blocks that
are associated with certain object classes. Next, we can assemble
neighboring blocks of the same predicted class, analyze their relation,
and build a global representation. For example, to detect a horse, we
can see the whole horse in a sliding window at a higher level of a
pyramid and observe parts of the same horse in multiple sliding windows
at lower levels of the same location. The local representation can be 3D
Saab coefficients (2D from the space and 1D from the scale). The above 
sketched idea remains to be tested.

\subsection{Multi-Domain Data Integration}

ML with multi-domain data such as images, videos, point clouds, texts,
speeches, audios, graphs is the emerging trend.  It is straightforward
to apply GL to a single data domain for representation learning.
Non-image-domain applications of GL can be found in some exploratory
work - videos \citep{wei2022expressionhop}, point clouds \citep{
kadam2021r, kadam2020unsupervised, kadam2022pcrp, liu20213d,
zhang2021gsip, zhang2020unsupervised, zhang2020pointhop++,
zhang2020pointhop}, texts \citep{wei2022synwmd, wei2022task}, graphs
\citep{xie2022graphhop,xie2022graphhop++} and knowledge graphs
\citep{ge2022compounde, wang2022kgboost}.  Generally, we can concatenate
representation vectors learned from each domain into ``hybrid"
representation vectors for multi-domain input samples.  The main
challenge is that the dimension of hybrid representation vectors is
extremely high. The training data will be too sparse in the
representation space to yield meaningful learning results. We need some
mechanism to zoom into a lower dimensional space. To this end, we may
leverage co-occurrence (e.g., joint or conditional probabilities),
priors, constraints. 

Co-occurrence of representations and labels has been exploited in attention
localization in \citep{yang2022statistical} as explained. We use the
problem of dog/cat image classification to illustrate it.  Each image
can only have one label (i.e., dog or cat) despite the existence of
diverse backgrounds (e.g., indoor or outdoor).  We partition input
images into local patches, assign the source image label to resulting
patches, and group patches of visual similarity into clusters.  There
are clusters of background regions (say, wall, window, furniture, floor,
sky, grass, street, etc.). They are typically shared by dog and cat
images.  As a result, background patches with dog and cat labels
co-exist in these clusters.  Such clusters have a higher entropy value.
In contrast, clusters contain regions specific to dog/cat images (e.g.,
their facial regions) are dominated by one class. Such clusters have a
lower entropy value. Patches in low entropy clusters can be chosen as
attention regions. This idea could be extended to object tracking and
video object segmentation, and further exploration along this direction
could be fruitful. 

To train an automatic GL-based image/video captioning system may
leverage the co-occurrence of multi-domain representations and labels.
First, we need build a link between simple visual data units (e.g.
objects and scenes) and their texts. This training can be conducted
based on the ImageNet or the Places dataset \citep{zhou2017places}.  The
scene recognition could be challenging due to the occlusion of the
foreground objects. On the other hand, the correlation between an object
and its environment can be exploited. For example, a giraffe is most
likely to be in the wild or zoo. In the inference stage, we can first
identify main objects and the background in an input image using their
image representations. Next, we need to describe interactions between objects
and/or interactions between object/environment. They could be learned
from the image/video captioning training datasets. Interaction in the
text domain is easier to capture.  One way is to build a dependency
parsing tree \citep{wei2022synwmd, wei2022task} whose child nodes contain
main objects and/or background. Various interactions between child nodes
define ``actions". We group actions into similar types, and go back to
the image domain to find similarities among their associated images
with an objective to connect image- and text-based action representations.

\subsection{Lightweight DL versus GL}

Development of lightweight DL networks of smaller sizes and faster
computation is an active research topic in the DL community.  One
approach is to compress large DL networks \citep{alvarez2017compression,
cheng2017survey, choudhary2020comprehensive, hoefler2021sparsity,
louizos2017bayesian, murshed2021machine} to smaller ones.  This can be
achieved by quantization and/or model pruning.  The former lowers the
precision of model parameters while the latter reduces the number of
model parameters. They can be applied on a given model one after the
other. Another approach is to design smaller DL architectures targeting
at mobile/edge computing platforms from scratch. Several well-known
networks belong to this class, e.g., MobileNets
\citep{howard2017mobilenets}, ShuffleNet \citep{zhang2018shufflenet} and
Light CNN \citep{wu2018light}. Usually, the lightweight DL networks trade
degraded performance for less memory and computation requirement. 

Although little performance benchmarking between GL models and
lightweight DL models has been conducted, we do find one example in
\citep{chen2022defakehop++} for deepfake video detection.  The work
compares a GL system called DefakeHop++ and MobileNet v3.  The numbers
of model parameters of DefakeHop++ and MobileNet v3 are 238K and 1.5M,
respectively. Furthermore, DefakeHop++ outperforms MobileNet V3 (without
pre-training) in detection performance. The performance gap is higher as
the percentages of training samples go lower. This preliminary study
demonstrates the enormous potential of GL models over simplified DL models.
More extensive evaluations of DL and GL comparison in a wider range of
settings and applications are needed to draw more definite conclusion. 

\section{Conclusion}\label{sec:conclusion}

The objective of this overview paper is to provide an introduction to
the new GL methodology with illustrative examples.  GL adopts a
modularized design, where each module is optimized independently for
implementational efficiency.  Initially, research on GL was conducted to
provide a better understanding of CNNs. The study led to the design of
an interpretable FF-CNN.  FF-CNN derives network parameters of the
current layer based on data statistics from the output of the previous
layer in a one-pass feedforward manner.  Filter weights in the
convolutional layers are conveniently computed through the Saab and c/w
Saab transforms without any labels.  Filter weights in the FC layers are
calculated with labels or pseudo-labels based on linear least squared
regression.  To reduce the performance gap between FF-CNN and the
classical BP-CNN (backpropagation-based CNN), ensemble learning is
incorporated in the FF design, leading to more powerful ML systems such
as PixelHop, PixelHop++ and IPHop. Recently, discriminant and relevant
feature selection tools and new classifiers/regressors that leverage
both feature and decision fusions have been devised to make GL a more
mature technology. 

Our overview paper intends to attract the attention and interests of
researchers to this emerging learning paradigm that is critical to
sustainability of our living environment and reliability of high stakes
decision.  The differences between GL and DL in terms of computational
complexity (or carbon footprint), storage requirement (or model size),
and logical transparency were compared with many examples throughout the
paper.  We have seen a few successful applications of GL with
performance comparable with state-of-the-art DL solutions demanding
significantly less resource.

Admittedly, GL is still at its infancy. Quite a few problems that are
well solved by DL do not have a competitive GL solution yet. We believe
that it is worthwhile to re-examine all of them from the GL angle.  Such
efforts are expected to lead to fruitful results.  Several topics on the
future research and development of GL are discussed in Sec.
\ref{sec:outlook}. The opportunities are plentiful.  Commitment to
further advancement of GL is intellectually rewarding and practically
meaningful. 

\section*{Acknowledgment}

The authors acknowledge the Center for Advanced Research Computing
(CARC) at the University of Southern California for providing computing
resources that have contributed to the research results reported in this
work. The first author would also like to thank his former and current
PhD students in the Media Communications Laboratory at the University of
Southern California for their collaborative efforts in exploring green
learning technologies. Special thanks go to Hao Xu, Yueru Chen, Yijing
Yang, Min Zhang, Pranav Kadam, Mozhdeh Rouhsedaghat, Hong-Shuo Chen,
Ruiyuan Lin, Hongyu Fu, Xuejing Lei, Zohreh Azizi, Yifan Wang, Tian Xie,
Zhiruo Zhou, Wei Wang, Yao Zhu, Kaitai Zhang, Bin Wang, Chengwei Wei,
Yun Cheng Wang, Xiou Ge, Siyang Li, Jiali Duan, Vasileios Magouliantis,
Zhanxuan Mei, Ganning Zhao, Xinyu Wang. 

\bibliographystyle{unsrtnat}
\bibliography{ref} 

\begin{thebibliography}{211}
\providecommand{\natexlab}[1]{#1}
\providecommand{\url}[1]{\texttt{#1}}
\expandafter\ifx\csname urlstyle\endcsname\relax
  \providecommand{\doi}[1]{doi: #1}\else
  \providecommand{\doi}{doi: \begingroup \urlstyle{rm}\Url}\fi

\bibitem[Gu et~al.(2018)Gu, Wang, Kuen, Ma, Shahroudy, Shuai, Liu, Wang, Wang,
  Cai, et~al.]{gu2018recent}
Jiuxiang Gu, Zhenhua Wang, Jason Kuen, Lianyang Ma, Amir Shahroudy, Bing Shuai,
  Ting Liu, Xingxing Wang, Gang Wang, Jianfei Cai, et~al.
\newblock Recent advances in convolutional neural networks.
\newblock \emph{Pattern recognition}, 77:\penalty0 354--377, 2018.

\bibitem[Pascanu et~al.(2013)Pascanu, Mikolov, and
  Bengio]{pascanu2013difficulty}
Razvan Pascanu, Tomas Mikolov, and Yoshua Bengio.
\newblock On the difficulty of training recurrent neural networks.
\newblock In \emph{International conference on machine learning}, pages
  1310--1318. PMLR, 2013.

\bibitem[Salehinejad et~al.(2017)Salehinejad, Sankar, Barfett, Colak, and
  Valaee]{salehinejad2017recent}
Hojjat Salehinejad, Sharan Sankar, Joseph Barfett, Errol Colak, and Shahrokh
  Valaee.
\newblock Recent advances in recurrent neural networks.
\newblock \emph{arXiv preprint arXiv:1801.01078}, 2017.

\bibitem[Su et~al.(2022)Su, Kuo, et~al.]{su2022recurrent}
Yuanhang Su, C-C~Jay Kuo, et~al.
\newblock Recurrent neural networks and their memory behavior: A survey.
\newblock \emph{APSIPA Transactions on Signal and Information Processing},
  11\penalty0 (1), 2022.

\bibitem[Greff et~al.(2016)Greff, Srivastava, Koutn{\'\i}k, Steunebrink, and
  Schmidhuber]{greff2016lstm}
Klaus Greff, Rupesh~K Srivastava, Jan Koutn{\'\i}k, Bas~R Steunebrink, and
  J{\"u}rgen Schmidhuber.
\newblock Lstm: A search space odyssey.
\newblock \emph{IEEE transactions on neural networks and learning systems},
  28\penalty0 (10):\penalty0 2222--2232, 2016.

\bibitem[Hochreiter and Schmidhuber(1997)]{hochreiter1997long}
Sepp Hochreiter and J{\"u}rgen Schmidhuber.
\newblock Long short-term memory.
\newblock \emph{Neural computation}, 9\penalty0 (8):\penalty0 1735--1780, 1997.

\bibitem[Su and Kuo(2019)]{su2019extended}
Yuanhang Su and C-C~Jay Kuo.
\newblock On extended long short-term memory and dependent bidirectional
  recurrent neural network.
\newblock \emph{Neurocomputing}, 356:\penalty0 151--161, 2019.

\bibitem[Krizhevsky et~al.(2012)Krizhevsky, Sutskever, and
  Hinton]{NIPS2012_AlexNet}
Alex Krizhevsky, Ilya Sutskever, and Geoffrey~E. Hinton.
\newblock Imagenet classification with deep convolutional neural networks.
\newblock In F.~Pereira, C.~J.~C. Burges, L.~Bottou, and K.~Q. Weinberger,
  editors, \emph{Advances in Neural Information Processing Systems 25}, pages
  1097--1105. Curran Associates, Inc., 2012.

\bibitem[LeCun et~al.(2015)LeCun, Bengio, and Hinton]{Nature2015}
Yann LeCun, Yoshua Bengio, and Geoffrey~E. Hinton.
\newblock Deep learning.
\newblock \emph{Nature}, 521:\penalty0 436--444, 2015.

\bibitem[He et~al.(2015)He, Zhang, Ren, and Sun]{he2015delving}
Kaiming He, Xiangyu Zhang, Shaoqing Ren, and Jian Sun.
\newblock Delving deep into rectifiers: Surpassing human-level performance on
  imagenet classification.
\newblock In \emph{Proceedings of the IEEE international conference on computer
  vision}, pages 1026--1034, 2015.

\bibitem[Huang et~al.(2017)Huang, Liu, Van Der~Maaten, and
  Weinberger]{huang2017densely}
Gao Huang, Zhuang Liu, Laurens Van Der~Maaten, and Kilian~Q Weinberger.
\newblock Densely connected convolutional networks.
\newblock In \emph{Proceedings of the IEEE conference on computer vision and
  pattern recognition}, pages 4700--4708, 2017.

\bibitem[Deng et~al.(2009)Deng, Dong, Socher, Li, Li, and
  Fei-Fei]{deng2009imagenet}
Jia Deng, Wei Dong, Richard Socher, Li-Jia Li, Kai Li, and Li~Fei-Fei.
\newblock Imagenet: A large-scale hierarchical image database.
\newblock In \emph{2009 IEEE conference on computer vision and pattern
  recognition}, pages 248--255. Ieee, 2009.

\bibitem[Han et~al.(2022)Han, Wang, Chen, Chen, Guo, Liu, Tang, Xiao, Xu, Xu,
  et~al.]{han2022survey}
Kai Han, Yunhe Wang, Hanting Chen, Xinghao Chen, Jianyuan Guo, Zhenhua Liu,
  Yehui Tang, An~Xiao, Chunjing Xu, Yixing Xu, et~al.
\newblock A survey on vision transformer.
\newblock \emph{IEEE transactions on pattern analysis and machine
  intelligence}, 2022.

\bibitem[Jaderberg et~al.(2015)Jaderberg, Simonyan, Zisserman,
  et~al.]{jaderberg2015spatial}
Max Jaderberg, Karen Simonyan, Andrew Zisserman, et~al.
\newblock Spatial transformer networks.
\newblock \emph{Advances in neural information processing systems}, 28, 2015.

\bibitem[Khan et~al.(2021)Khan, Naseer, Hayat, Zamir, Khan, and
  Shah]{khan2021transformers}
Salman Khan, Muzammal Naseer, Munawar Hayat, Syed~Waqas Zamir, Fahad~Shahbaz
  Khan, and Mubarak Shah.
\newblock Transformers in vision: A survey.
\newblock \emph{ACM Computing Surveys (CSUR)}, 2021.

\bibitem[Vaswani et~al.(2017)Vaswani, Shazeer, Parmar, Uszkoreit, Jones, Gomez,
  Kaiser, and Polosukhin]{vaswani2017attention}
Ashish Vaswani, Noam Shazeer, Niki Parmar, Jakob Uszkoreit, Llion Jones,
  Aidan~N Gomez, {\L}ukasz Kaiser, and Illia Polosukhin.
\newblock Attention is all you need.
\newblock \emph{Advances in neural information processing systems}, 30, 2017.

\bibitem[Wolf et~al.(2020)Wolf, Debut, Sanh, Chaumond, Delangue, Moi, Cistac,
  Rault, Louf, Funtowicz, et~al.]{wolf2020transformers}
Thomas Wolf, Lysandre Debut, Victor Sanh, Julien Chaumond, Clement Delangue,
  Anthony Moi, Pierric Cistac, Tim Rault, R{\'e}mi Louf, Morgan Funtowicz,
  et~al.
\newblock Transformers: State-of-the-art natural language processing.
\newblock In \emph{Proceedings of the 2020 conference on empirical methods in
  natural language processing: system demonstrations}, pages 38--45, 2020.

\bibitem[Akhtar and Mian(2018)]{akhtar2018threat}
Naveed Akhtar and Ajmal Mian.
\newblock Threat of adversarial attacks on deep learning in computer vision: A
  survey.
\newblock \emph{Ieee Access}, 6:\penalty0 14410--14430, 2018.

\bibitem[Chan et~al.(2022)Chan, You, Qi, Wright, and Ma]{chan2022redunet}
Kwan Ho~Ryan Chan, Chong You, Haozhi Qi, John Wright, and Yi~Ma.
\newblock Redunet: A white-box deep network from the principle of maximizing
  rate reduction.
\newblock \emph{J Mach Learn Res}, 23\penalty0 (114):\penalty0 1--103, 2022.

\bibitem[Damian et~al.(2022)Damian, Lee, and Soltanolkotabi]{damian2022neural}
Alexandru Damian, Jason Lee, and Mahdi Soltanolkotabi.
\newblock Neural networks can learn representations with gradient descent.
\newblock In \emph{Conference on Learning Theory}, pages 5413--5452. PMLR,
  2022.

\bibitem[Ma et~al.(2022)Ma, Tsao, and Shum]{ma2022principles}
Yi~Ma, Doris Tsao, and Heung-Yeung Shum.
\newblock On the principles of parsimony and self-consistency for the emergence
  of intelligence.
\newblock \emph{Frontiers of Information Technology \& Electronic Engineering},
  pages 1--26, 2022.

\bibitem[Oymak et~al.(2021)Oymak, Li, and
  Soltanolkotabi]{oymak2021generalization}
Samet Oymak, Mingchen Li, and Mahdi Soltanolkotabi.
\newblock Generalization guarantees for neural architecture search with
  train-validation split.
\newblock In \emph{International Conference on Machine Learning}, pages
  8291--8301. PMLR, 2021.

\bibitem[Soltanolkotabi et~al.(2018)Soltanolkotabi, Javanmard, and
  Lee]{soltanolkotabi2018theoretical}
Mahdi Soltanolkotabi, Adel Javanmard, and Jason~D Lee.
\newblock Theoretical insights into the optimization landscape of
  over-parameterized shallow neural networks.
\newblock \emph{IEEE Transactions on Information Theory}, 65\penalty0
  (2):\penalty0 742--769, 2018.

\bibitem[Wright and Ma(2022)]{wright2022high}
John Wright and Yi~Ma.
\newblock \emph{High-dimensional data analysis with low-dimensional models:
  Principles, computation, and applications}.
\newblock Cambridge University Press, 2022.

\bibitem[Zhang et~al.(2019{\natexlab{a}})Zhang, Yu, Jiao, Xing, El~Ghaoui, and
  Jordan]{zhang2019theoretically}
Hongyang Zhang, Yaodong Yu, Jiantao Jiao, Eric Xing, Laurent El~Ghaoui, and
  Michael Jordan.
\newblock Theoretically principled trade-off between robustness and accuracy.
\newblock In \emph{International conference on machine learning}, pages
  7472--7482. PMLR, 2019{\natexlab{a}}.

\bibitem[Misra and Maaten(2020)]{misra2020self}
Ishan Misra and Laurens van~der Maaten.
\newblock Self-supervised learning of pretext-invariant representations.
\newblock In \emph{Proceedings of the IEEE/CVF Conference on Computer Vision
  and Pattern Recognition}, pages 6707--6717, 2020.

\bibitem[Sohn et~al.(2020)Sohn, Berthelot, Carlini, Zhang, Zhang, Raffel,
  Cubuk, Kurakin, and Li]{sohn2020fixmatch}
Kihyuk Sohn, David Berthelot, Nicholas Carlini, Zizhao Zhang, Han Zhang,
  Colin~A Raffel, Ekin~Dogus Cubuk, Alexey Kurakin, and Chun-Liang Li.
\newblock Fixmatch: Simplifying semi-supervised learning with consistency and
  confidence.
\newblock \emph{Advances in neural information processing systems},
  33:\penalty0 596--608, 2020.

\bibitem[Van~Engelen and Hoos(2020)]{van2020survey}
Jesper~E Van~Engelen and Holger~H Hoos.
\newblock A survey on semi-supervised learning.
\newblock \emph{Machine Learning}, 109\penalty0 (2):\penalty0 373--440, 2020.

\bibitem[Lannelongue et~al.(2021)Lannelongue, Grealey, and
  Inouye]{lannelongue2021green}
Lo{\"\i}c Lannelongue, Jason Grealey, and Michael Inouye.
\newblock Green algorithms: quantifying the carbon footprint of computation.
\newblock \emph{Advanced science}, 8\penalty0 (12):\penalty0 2100707, 2021.

\bibitem[Schwartz et~al.(2020)Schwartz, Dodge, Smith, and
  Etzioni]{schwartz2020green}
Roy Schwartz, Jesse Dodge, Noah~A Smith, and Oren Etzioni.
\newblock Green {AI}.
\newblock \emph{Communications of the ACM}, 63\penalty0 (12):\penalty0 54--63,
  2020.

\bibitem[Wu et~al.(2022)Wu, Raghavendra, Gupta, Acun, Ardalani, Maeng, Chang,
  Aga, Huang, Bai, et~al.]{wu2022sustainable}
Carole-Jean Wu, Ramya Raghavendra, Udit Gupta, Bilge Acun, Newsha Ardalani,
  Kiwan Maeng, Gloria Chang, Fiona Aga, Jinshi Huang, Charles Bai, et~al.
\newblock Sustainable {AI}: Environmental implications, challenges and
  opportunities.
\newblock \emph{Proceedings of Machine Learning and Systems}, 4:\penalty0
  795--813, 2022.

\bibitem[Xu et~al.(2021)Xu, Zhou, Fu, Zhou, and Li]{xu2021survey}
Jingjing Xu, Wangchunshu Zhou, Zhiyi Fu, Hao Zhou, and Lei Li.
\newblock A survey on green deep learning.
\newblock \emph{arXiv preprint arXiv:2111.05193}, 2021.

\bibitem[Sanh et~al.(2019)Sanh, Debut, Chaumond, and Wolf]{sanh2019distilbert}
Victor Sanh, Lysandre Debut, Julien Chaumond, and Thomas Wolf.
\newblock {DistilBERT}, a distilled version of {BERT}: smaller, faster, cheaper
  and lighter.
\newblock \emph{arXiv preprint arXiv:1910.01108}, 2019.

\bibitem[Sharir et~al.(2020)Sharir, Peleg, and Shoham]{sharir2020cost}
Or~Sharir, Barak Peleg, and Yoav Shoham.
\newblock The cost of training {NLP} models: A concise overview.
\newblock \emph{arXiv preprint arXiv:2004.08900}, 2020.

\bibitem[Strubell et~al.(2019)Strubell, Ganesh, and
  McCallum]{strubell2019energy}
Emma Strubell, Ananya Ganesh, and Andrew McCallum.
\newblock Energy and policy considerations for deep learning in {NLP}.
\newblock \emph{arXiv preprint arXiv:1906.02243}, 2019.

\bibitem[Arrieta et~al.(2020)Arrieta, D{\'\i}az-Rodr{\'\i}guez, Del~Ser,
  Bennetot, Tabik, Barbado, Garc{\'\i}a, Gil-L{\'o}pez, Molina, Benjamins,
  et~al.]{arrieta2020explainable}
Alejandro~Barredo Arrieta, Natalia D{\'\i}az-Rodr{\'\i}guez, Javier Del~Ser,
  Adrien Bennetot, Siham Tabik, Alberto Barbado, Salvador Garc{\'\i}a, Sergio
  Gil-L{\'o}pez, Daniel Molina, Richard Benjamins, et~al.
\newblock Explainable artificial intelligence (xai): Concepts, taxonomies,
  opportunities and challenges toward responsible ai.
\newblock \emph{Information fusion}, 58:\penalty0 82--115, 2020.

\bibitem[Poursabzi-Sangdeh et~al.(2021)Poursabzi-Sangdeh, Goldstein, Hofman,
  Wortman~Vaughan, and Wallach]{poursabzi2021manipulating}
Forough Poursabzi-Sangdeh, Daniel~G Goldstein, Jake~M Hofman, Jennifer~Wortman
  Wortman~Vaughan, and Hanna Wallach.
\newblock Manipulating and measuring model interpretability.
\newblock In \emph{Proceedings of the 2021 CHI conference on human factors in
  computing systems}, pages 1--52, 2021.

\bibitem[Rudin(2019)]{rudin2019stop}
Cynthia Rudin.
\newblock Stop explaining black box machine learning models for high stakes
  decisions and use interpretable models instead.
\newblock \emph{Nature Machine Intelligence}, 1\penalty0 (5):\penalty0
  206--215, 2019.

\bibitem[Kuo(2017)]{kuo2017cnn}
C-C~Jay Kuo.
\newblock The cnn as a guided multilayer recos transform [lecture notes].
\newblock \emph{IEEE signal processing magazine}, 34\penalty0 (3):\penalty0
  81--89, 2017.

\bibitem[Kuo(2016)]{kuo2016understanding}
C-C~Jay Kuo.
\newblock Understanding convolutional neural networks with a mathematical
  model.
\newblock \emph{Journal of Visual Communication and Image Representation},
  41:\penalty0 406--413, 2016.

\bibitem[Kuo and Chen(2018)]{kuo2018data}
C-C~Jay Kuo and Yueru Chen.
\newblock On data-driven saak transform.
\newblock \emph{Journal of Visual Communication and Image Representation},
  50:\penalty0 237--246, 2018.

\bibitem[Kuo et~al.(2019)Kuo, Zhang, Li, Duan, and Chen]{kuo2019interpretable}
C-C~Jay Kuo, Min Zhang, Siyang Li, Jiali Duan, and Yueru Chen.
\newblock Interpretable convolutional neural networks via feedforward design.
\newblock \emph{Journal of Visual Communication and Image Representation},
  60:\penalty0 346--359, 2019.

\bibitem[Chen and Kuo(2020)]{chen2020pixelhop}
Yueru Chen and C-C~Jay Kuo.
\newblock Pixelhop: A successive subspace learning ({SSL}) method for object
  recognition.
\newblock \emph{Journal of Visual Communication and Image Representation}, page
  102749, 2020.

\bibitem[Chen et~al.(2020{\natexlab{a}})Chen, Rouhsedaghat, You, Rao, and
  Kuo]{chen2020pixelhop++}
Yueru Chen, Mozhdeh Rouhsedaghat, Suya You, Raghuveer Rao, and C-C~Jay Kuo.
\newblock Pixelhop++: A small successive-subspace-learning-based ({SSL}-based)
  model for image classification.
\newblock In \emph{2020 IEEE International Conference on Image Processing
  (ICIP)}, pages 3294--3298. IEEE, 2020{\natexlab{a}}.

\bibitem[Yang et~al.(2022{\natexlab{a}})Yang, Fu, and Kuo]{yang2022design}
Yijing Yang, Hongyu Fu, and C-C~Jay Kuo.
\newblock Design of supervision-scalable learning systems: Methodology and
  performance benchmarking.
\newblock \emph{arXiv preprint arXiv:2206.09061}, 2022{\natexlab{a}}.

\bibitem[Yang et~al.(2022{\natexlab{b}})Yang, Wang, Fu, and
  Kuo]{yang2022supervised}
Yijing Yang, Wei Wang, Hongyu Fu, and C-C~Jay Kuo.
\newblock On supervised feature selection from high dimensional feature spaces.
\newblock \emph{arXiv preprint arXiv:2203.11924}, 2022{\natexlab{b}}.

\bibitem[Fu et~al.(2022{\natexlab{a}})Fu, Yang, Mishra, and
  Kuo]{fu2022subspace}
Hongyu Fu, Yijing Yang, Vinod~K Mishra, and C-C~Jay Kuo.
\newblock Subspace learning machine (slm): Methodology and performance.
\newblock \emph{arXiv preprint arXiv:2205.05296}, 2022{\natexlab{a}}.

\bibitem[Xu et~al.(2017)Xu, Chen, Lin, and Kuo]{xu2017understanding}
Hao Xu, Yueru Chen, Ruiyuan Lin, and C-C~Jay Kuo.
\newblock Understanding cnn via deep features analysis.
\newblock In \emph{2017 Asia-Pacific Signal and Information Processing
  Association Annual Summit and Conference (APSIPA ASC)}, pages 1052--1060.
  IEEE, 2017.

\bibitem[Lin et~al.(2022)Lin, Zhou, You, Rao, and Kuo]{lin2022geometrical}
Ruiyuan Lin, Zhiruo Zhou, Suya You, Raghuveer Rao, and C-C~Jay Kuo.
\newblock Geometrical interpretation and design of multilayer perceptron
  design.
\newblock \emph{IEEE Trans. on Neural Networks and Learning Systems}, 2022.

\bibitem[LeCun et~al.(1998)LeCun, Bottou, Bengio, and Haffner]{LeNet1998}
Yann LeCun, L\'{e}on Bottou, Yoshua Bengio, and Patrick Haffner.
\newblock Gradient-based learning applied to document recognition.
\newblock \emph{Proc. IEEE}, 86\penalty0 (11):\penalty0 2278--2324, 1998.

\bibitem[Simonyan and Zisserman(2014)]{simonyan2014very}
Karen Simonyan and Andrew Zisserman.
\newblock Very deep convolutional networks for large-scale image recognition.
\newblock \emph{arXiv preprint arXiv:1409.1556}, 2014.

\bibitem[Elsken et~al.(2019)Elsken, Metzen, and Hutter]{elsken2019neural}
Thomas Elsken, Jan~Hendrik Metzen, and Frank Hutter.
\newblock Neural architecture search: A survey.
\newblock \emph{The Journal of Machine Learning Research}, 20\penalty0
  (1):\penalty0 1997--2017, 2019.

\bibitem[Cybenko(1989)]{cybenko1989approximation}
George Cybenko.
\newblock Approximation by superpositions of a sigmoidal function.
\newblock \emph{Mathematics of control, signals and systems}, 2\penalty0
  (4):\penalty0 303--314, 1989.

\bibitem[Hornik et~al.(1989)Hornik, Stinchcombe, and
  White]{hornik1989multilayer}
K~Hornik, M~Stinchcombe, and H~White.
\newblock Multilayer feedforward networks are universal approximators.
\newblock \emph{Neural Networks}, 2\penalty0 (5):\penalty0 359--366, 1989.

\bibitem[Lin et~al.(2021)Lin, You, Rao, and Kuo]{lin2021relationship}
Ruiyuan Lin, Suya You, Raghuveer Rao, and C-C~Jay Kuo.
\newblock On relationship of multilayer perceptrons and piecewise polynomial
  approximators.
\newblock \emph{IEEE Signal Processing Letters}, 28:\penalty0 1813--1817, 2021.

\bibitem[Hu et~al.(2020)Hu, Wang, et~al.]{hu2020slfb}
Yuhang Hu, Jinyan Wang, et~al.
\newblock Slfb-cnn: An interpretable neural network privacy protection
  framework.
\newblock In \emph{2020 16th International Conference on Computational
  Intelligence and Security (CIS)}, pages 298--302. IEEE, 2020.

\bibitem[Wang et~al.(2022{\natexlab{a}})Wang, Li, Hu, Li,
  et~al.]{wang2022privacy}
Jinyan Wang, Qiyu Li, Yuhang Hu, Xianxian Li, et~al.
\newblock A privacy preservation framework for feedforward-designed
  convolutional neural networks.
\newblock \emph{Neural Networks}, 2022{\natexlab{a}}.

\bibitem[Chen et~al.(2019{\natexlab{a}})Chen, Yang, Wang, and
  Kuo]{chen2019ensembles}
Yueru Chen, Yijing Yang, Wei Wang, and C-C~Jay Kuo.
\newblock Ensembles of feedforward-designed convolutional neural networks.
\newblock In \emph{2019 IEEE International Conference on Image Processing
  (ICIP)}, pages 3796--3800. IEEE, 2019{\natexlab{a}}.

\bibitem[Chen et~al.(2019{\natexlab{b}})Chen, Yang, Zhang, and
  Kuo]{chen2019semi}
Yueru Chen, Yijing Yang, Min Zhang, and C-C~Jay Kuo.
\newblock Semi-supervised learning via feedforward-designed convolutional
  neural networks.
\newblock In \emph{2019 IEEE International Conference on Image Processing
  (ICIP)}, pages 365--369. IEEE, 2019{\natexlab{b}}.

\bibitem[Solorio-Fern{\'a}ndez et~al.(2020)Solorio-Fern{\'a}ndez,
  Carrasco-Ochoa, and Mart{\'\i}nez-Trinidad]{solorio2020review}
Sa{\'u}l Solorio-Fern{\'a}ndez, J~Ariel Carrasco-Ochoa, and Jos{\'e}~Fco
  Mart{\'\i}nez-Trinidad.
\newblock A review of unsupervised feature selection methods.
\newblock \emph{Artificial Intelligence Review}, 53\penalty0 (2):\penalty0
  907--948, 2020.

\bibitem[Sheikhpour et~al.(2017)Sheikhpour, Sarram, Gharaghani, and
  Chahooki]{sheikhpour2017survey}
Razieh Sheikhpour, Mehdi~Agha Sarram, Sajjad Gharaghani, and Mohammad Ali~Zare
  Chahooki.
\newblock A survey on semi-supervised feature selection methods.
\newblock \emph{Pattern Recognition}, 64:\penalty0 141--158, 2017.

\bibitem[Huang(2015)]{huang2015supervised}
Samuel~H Huang.
\newblock Supervised feature selection: A tutorial.
\newblock \emph{Artif. Intell. Res.}, 4\penalty0 (2):\penalty0 22--37, 2015.

\bibitem[Dreiseitl and Ohno-Machado(2002)]{dreiseitl2002logistic}
Stephan Dreiseitl and Lucila Ohno-Machado.
\newblock Logistic regression and artificial neural network classification
  models: a methodology review.
\newblock \emph{Journal of biomedical informatics}, 35\penalty0 (5-6):\penalty0
  352--359, 2002.

\bibitem[Loh(2011)]{loh2011classification}
Wei-Yin Loh.
\newblock Classification and regression trees.
\newblock \emph{Wiley interdisciplinary reviews: data mining and knowledge
  discovery}, 1\penalty0 (1):\penalty0 14--23, 2011.

\bibitem[Rosenblatt(1958)]{rosenblatt1958perceptron}
Frank Rosenblatt.
\newblock The perceptron: a probabilistic model for information storage and
  organization in the brain.
\newblock \emph{Psychological review}, 65\penalty0 (6):\penalty0 386, 1958.

\bibitem[Cortes and Vapnik(1995)]{svm}
Corinna Cortes and Vladimir Vapnik.
\newblock Support-vector networks.
\newblock \emph{Machine learning}, 20\penalty0 (3):\penalty0 273--297, 1995.

\bibitem[Breiman et~al.(1984)Breiman, Friedman, Stone, and Olshen]{CART}
L~Breiman, J~Friedman, CJ~Stone, and RA~Olshen.
\newblock \emph{Classification and Regression Trees}.
\newblock CRC, Boca Raton, FL, 1984.

\bibitem[Breiman(2001)]{RF}
Leo Breiman.
\newblock Random forests.
\newblock \emph{Machine learning}, 45\penalty0 (1):\penalty0 5--32, 2001.

\bibitem[Chen and Guestrin(2016)]{chen2016xgboost}
Tianqi Chen and Carlos Guestrin.
\newblock Xgboost: A scalable tree boosting system.
\newblock In \emph{Proceedings of the 22nd acm sigkdd international conference
  on knowledge discovery and data mining}, pages 785--794, 2016.

\bibitem[Chen et~al.(2015)Chen, He, Benesty, Khotilovich, Tang, Cho, Chen,
  et~al.]{chen2015xgboost}
Tianqi Chen, Tong He, Michael Benesty, Vadim Khotilovich, Yuan Tang, Hyunsu
  Cho, Kailong Chen, et~al.
\newblock Xgboost: extreme gradient boosting.
\newblock \emph{R package version 0.4-2}, 1\penalty0 (4):\penalty0 1--4, 2015.

\bibitem[Turk and Pentland(1991)]{turk1991eigenfaces}
Matthew Turk and Alex Pentland.
\newblock Eigenfaces for recognition.
\newblock \emph{Journal of cognitive neuroscience}, 3\penalty0 (1):\penalty0
  71--86, 1991.

\bibitem[Kohavi and John(1997)]{kohavi1997wrappers}
Ron Kohavi and George~H John.
\newblock Wrappers for feature subset selection.
\newblock \emph{Artificial intelligence}, 97\penalty0 (1-2):\penalty0 273--324,
  1997.

\bibitem[Guyon et~al.(2002)Guyon, Weston, Barnhill, and Vapnik]{rfe}
Isabelle Guyon, Jason Weston, Stephen Barnhill, and Vladimir Vapnik.
\newblock Gene selection for cancer classification using support vector
  machines.
\newblock \emph{Machine learning}, 46\penalty0 (1):\penalty0 389--422, 2002.

\bibitem[Scheffe(1999)]{anova}
Henry Scheffe.
\newblock \emph{The analysis of variance}, volume~72.
\newblock John Wiley \& Sons, 1999.

\bibitem[Fu et~al.(2022{\natexlab{b}})Fu, Yang, Liu, Lin, Harrison, Mishra, and
  Kuo]{fu2022acceleration}
Hongyu Fu, Yijing Yang, Yuhuai Liu, Joseph Lin, Ethan Harrison, Vinod~K Mishra,
  and C-C~Jay Kuo.
\newblock Acceleration of subspace learning machine via particle swarm
  optimization and parallel processing.
\newblock \emph{arXiv preprint arXiv:2208.07023}, 2022{\natexlab{b}}.

\bibitem[Friedman(2001)]{GBDT}
Jerome~H Friedman.
\newblock Greedy function approximation: a gradient boosting machine.
\newblock \emph{Annals of statistics}, pages 1189--1232, 2001.

\bibitem[Zhang et~al.(2020{\natexlab{a}})Zhang, You, Kadam, Liu, and
  Kuo]{zhang2020pointhop}
Min Zhang, Haoxuan You, Pranav Kadam, Shan Liu, and C-C~Jay Kuo.
\newblock Pointhop: An explainable machine learning method for point cloud
  classification.
\newblock \emph{IEEE Transactions on Multimedia}, 2020{\natexlab{a}}.

\bibitem[Zhang et~al.(2020{\natexlab{b}})Zhang, Wang, Kadam, Liu, and
  Kuo]{zhang2020pointhop++}
Min Zhang, Yifan Wang, Pranav Kadam, Shan Liu, and C-C~Jay Kuo.
\newblock Pointhop++: A lightweight learning model on point sets for 3d
  classification.
\newblock \emph{arXiv preprint arXiv:2002.03281}, 2020{\natexlab{b}}.

\bibitem[Yang et~al.(2021)Yang, Magoulianitis, and Kuo]{yang2021pixelhop}
Yijing Yang, Vasileios Magoulianitis, and C-C~Jay Kuo.
\newblock E-pixelhop: An enhanced pixelhop method for object classification.
\newblock In \emph{2021 Asia-Pacific Signal and Information Processing
  Association Annual Summit and Conference (APSIPA ASC)}, pages 1475--1482.
  IEEE, 2021.

\bibitem[Azizi et~al.(2020)Azizi, Lei, and Kuo]{azizi2020noise}
Zohreh Azizi, Xuejing Lei, and C-C~Jay Kuo.
\newblock Noise-aware texture-preserving low-light enhancement.
\newblock In \emph{2020 IEEE International Conference on Visual Communications
  and Image Processing (VCIP)}, pages 443--446. IEEE, 2020.

\bibitem[Mei et~al.(2022)Mei, Wang, He, and Kuo]{mei2022greenbiqa}
Zhanxuan Mei, Yun-Cheng Wang, Xingze He, and C-C~Jay Kuo.
\newblock Greenbiqa: A lightweight blind image quality assessment method.
\newblock \emph{arXiv preprint arXiv:2206.14400}, 2022.

\bibitem[Zhang et~al.(2020{\natexlab{c}})Zhang, Kwong, and Kuo]{zhang2020data}
Xinfeng Zhang, Sam Kwong, and C-C~Jay Kuo.
\newblock Data-driven transform-based compressed image quality assessment.
\newblock \emph{IEEE Transactions on Circuits and Systems for Video
  Technology}, 31\penalty0 (9):\penalty0 3352--3365, 2020{\natexlab{c}}.

\bibitem[Chen et~al.(2022)Chen, Hu, You, and Kuo]{chen2022defakehop++}
Hong-Shuo Chen, Shuowen Hu, Suya You, and C-C~Jay Kuo.
\newblock Defakehop++: An enhanced lightweight deepfake detector.
\newblock \emph{arXiv preprint arXiv:2205.00211}, 2022.

\bibitem[Chen et~al.(2021{\natexlab{a}})Chen, Rouhsedaghat, Ghani, Hu, You, and
  Kuo]{chen2021defakehop}
Hong-Shuo Chen, Mozhdeh Rouhsedaghat, Hamza Ghani, Shuowen Hu, Suya You, and
  C-C~Jay Kuo.
\newblock Defakehop: A light-weight high-performance deepfake detector.
\newblock In \emph{2021 IEEE International Conference on Multimedia and Expo
  (ICME)}, pages 1--6. IEEE, 2021{\natexlab{a}}.

\bibitem[Chen et~al.(2021{\natexlab{b}})Chen, Zhang, Hu, You, and
  Kuo]{chen2021geo}
Hong-Shuo Chen, Kaitai Zhang, Shuowen Hu, Suya You, and C-C~Jay Kuo.
\newblock Geo-defakehop: High-performance geographic fake image detection.
\newblock \emph{arXiv preprint arXiv:2110.09795}, 2021{\natexlab{b}}.

\bibitem[Zhu et~al.(2022{\natexlab{a}})Zhu, Wang, Chen, Salloum, and
  Kuo]{zhu2022pixelhop}
Yao Zhu, Xinyu Wang, Hong-Shuo Chen, Ronald Salloum, and C-C~Jay Kuo.
\newblock A-pixelhop: A green, robust and explainable fake-image detector.
\newblock In \emph{ICASSP 2022-2022 IEEE International Conference on Acoustics,
  Speech and Signal Processing (ICASSP)}, pages 8947--8951. IEEE,
  2022{\natexlab{a}}.

\bibitem[Kadam et~al.(2021)Kadam, Zhang, Liu, and Kuo]{kadam2021r}
Pranav Kadam, Min Zhang, Shan Liu, and C-C~Jay Kuo.
\newblock R-pointhop: A green, accurate and unsupervised point cloud
  registration method.
\newblock \emph{arXiv preprint arXiv:2103.08129}, 2021.

\bibitem[Kadam et~al.(2020)Kadam, Zhang, Liu, and Kuo]{kadam2020unsupervised}
Pranav Kadam, Min Zhang, Shan Liu, and C-C~Jay Kuo.
\newblock Unsupervised point cloud registration via salient points analysis
  (spa).
\newblock In \emph{2020 IEEE International Conference on Visual Communications
  and Image Processing (VCIP)}, pages 5--8. IEEE, 2020.

\bibitem[Kadam et~al.(2022{\natexlab{a}})Kadam, Zhou, Liu, and
  Kuo]{kadam2022pcrp}
Pranav Kadam, Qingyang Zhou, Shan Liu, and C-C~Jay Kuo.
\newblock Pcrp: Unsupervised point cloud object retrieval and pose estimation.
\newblock \emph{arXiv preprint arXiv:2202.07843}, 2022{\natexlab{a}}.

\bibitem[Liu et~al.(2021{\natexlab{a}})Liu, Zhang, Kadam, and Kuo]{liu20213d}
Shan Liu, Min Zhang, Pranav Kadam, and C.-C.~Jay Kuo.
\newblock \emph{3D Point Cloud Analysis: Traditional, Deep Learning, and
  Explainable Machine Learning Methods}.
\newblock Springer, 2021{\natexlab{a}}.

\bibitem[Zhang et~al.(2021)Zhang, Kadam, Liu, and Kuo]{zhang2021gsip}
Min Zhang, Pranav Kadam, Shan Liu, and C-C~Jay Kuo.
\newblock Gsip: Green semantic segmentation of large-scale indoor point clouds.
\newblock \emph{arXiv preprint arXiv:2109.11835}, 2021.

\bibitem[Zhang et~al.(2020{\natexlab{d}})Zhang, Kadam, Liu, and
  Kuo]{zhang2020unsupervised}
Min Zhang, Pranav Kadam, Shan Liu, and C-C~Jay Kuo.
\newblock Unsupervised feedforward feature ({UFF}) learning for point cloud
  classification and segmentation.
\newblock In \emph{2020 IEEE International Conference on Visual Communications
  and Image Processing (VCIP)}, pages 144--147. IEEE, 2020{\natexlab{d}}.

\bibitem[Rouhsedaghat et~al.(2021{\natexlab{a}})Rouhsedaghat, Wang, Ge, Hu,
  You, and Kuo]{rouhsedaghat2021facehop}
Mozhdeh Rouhsedaghat, Yifan Wang, Xiou Ge, Shuowen Hu, Suya You, and C-C~Jay
  Kuo.
\newblock Facehop: A light-weight low-resolution face gender classification
  method.
\newblock In \emph{International Conference on Pattern Recognition}, pages
  169--183. Springer, 2021{\natexlab{a}}.

\bibitem[Rouhsedaghat et~al.(2021{\natexlab{b}})Rouhsedaghat, Wang, Hu, You,
  and Kuo]{rouhsedaghat2021low}
Mozhdeh Rouhsedaghat, Yifan Wang, Shuowen Hu, Suya You, and C-C~Jay Kuo.
\newblock Low-resolution face recognition in resource-constrained environments.
\newblock \emph{Pattern Recognition Letters}, 149:\penalty0 193--199,
  2021{\natexlab{b}}.

\bibitem[Lei et~al.(2020)Lei, Zhao, and Kuo]{lei2020nites}
Xuejing Lei, Ganning Zhao, and C.-C.~Jay Kuo.
\newblock {NITES}: A non-parametric interpretable texture synthesis method.
\newblock In \emph{2020 Asia-Pacific Signal and Information Processing
  Association Annual Summit and Conference (APSIPA ASC)}, pages 1698--1706.
  IEEE, 2020.

\bibitem[Lei et~al.(2021)Lei, Zhao, Zhang, and Kuo]{lei2021tghop}
Xuejing Lei, Ganning Zhao, Kaitai Zhang, and C-C~Jay Kuo.
\newblock Tghop: an explainable, efficient, and lightweight method for texture
  generation.
\newblock \emph{APSIPA Transactions on Signal and Information Processing}, 10,
  2021.

\bibitem[Zhang et~al.(2019{\natexlab{b}})Zhang, Chen, Wang, Ji, and
  Kuo]{zhang2019texture}
Kaitai Zhang, Hong-Shuo Chen, Ye~Wang, Xiangyang Ji, and C.-C.~Jay Kuo.
\newblock Texture analysis via hierarchical spatial-spectral correlation
  ({H}{S}{S}{C}).
\newblock In \emph{2019 IEEE International Conference on Image Processing
  (ICIP)}, pages 4419--4423. IEEE, 2019{\natexlab{b}}.

\bibitem[Xie et~al.(2022{\natexlab{a}})Xie, Kannan, and Kuo]{xie2022graphhop++}
Tian Xie, Rajgopal Kannan, and C-C~Jay Kuo.
\newblock Graphhop++: New insights into graphhop and its enhancement.
\newblock \emph{arXiv preprint arXiv:2204.08646}, 2022{\natexlab{a}}.

\bibitem[Xie et~al.(2022{\natexlab{b}})Xie, Wang, and Kuo]{xie2022graphhop}
Tian Xie, Bin Wang, and C-C~Jay Kuo.
\newblock Graphhop: An enhanced label propagation method for node
  classification.
\newblock \emph{IEEE Transactions on Neural Networks and Learning Systems},
  2022{\natexlab{b}}.

\bibitem[Chen et~al.(2020{\natexlab{b}})Chen, Shao, Wang, Li, and
  Kuo]{chen2020point}
Yueru Chen, Yiting Shao, Jing Wang, Ge~Li, and C-C~Jay Kuo.
\newblock Point cloud attribute compression via successive subspace graph
  transform.
\newblock In \emph{2020 IEEE International Conference on Visual Communications
  and Image Processing (VCIP)}, pages 66--69. IEEE, 2020{\natexlab{b}}.

\bibitem[Zhang et~al.(2020{\natexlab{e}})Zhang, Yang, Li, Liu, Yang,
  Katsavounidis, Lei, and Kuo]{zhang2020image}
Xinfeng Zhang, Chao Yang, Xiaoguang Li, Shan Liu, Haitao Yang, Ioannis
  Katsavounidis, Shaw-Min Lei, and C-C~Jay Kuo.
\newblock Image coding with data-driven transforms: Methodology, performance
  and potential.
\newblock \emph{IEEE Transactions on Image Processing}, 29:\penalty0
  9292--9304, 2020{\natexlab{e}}.

\bibitem[Tseng et~al.(2020)Tseng, Yang, Kuo, and Tsai]{tseng2020interpretable}
Tzu-Wei Tseng, Kai-Jiun Yang, C-C~Jay Kuo, and Shang-Ho Tsai.
\newblock An interpretable compression and classification system: Theory and
  applications.
\newblock \emph{IEEE Access}, 8:\penalty0 143962--143974, 2020.

\bibitem[Ding et~al.(2020)Ding, Gudapati, Lin, Fei, Packard, Song, Chang, Baek,
  Wang, Roustaei, et~al.]{ding2020saak}
Yichen Ding, Varun Gudapati, Ruiyuan Lin, Yanan Fei, Ren{\'e} R~Sevag Packard,
  Sibo Song, Chih-Chiang Chang, Kyung~In Baek, Zhaoqiang Wang, Mehrdad
  Roustaei, et~al.
\newblock Saak transform-based machine learning for light-sheet imaging of
  cardiac trabeculation.
\newblock \emph{IEEE Transactions on Biomedical Engineering}, 68\penalty0
  (1):\penalty0 225--235, 2020.

\bibitem[Liu et~al.(2021{\natexlab{b}})Liu, Xing, Yang, Kuo, Babu, Fakhri,
  Jenkins, and Woo]{liu2021voxelhop}
Xiaofeng Liu, Fangxu Xing, Chao Yang, C-C~Jay Kuo, Suma Babu, Georges~El
  Fakhri, Thomas Jenkins, and Jonghye Woo.
\newblock Voxelhop: Successive subspace learning for als disease classification
  using structural mri.
\newblock \emph{arXiv preprint arXiv:2101.05131}, 2021{\natexlab{b}}.

\bibitem[Li et~al.(2020)Li, Yang, Sun, Qi, and Lyu]{li2020celeb}
Yuezun Li, Xin Yang, Pu~Sun, Honggang Qi, and Siwei Lyu.
\newblock Celeb-df: A large-scale challenging dataset for deepfake forensics.
\newblock In \emph{Proceedings of the IEEE/CVF Conference on Computer Vision
  and Pattern Recognition}, pages 3207--3216, 2020.

\bibitem[Zhou et~al.(2017{\natexlab{a}})Zhou, Han, Morariu, and
  Davis]{zhou2017two}
Peng Zhou, Xintong Han, Vlad~I Morariu, and Larry~S Davis.
\newblock Two-stream neural networks for tampered face detection.
\newblock In \emph{2017 IEEE Conference on Computer Vision and Pattern
  Recognition Workshops (CVPRW)}, pages 1831--1839. IEEE, 2017{\natexlab{a}}.

\bibitem[Afchar et~al.(2018)Afchar, Nozick, Yamagishi, and
  Echizen]{afchar2018mesonet}
Darius Afchar, Vincent Nozick, Junichi Yamagishi, and Isao Echizen.
\newblock Mesonet: a compact facial video forgery detection network.
\newblock In \emph{2018 IEEE International Workshop on Information Forensics
  and Security (WIFS)}, pages 1--7. IEEE, 2018.

\bibitem[Yang et~al.(2019{\natexlab{a}})Yang, Li, and Lyu]{yang2019exposing}
Xin Yang, Yuezun Li, and Siwei Lyu.
\newblock Exposing deep fakes using inconsistent head poses.
\newblock In \emph{ICASSP 2019-2019 IEEE International Conference on Acoustics,
  Speech and Signal Processing (ICASSP)}, pages 8261--8265. IEEE,
  2019{\natexlab{a}}.

\bibitem[Li and Lyu(2018)]{li2018exposing}
Yuezun Li and Siwei Lyu.
\newblock Exposing deepfake videos by detecting face warping artifacts.
\newblock \emph{arXiv preprint arXiv:1811.00656}, 2018.

\bibitem[Matern et~al.(2019)Matern, Riess, and
  Stamminger]{matern2019exploiting}
Falko Matern, Christian Riess, and Marc Stamminger.
\newblock Exploiting visual artifacts to expose deepfakes and face
  manipulations.
\newblock In \emph{2019 IEEE Winter Applications of Computer Vision Workshops
  (WACVW)}, pages 83--92. IEEE, 2019.

\bibitem[Rossler et~al.(2019)Rossler, Cozzolino, Verdoliva, Riess, Thies, and
  Nie{\ss}ner]{rossler2019faceforensics++}
Andreas Rossler, Davide Cozzolino, Luisa Verdoliva, Christian Riess, Justus
  Thies, and Matthias Nie{\ss}ner.
\newblock Faceforensics++: Learning to detect manipulated facial images.
\newblock In \emph{Proceedings of the IEEE International Conference on Computer
  Vision}, pages 1--11, 2019.

\bibitem[Nguyen et~al.(2019{\natexlab{a}})Nguyen, Fang, Yamagishi, and
  Echizen]{nguyen2019multi}
Huy~H Nguyen, Fuming Fang, Junichi Yamagishi, and Isao Echizen.
\newblock Multi-task learning for detecting and segmenting manipulated facial
  images and videos.
\newblock \emph{arXiv preprint arXiv:1906.06876}, 2019{\natexlab{a}}.

\bibitem[Nguyen et~al.(2019{\natexlab{b}})Nguyen, Yamagishi, and
  Echizen]{nguyen2019use}
Huy~H Nguyen, Junichi Yamagishi, and Isao Echizen.
\newblock Use of a capsule network to detect fake images and videos.
\newblock \emph{arXiv preprint arXiv:1910.12467}, 2019{\natexlab{b}}.

\bibitem[Li and Lyu(2019)]{li2019exposing}
Yuezun Li and Siwei Lyu.
\newblock Exposing deepfake videos by detecting face warping artifacts.
\newblock In \emph{IEEE Conference on Computer Vision and Pattern Recognition
  Workshops (CVPRW)}, 2019.

\bibitem[Zhao et~al.(2021)Zhao, Zhou, Chen, Wei, Zhang, and Yu]{zhao2021multi}
Hanqing Zhao, Wenbo Zhou, Dongdong Chen, Tianyi Wei, Weiming Zhang, and Nenghai
  Yu.
\newblock Multi-attentional deepfake detection.
\newblock In \emph{Proceedings of the IEEE/CVF Conference on Computer Vision
  and Pattern Recognition}, pages 2185--2194, 2021.

\bibitem[Dolhansky et~al.(2020)Dolhansky, Bitton, Pflaum, Lu, Howes, Wang, and
  Ferrer]{dolhansky2020deepfake}
Brian Dolhansky, Joanna Bitton, Ben Pflaum, Jikuo Lu, Russ Howes, Menglin Wang,
  and Cristian~Canton Ferrer.
\newblock The deepfake detection challenge (dfdc) dataset.
\newblock \emph{arXiv preprint arXiv:2006.07397}, 2020.

\bibitem[Seferbekov(2020)]{seferbekov2020dfdc}
Selim Seferbekov.
\newblock A prize winning solution for dfdc challenge.
\newblock \url{https://github.com/selimsef/dfdc_deepfake_challenge}, 2020.

\bibitem[Lin and Kuo(2011)]{lin2011perceptual}
Weisi Lin and C-C~Jay Kuo.
\newblock Perceptual visual quality metrics: A survey.
\newblock \emph{Journal of visual communication and image representation},
  22\penalty0 (4):\penalty0 297--312, 2011.

\bibitem[Wang et~al.(2004)Wang, Bovik, Sheikh, and Simoncelli]{wang2004image}
Zhou Wang, Alan~C Bovik, Hamid~R Sheikh, and Eero~P Simoncelli.
\newblock Image quality assessment: from error visibility to structural
  similarity.
\newblock \emph{IEEE transactions on image processing}, 13\penalty0
  (4):\penalty0 600--612, 2004.

\bibitem[Zhang et~al.(2011)Zhang, Zhang, Mou, and Zhang]{zhang2011fsim}
Lin Zhang, Lei Zhang, Xuanqin Mou, and David Zhang.
\newblock Fsim: A feature similarity index for image quality assessment.
\newblock \emph{IEEE transactions on Image Processing}, 20\penalty0
  (8):\penalty0 2378--2386, 2011.

\bibitem[Li et~al.(2018{\natexlab{a}})Li, Bampis, Novak, Aaron, Swanson,
  Moorthy, and Cock]{li2018vmaf}
Zhi Li, Christos Bampis, Julie Novak, Anne Aaron, Kyle Swanson, Anush Moorthy,
  and JD~Cock.
\newblock Vmaf: The journey continues.
\newblock \emph{Netflix Technology Blog}, 25, 2018{\natexlab{a}}.

\bibitem[Yang et~al.(2019{\natexlab{b}})Yang, Li, and Liu]{yang2019survey}
Xiaohan Yang, Fan Li, and Hantao Liu.
\newblock A survey of dnn methods for blind image quality assessment.
\newblock \emph{IEEE Access}, 7:\penalty0 123788--123806, 2019{\natexlab{b}}.

\bibitem[Larson and Chandler(2010)]{larson2010most}
Eric~Cooper Larson and Damon~Michael Chandler.
\newblock Most apparent distortion: full-reference image quality assessment and
  the role of strategy.
\newblock \emph{Journal of electronic imaging}, 19\penalty0 (1):\penalty0
  011006, 2010.

\bibitem[Lin et~al.(2019)Lin, Hosu, and Saupe]{lin2019kadid}
Hanhe Lin, Vlad Hosu, and Dietmar Saupe.
\newblock Kadid-10k: A large-scale artificially distorted iqa database.
\newblock In \emph{2019 Eleventh International Conference on Quality of
  Multimedia Experience (QoMEX)}, pages 1--3. IEEE, 2019.

\bibitem[Ghadiyaram and Bovik(2015)]{ghadiyaram2015massive}
Deepti Ghadiyaram and Alan~C Bovik.
\newblock Massive online crowdsourced study of subjective and objective picture
  quality.
\newblock \emph{IEEE Transactions on Image Processing}, 25\penalty0
  (1):\penalty0 372--387, 2015.

\bibitem[Hosu et~al.(2020)Hosu, Lin, Sziranyi, and Saupe]{hosu2020koniq}
Vlad Hosu, Hanhe Lin, Tamas Sziranyi, and Dietmar Saupe.
\newblock Koniq-10k: An ecologically valid database for deep learning of blind
  image quality assessment.
\newblock \emph{IEEE Transactions on Image Processing}, 29:\penalty0
  4041--4056, 2020.

\bibitem[Mittal et~al.(2012{\natexlab{a}})Mittal, Soundararajan, and
  Bovik]{mittal2012making}
Anish Mittal, Rajiv Soundararajan, and Alan~C Bovik.
\newblock Making a completely blind image quality analyzer.
\newblock \emph{IEEE Signal processing letters}, 20\penalty0 (3):\penalty0
  209--212, 2012{\natexlab{a}}.

\bibitem[Mittal et~al.(2012{\natexlab{b}})Mittal, Moorthy, and
  Bovik]{mittal2012no}
Anish Mittal, Anush~Krishna Moorthy, and Alan~Conrad Bovik.
\newblock No-reference image quality assessment in the spatial domain.
\newblock \emph{IEEE Transactions on image processing}, 21\penalty0
  (12):\penalty0 4695--4708, 2012{\natexlab{b}}.

\bibitem[Ye et~al.(2012)Ye, Kumar, Kang, and Doermann]{ye2012unsupervised}
Peng Ye, Jayant Kumar, Le~Kang, and David Doermann.
\newblock Unsupervised feature learning framework for no-reference image
  quality assessment.
\newblock In \emph{2012 IEEE conference on computer vision and pattern
  recognition}, pages 1098--1105. IEEE, 2012.

\bibitem[Xu et~al.(2016)Xu, Ye, Li, Du, Liu, and Doermann]{xu2016blind}
Jingtao Xu, Peng Ye, Qiaohong Li, Haiqing Du, Yong Liu, and David Doermann.
\newblock Blind image quality assessment based on high order statistics
  aggregation.
\newblock \emph{IEEE Transactions on Image Processing}, 25\penalty0
  (9):\penalty0 4444--4457, 2016.

\bibitem[Kim and Lee(2016)]{kim2016fully}
Jongyoo Kim and Sanghoon Lee.
\newblock Fully deep blind image quality predictor.
\newblock \emph{IEEE Journal of selected topics in signal processing},
  11\penalty0 (1):\penalty0 206--220, 2016.

\bibitem[Bosse et~al.(2017)Bosse, Maniry, M{\"u}ller, Wiegand, and
  Samek]{bosse2017deep}
Sebastian Bosse, Dominique Maniry, Klaus-Robert M{\"u}ller, Thomas Wiegand, and
  Wojciech Samek.
\newblock Deep neural networks for no-reference and full-reference image
  quality assessment.
\newblock \emph{IEEE Transactions on image processing}, 27\penalty0
  (1):\penalty0 206--219, 2017.

\bibitem[Zeng et~al.(2018)Zeng, Zhang, and Bovik]{zeng2018blind}
Hui Zeng, Lei Zhang, and Alan~C Bovik.
\newblock Blind image quality assessment with a probabilistic quality
  representation.
\newblock In \emph{2018 25th IEEE International Conference on Image Processing
  (ICIP)}, pages 609--613. IEEE, 2018.

\bibitem[Zhang et~al.(2018{\natexlab{a}})Zhang, Ma, Yan, Deng, and
  Wang]{zhang2018blind}
Weixia Zhang, Kede Ma, Jia Yan, Dexiang Deng, and Zhou Wang.
\newblock Blind image quality assessment using a deep bilinear convolutional
  neural network.
\newblock \emph{IEEE Transactions on Circuits and Systems for Video
  Technology}, 30\penalty0 (1):\penalty0 36--47, 2018{\natexlab{a}}.

\bibitem[Su et~al.(2020)Su, Yan, Zhu, Zhang, Ge, Sun, and Zhang]{su2020blindly}
Shaolin Su, Qingsen Yan, Yu~Zhu, Cheng Zhang, Xin Ge, Jinqiu Sun, and Yanning
  Zhang.
\newblock Blindly assess image quality in the wild guided by a self-adaptive
  hyper network.
\newblock In \emph{Proceedings of the IEEE/CVF Conference on Computer Vision
  and Pattern Recognition}, pages 3667--3676, 2020.

\bibitem[Tombari et~al.(2010)Tombari, Salti, and Di~Stefano]{tombari2010unique}
Federico Tombari, Samuele Salti, and Luigi Di~Stefano.
\newblock Unique signatures of histograms for local surface description.
\newblock In \emph{Proceedings of the European Conference on Computer Vision},
  pages 356--369. Springer, 2010.

\bibitem[Johnson(1997)]{johnson1997spin}
Andrew~E Johnson.
\newblock Spin-images: A representation for 3-d surface matching.
\newblock \emph{PhD thesis, Robotics Institute, Carnegie Mellon University},
  1997.

\bibitem[Kadam et~al.(2022{\natexlab{b}})Kadam, Zhang, Liu, and
  Kuo]{kadam2022r}
Pranav Kadam, Min Zhang, Shan Liu, and C-C~Jay Kuo.
\newblock R-pointhop: A green, accurate, and unsupervised point cloud
  registration method.
\newblock \emph{IEEE Transactions on Image Processing}, 31:\penalty0
  2710--2725, 2022{\natexlab{b}}.

\bibitem[Wu et~al.(2015)Wu, Song, Khosla, Yu, Zhang, Tang, and Xiao]{wu20153d}
Zhirong Wu, Shuran Song, Aditya Khosla, Fisher Yu, Linguang Zhang, Xiaoou Tang,
  and Jianxiong Xiao.
\newblock 3d shapenets: A deep representation for volumetric shapes.
\newblock In \emph{Proceedings of the IEEE conference on computer vision and
  pattern recognition}, pages 1912--1920, 2015.

\bibitem[Besl and McKay(1992)]{besl1992method}
Paul~J Besl and Neil~D McKay.
\newblock Method for registration of 3-d shapes.
\newblock In \emph{Sensor fusion IV: control paradigms and data structures},
  volume 1611, pages 586--606. International Society for Optics and Photonics,
  1992.

\bibitem[Yang et~al.(2015)Yang, Li, Campbell, and Jia]{yang2015go}
Jiaolong Yang, Hongdong Li, Dylan Campbell, and Yunde Jia.
\newblock Go-icp: A globally optimal solution to 3d icp point-set registration.
\newblock \emph{IEEE transactions on pattern analysis and machine
  intelligence}, 38\penalty0 (11):\penalty0 2241--2254, 2015.

\bibitem[Zhou et~al.(2016)Zhou, Park, and Koltun]{zhou2016fast}
Qian-Yi Zhou, Jaesik Park, and Vladlen Koltun.
\newblock Fast global registration.
\newblock In \emph{Proceedings of the European Conference on Computer Vision},
  pages 766--782. Springer, 2016.

\bibitem[Aoki et~al.(2019)Aoki, Goforth, Srivatsan, and
  Lucey]{aoki2019pointnetlk}
Yasuhiro Aoki, Hunter Goforth, Rangaprasad~Arun Srivatsan, and Simon Lucey.
\newblock Pointnetlk: Robust \& efficient point cloud registration using
  pointnet.
\newblock In \emph{Proceedings of the IEEE conference on computer vision and
  pattern recognition}, pages 7163--7172, 2019.

\bibitem[Wang and Solomon(2019{\natexlab{a}})]{wang2019deep}
Yue Wang and Justin~M Solomon.
\newblock Deep closest point: Learning representations for point cloud
  registration.
\newblock In \emph{Proceedings of the IEEE International Conference on Computer
  Vision}, pages 3523--3532, 2019{\natexlab{a}}.

\bibitem[Wang and Solomon(2019{\natexlab{b}})]{wang2019prnet}
Yue Wang and Justin~M Solomon.
\newblock Prnet: Self-supervised learning for partial-to-partial registration.
\newblock In \emph{Advances in neural information processing systems}, pages
  8814--8826, 2019{\natexlab{b}}.

\bibitem[Kipf and Welling(2016)]{kipf2016semi}
Thomas~N Kipf and Max Welling.
\newblock Semi-supervised classification with graph convolutional networks.
\newblock \emph{International Conference on Learning Representations}, 2016.

\bibitem[Hamilton et~al.(2017)Hamilton, Ying, and
  Leskovec]{hamilton2017inductive}
Will Hamilton, Zhitao Ying, and Jure Leskovec.
\newblock Inductive representation learning on large graphs.
\newblock In \emph{Advances in neural information processing systems}, pages
  1024--1034, 2017.

\bibitem[Song et~al.(2022)Song, Yang, Xu, and King]{song2022graph}
Zixing Song, Xiangli Yang, Zenglin Xu, and Irwin King.
\newblock Graph-based semi-supervised learning: A comprehensive review.
\newblock \emph{IEEE Transactions on Neural Networks and Learning Systems},
  2022.

\bibitem[Rumelhart et~al.(1986)Rumelhart, Hinton, and
  Williams]{rumelhart1986learning}
David~E Rumelhart, Geoffrey~E Hinton, and Ronald~J Williams.
\newblock Learning representations by back-propagating errors.
\newblock \emph{nature}, 323\penalty0 (6088):\penalty0 533--536, 1986.

\bibitem[Zhu et~al.(2003)Zhu, Ghahramani, and Lafferty]{zhu2003semi}
Xiaojin Zhu, Zoubin Ghahramani, and John~D Lafferty.
\newblock Semi-supervised learning using gaussian fields and harmonic
  functions.
\newblock In \emph{Proceedings of the 20th International conference on Machine
  learning (ICML-03)}, pages 912--919, 2003.

\bibitem[Zhou et~al.(2003)Zhou, Bousquet, Lal, Weston, and
  Sch{\"o}lkopf]{zhou2003learning}
Dengyong Zhou, Olivier Bousquet, Thomas Lal, Jason Weston, and Bernhard
  Sch{\"o}lkopf.
\newblock Learning with local and global consistency.
\newblock \emph{Advances in neural information processing systems},
  16:\penalty0 321--328, 2003.

\bibitem[Li et~al.(2018{\natexlab{b}})Li, Han, and Wu]{li2018deeper}
Qimai Li, Zhichao Han, and Xiao-Ming Wu.
\newblock Deeper insights into graph convolutional networks for semi-supervised
  learning.
\newblock In \emph{Thirty-Second AAAI conference on artificial intelligence},
  2018{\natexlab{b}}.

\bibitem[Li et~al.(2019)Li, Wu, Liu, Zhang, and Guan]{li2019label}
Qimai Li, Xiao-Ming Wu, Han Liu, Xiaotong Zhang, and Zhichao Guan.
\newblock Label efficient semi-supervised learning via graph filtering.
\newblock In \emph{Proceedings of the IEEE Conference on Computer Vision and
  Pattern Recognition}, pages 9582--9591, 2019.

\bibitem[Veli{\v{c}}kovi{\'{c}} et~al.(2018)Veli{\v{c}}kovi{\'{c}}, Cucurull,
  Casanova, Romero, Li{\`{o}}, and Bengio]{velickovic2018graph}
Petar Veli{\v{c}}kovi{\'{c}}, Guillem Cucurull, Arantxa Casanova, Adriana
  Romero, Pietro Li{\`{o}}, and Yoshua Bengio.
\newblock {Graph Attention Networks}.
\newblock \emph{International Conference on Learning Representations}, 2018.

\bibitem[Abu-El-Haija et~al.(2019)Abu-El-Haija, Perozzi, Kapoor, Alipourfard,
  Lerman, Harutyunyan, Ver~Steeg, and Galstyan]{abu2019mixhop}
Sami Abu-El-Haija, Bryan Perozzi, Amol Kapoor, Nazanin Alipourfard, Kristina
  Lerman, Hrayr Harutyunyan, Greg Ver~Steeg, and Aram Galstyan.
\newblock Mixhop: Higher-order graph convolutional architectures via sparsified
  neighborhood mixing.
\newblock In \emph{international conference on machine learning}, pages 21--29.
  PMLR, 2019.

\bibitem[Zeng et~al.(2021{\natexlab{a}})Zeng, Zhang, Xia, Srivastava, Malevich,
  Kannan, Prasanna, Jin, and Chen]{zeng2021decoupling}
Hanqing Zeng, Muhan Zhang, Yinglong Xia, Ajitesh Srivastava, Andrey Malevich,
  Rajgopal Kannan, Viktor Prasanna, Long Jin, and Ren Chen.
\newblock Decoupling the depth and scope of graph neural networks.
\newblock \emph{Advances in Neural Information Processing Systems}, 34,
  2021{\natexlab{a}}.

\bibitem[Wang and Zhang(2007)]{wang2007label}
Fei Wang and Changshui Zhang.
\newblock Label propagation through linear neighborhoods.
\newblock \emph{IEEE Transactions on Knowledge and Data Engineering},
  20\penalty0 (1):\penalty0 55--67, 2007.

\bibitem[Nie et~al.(2010)Nie, Xiang, Liu, and Zhang]{nie2010general}
Feiping Nie, Shiming Xiang, Yun Liu, and Changshui Zhang.
\newblock A general graph-based semi-supervised learning with novel class
  discovery.
\newblock \emph{Neural Computing and Applications}, 19\penalty0 (4):\penalty0
  549--555, 2010.

\bibitem[Klicpera et~al.(2018)Klicpera, Bojchevski, and
  G{\"u}nnemann]{klicpera2018predict}
Johannes Klicpera, Aleksandar Bojchevski, and Stephan G{\"u}nnemann.
\newblock Predict then propagate: Graph neural networks meet personalized
  pagerank.
\newblock \emph{International Conference on Learning Representations}, 2018.

\bibitem[Huang et~al.(2020)Huang, He, Singh, Lim, and
  Benson]{huang2020combining}
Qian Huang, Horace He, Abhay Singh, Ser-Nam Lim, and Austin~R Benson.
\newblock Combining label propagation and simple models out-performs graph
  neural networks.
\newblock \emph{arXiv preprint arXiv:2010.13993}, 2020.

\bibitem[Wan et~al.(2021)Wan, Zhan, Liu, Yu, Pan, and Gong]{wan2021contrastive}
Sheng Wan, Yibing Zhan, Liu Liu, Baosheng Yu, Shirui Pan, and Chen Gong.
\newblock Contrastive graph poisson networks: Semi-supervised learning with
  extremely limited labels.
\newblock \emph{Advances in Neural Information Processing Systems}, 34, 2021.

\bibitem[Wang et~al.(2022{\natexlab{b}})Wang, Ge, Wang, and
  Kuo]{wang2022greenkgc}
Yun-Cheng Wang, Xiou Ge, Bin Wang, and C-C~Jay Kuo.
\newblock Greenkgc: A lightweight knowledge graph completion method.
\newblock \emph{arXiv preprint arXiv:2208.09137}, 2022{\natexlab{b}}.

\bibitem[Bordes et~al.(2013)Bordes, Usunier, Garcia-Duran, Weston, and
  Yakhnenko]{bordes2013translating}
Antoine Bordes, Nicolas Usunier, Alberto Garcia-Duran, Jason Weston, and Oksana
  Yakhnenko.
\newblock Translating embeddings for modeling multi-relational data.
\newblock \emph{Advances in neural information processing systems}, 26, 2013.

\bibitem[Toutanova and Chen(2015)]{toutanova2015observed}
Kristina Toutanova and Danqi Chen.
\newblock Observed versus latent features for knowledge base and text
  inference.
\newblock In \emph{Proceedings of the 3rd workshop on continuous vector space
  models and their compositionality}, pages 57--66, 2015.

\bibitem[Dettmers et~al.(2018)Dettmers, Minervini, Stenetorp, and
  Riedel]{dettmers2018convolutional}
Tim Dettmers, Pasquale Minervini, Pontus Stenetorp, and Sebastian Riedel.
\newblock Convolutional 2d knowledge graph embeddings.
\newblock In \emph{Thirty-second AAAI conference on artificial intelligence},
  2018.

\bibitem[Bollacker et~al.(2008)Bollacker, Evans, Paritosh, Sturge, and
  Taylor]{bollacker2008freebase}
Kurt Bollacker, Colin Evans, Praveen Paritosh, Tim Sturge, and Jamie Taylor.
\newblock Freebase: a collaboratively created graph database for structuring
  human knowledge.
\newblock In \emph{Proceedings of the 2008 ACM SIGMOD international conference
  on Management of data}, pages 1247--1250, 2008.

\bibitem[Miller(1995)]{miller1995wordnet}
George~A Miller.
\newblock Wordnet: a lexical database for english.
\newblock \emph{Communications of the ACM}, 38\penalty0 (11):\penalty0 39--41,
  1995.

\bibitem[Mahdisoltani et~al.(2014)Mahdisoltani, Biega, and
  Suchanek]{mahdisoltani2014yago3}
Farzaneh Mahdisoltani, Joanna Biega, and Fabian Suchanek.
\newblock Yago3: A knowledge base from multilingual wikipedias.
\newblock In \emph{7th biennial conference on innovative data systems
  research}. CIDR Conference, 2014.

\bibitem[Sun et~al.(2019)Sun, Deng, Nie, and Tang]{sun2019rotate}
Zhiqing Sun, Zhi-Hong Deng, Jian-Yun Nie, and Jian Tang.
\newblock Rotate: Knowledge graph embedding by relational rotation in complex
  space.
\newblock \emph{arXiv preprint arXiv:1902.10197}, 2019.

\bibitem[Nguyen et~al.(2017)Nguyen, Nguyen, Nguyen, and Phung]{nguyen2017novel}
Dai~Quoc Nguyen, Tu~Dinh Nguyen, Dat~Quoc Nguyen, and Dinh Phung.
\newblock A novel embedding model for knowledge base completion based on
  convolutional neural network.
\newblock \emph{arXiv preprint arXiv:1712.02121}, 2017.

\bibitem[Chami et~al.(2020)Chami, Wolf, Juan, Sala, Ravi, and
  R{\'e}]{chami2020low}
Ines Chami, Adva Wolf, Da-Cheng Juan, Frederic Sala, Sujith Ravi, and
  Christopher R{\'e}.
\newblock Low-dimensional hyperbolic knowledge graph embeddings.
\newblock \emph{arXiv preprint arXiv:2005.00545}, 2020.

\bibitem[Zhu et~al.(2022{\natexlab{b}})Zhu, Zhang, Chen, Chen, Cheng, Zhang,
  and Chen]{zhu2022dualde}
Yushan Zhu, Wen Zhang, Mingyang Chen, Hui Chen, Xu~Cheng, Wei Zhang, and Huajun
  Chen.
\newblock Dualde: Dually distilling knowledge graph embedding for faster and
  cheaper reasoning.
\newblock In \emph{Proceedings of the Fifteenth ACM International Conference on
  Web Search and Data Mining}, pages 1516--1524, 2022{\natexlab{b}}.

\bibitem[Madry et~al.(2017)Madry, Makelov, Schmidt, Tsipras, and
  Vladu]{madry2017towards}
Aleksander Madry, Aleksandar Makelov, Ludwig Schmidt, Dimitris Tsipras, and
  Adrian Vladu.
\newblock Towards deep learning models resistant to adversarial attacks.
\newblock \emph{arXiv preprint arXiv:1706.06083}, 2017.

\bibitem[Mehrabi et~al.(2021)Mehrabi, Morstatter, Saxena, Lerman, and
  Galstyan]{mehrabi2021survey}
Ninareh Mehrabi, Fred Morstatter, Nripsuta Saxena, Kristina Lerman, and Aram
  Galstyan.
\newblock A survey on bias and fairness in machine learning.
\newblock \emph{ACM Computing Surveys (CSUR)}, 54\penalty0 (6):\penalty0 1--35,
  2021.

\bibitem[Zadrozny(2004)]{zadrozny2004learning}
Bianca Zadrozny.
\newblock Learning and evaluating classifiers under sample selection bias.
\newblock In \emph{Proceedings of the twenty-first international conference on
  Machine learning}, page 114, 2004.

\bibitem[Northcutt et~al.(2021)Northcutt, Athalye, and
  Mueller]{northcutt2021pervasive}
Curtis~G Northcutt, Anish Athalye, and Jonas Mueller.
\newblock Pervasive label errors in test sets destabilize machine learning
  benchmarks.
\newblock \emph{arXiv preprint arXiv:2103.14749}, 2021.

\bibitem[Kingma and Welling(2013)]{kingma2013auto}
Diederik~P Kingma and Max Welling.
\newblock Auto-encoding variational bayes.
\newblock \emph{arXiv preprint arXiv:1312.6114}, 2013.

\bibitem[Bojanowski et~al.(2018)Bojanowski, Joulin, Lopez-Pas, and
  Szlam]{bojanowski2018optimizing}
Piotr Bojanowski, Armand Joulin, David Lopez-Pas, and Arthur Szlam.
\newblock Optimizing the latent space of generative networks.
\newblock In \emph{International Conference on Machine Learning}, pages
  600--609, 2018.

\bibitem[Li and Malik(2018)]{li2018implicit}
Ke~Li and Jitendra Malik.
\newblock Implicit maximum likelihood estimation.
\newblock \emph{arXiv preprint arXiv:1809.09087}, 2018.

\bibitem[Hoshen et~al.(2019)Hoshen, Li, and Malik]{hoshen2019non}
Yedid Hoshen, Ke~Li, and Jitendra Malik.
\newblock Non-adversarial image synthesis with generative latent nearest
  neighbors.
\newblock In \emph{Proceedings of the IEEE Conference on Computer Vision and
  Pattern Recognition}, pages 5811--5819, 2019.

\bibitem[Goodfellow et~al.(2014)Goodfellow, Pouget-Abadie, Mirza, Xu,
  Warde-Farley, Ozair, Courville, and Bengio]{goodfellow2014generative}
Ian Goodfellow, Jean Pouget-Abadie, Mehdi Mirza, Bing Xu, David Warde-Farley,
  Sherjil Ozair, Aaron Courville, and Yoshua Bengio.
\newblock Generative adversarial nets.
\newblock In \emph{Advances in neural information processing systems}, pages
  2672--2680, 2014.

\bibitem[Mao et~al.(2017)Mao, Li, Xie, Lau, Wang, and
  Paul~Smolley]{mao2017least}
Xudong Mao, Qing Li, Haoran Xie, Raymond~YK Lau, Zhen Wang, and Stephen
  Paul~Smolley.
\newblock Least squares generative adversarial networks.
\newblock In \emph{Proceedings of the IEEE international conference on computer
  vision}, pages 2794--2802, 2017.

\bibitem[Arjovsky et~al.(2017)Arjovsky, Chintala, and
  Bottou]{arjovsky2017wasserstein}
Martin Arjovsky, Soumith Chintala, and L{\'e}on Bottou.
\newblock Wasserstein generative adversarial networks.
\newblock In \emph{International Conference on Machine Learning}, pages
  214--223, 2017.

\bibitem[Azizi and Kuo(2022)]{azizi2022pager}
Zohreh Azizi and C-C~Jay Kuo.
\newblock Pager: Progressive attribute-guided extendable robust image
  generation.
\newblock \emph{arXiv preprint arXiv:2206.00162}, 2022.

\bibitem[Granot et~al.(2022)Granot, Feinstein, Shocher, Bagon, and
  Irani]{granot2022drop}
Niv Granot, Ben Feinstein, Assaf Shocher, Shai Bagon, and Michal Irani.
\newblock Drop the gan: In defense of patches nearest neighbors as single image
  generative models.
\newblock In \emph{Proceedings of the IEEE/CVF Conference on Computer Vision
  and Pattern Recognition}, pages 13460--13469, 2022.

\bibitem[Everingham et~al.(2010)Everingham, Van~Gool, Williams, Winn, and
  Zisserman]{everingham2010pascal}
Mark Everingham, Luc Van~Gool, Christopher~KI Williams, John Winn, and Andrew
  Zisserman.
\newblock The pascal visual object classes (voc) challenge.
\newblock \emph{International journal of computer vision}, 88\penalty0
  (2):\penalty0 303--338, 2010.

\bibitem[Lin et~al.(2014)Lin, Maire, Belongie, Hays, Perona, Ramanan,
  Doll{\'a}r, and Zitnick]{lin2014microsoft}
Tsung-Yi Lin, Michael Maire, Serge Belongie, James Hays, Pietro Perona, Deva
  Ramanan, Piotr Doll{\'a}r, and C~Lawrence Zitnick.
\newblock Microsoft coco: Common objects in context.
\newblock In \emph{European conference on computer vision}, pages 740--755.
  Springer, 2014.

\bibitem[Zhou et~al.(2017{\natexlab{b}})Zhou, Lapedriza, Khosla, Oliva, and
  Torralba]{zhou2017places}
Bolei Zhou, Agata Lapedriza, Aditya Khosla, Aude Oliva, and Antonio Torralba.
\newblock Places: A 10 million image database for scene recognition.
\newblock \emph{IEEE Transactions on Pattern Analysis and Machine
  Intelligence}, 2017{\natexlab{b}}.

\bibitem[Cordts et~al.(2016)Cordts, Omran, Ramos, Rehfeld, Enzweiler, Benenson,
  Franke, Roth, and Schiele]{cordts2016cityscapes}
Marius Cordts, Mohamed Omran, Sebastian Ramos, Timo Rehfeld, Markus Enzweiler,
  Rodrigo Benenson, Uwe Franke, Stefan Roth, and Bernt Schiele.
\newblock The cityscapes dataset for semantic urban scene understanding.
\newblock In \emph{Proceedings of the IEEE conference on computer vision and
  pattern recognition}, pages 3213--3223, 2016.

\bibitem[Zeng et~al.(2021{\natexlab{b}})Zeng, Liao, Tavakolian, Guo, Zhou, Hu,
  Pietik{\"a}inen, and Liu]{zeng2021deep}
Delu Zeng, Minyu Liao, Mohammad Tavakolian, Yulan Guo, Bolei Zhou, Dewen Hu,
  Matti Pietik{\"a}inen, and Li~Liu.
\newblock Deep learning for scene classification: A survey.
\newblock \emph{arXiv preprint arXiv:2101.10531}, 2021{\natexlab{b}}.

\bibitem[Zhao et~al.(2019)Zhao, Zheng, Xu, and Wu]{zhao2019object}
Zhong-Qiu Zhao, Peng Zheng, Shou-tao Xu, and Xindong Wu.
\newblock Object detection with deep learning: A review.
\newblock \emph{IEEE transactions on neural networks and learning systems},
  30\penalty0 (11):\penalty0 3212--3232, 2019.

\bibitem[Minaee et~al.(2022)Minaee, Boykov, Porikli, Plaza, Kehtarnavaz, and
  Terzopoulos]{minaee2021image}
Shervin Minaee, Yuri~Y Boykov, Fatih Porikli, Antonio~J Plaza, Nasser
  Kehtarnavaz, and Demetri Terzopoulos.
\newblock Image segmentation using deep learning: A survey.
\newblock \emph{IEEE transactions on pattern analysis and machine
  intelligence}, 44\penalty0 (7):\penalty0 3523--3542, 2022.

\bibitem[Redmon et~al.(2016)Redmon, Divvala, Girshick, and
  Farhadi]{redmon2016you}
Joseph Redmon, Santosh Divvala, Ross Girshick, and Ali Farhadi.
\newblock You only look once: Unified, real-time object detection.
\newblock In \emph{Proceedings of the IEEE conference on computer vision and
  pattern recognition}, pages 779--788, 2016.

\bibitem[Zou et~al.(2019)Zou, Shi, Guo, and Ye]{zou2019object}
Zhengxia Zou, Zhenwei Shi, Yuhong Guo, and Jieping Ye.
\newblock Object detection in 20 years: A survey.
\newblock \emph{arXiv preprint arXiv:1905.05055}, 2019.

\bibitem[Dalal and Triggs(2005)]{dalal2005histograms}
Navneet Dalal and Bill Triggs.
\newblock Histograms of oriented gradients for human detection.
\newblock In \emph{2005 IEEE computer society conference on computer vision and
  pattern recognition (CVPR'05)}, volume~1, pages 886--893. Ieee, 2005.

\bibitem[Felzenszwalb et~al.(2008)Felzenszwalb, McAllester, and
  Ramanan]{felzenszwalb2008discriminatively}
Pedro Felzenszwalb, David McAllester, and Deva Ramanan.
\newblock A discriminatively trained, multiscale, deformable part model.
\newblock In \emph{2008 IEEE conference on computer vision and pattern
  recognition}, pages 1--8. Ieee, 2008.

\bibitem[Wei et~al.(2022{\natexlab{a}})Wei, Kuo, Testa, Machado-Lima, and
  Fatima]{wei2022expressionhop}
Chengwei Wei, C.-C.~Jay Kuo, Rafael~Liuz Testa, Ariane Machado-Lima, and
  L.~S.~Nunes Fatima.
\newblock Expression{H}op: A lightweight human facial expression classifier.
\newblock In \emph{Proceedings of the IEEE Conference on Multimedia Information
  Processing and Retrieval}, 2022{\natexlab{a}}.

\bibitem[Wei et~al.(2022{\natexlab{b}})Wei, Wang, and Kuo]{wei2022synwmd}
Chengwei Wei, Bin Wang, and C-C~Jay Kuo.
\newblock Synwmd: Syntax-aware word mover's distance for sentence similarity
  evaluation.
\newblock \emph{arXiv preprint arXiv:2206.10029}, 2022{\natexlab{b}}.

\bibitem[Wei et~al.(2022{\natexlab{c}})Wei, Wang, and Kuo]{wei2022task}
Chengwei Wei, Bin Wang, and C-C~Jay Kuo.
\newblock Task-specific dependency-based word embedding methods.
\newblock \emph{Pattern Recognition Letters}, 2022{\natexlab{c}}.

\bibitem[Ge et~al.(2022)Ge, Wang, Wang, and Kuo]{ge2022compounde}
Xiou Ge, Yun-Cheng Wang, Bin Wang, and C-C~Jay Kuo.
\newblock Compounde: Knowledge graph embedding with translation, rotation and
  scaling compound operations.
\newblock \emph{arXiv preprint arXiv:2207.05324}, 2022.

\bibitem[Wang et~al.(2022{\natexlab{c}})Wang, Ge, Wang, and
  Kuo]{wang2022kgboost}
Yun-Cheng Wang, Xiou Ge, Bin Wang, and C-C~Jay Kuo.
\newblock Kgboost: A classification-based knowledge base completion method with
  negative sampling.
\newblock \emph{Pattern Recognition Letters}, 157:\penalty0 104--111,
  2022{\natexlab{c}}.

\bibitem[Yang et~al.(2022{\natexlab{c}})Yang, Magoulianitis, Wang, and
  Kuo]{yang2022statistical}
Yijing Yang, Vasileios Magoulianitis, Xinyu Wang, and C-C~Jay Kuo.
\newblock Statistical attention localization (sal): Methodology and application
  to object classification.
\newblock \emph{arXiv preprint arXiv:2208.01823}, 2022{\natexlab{c}}.

\bibitem[Alvarez and Salzmann(2017)]{alvarez2017compression}
Jose~M Alvarez and Mathieu Salzmann.
\newblock Compression-aware training of deep networks.
\newblock \emph{Advances in neural information processing systems}, 30, 2017.

\bibitem[Cheng et~al.(2017)Cheng, Wang, Zhou, and Zhang]{cheng2017survey}
Yu~Cheng, Duo Wang, Pan Zhou, and Tao Zhang.
\newblock A survey of model compression and acceleration for deep neural
  networks.
\newblock \emph{arXiv preprint arXiv:1710.09282}, 2017.

\bibitem[Choudhary et~al.(2020)Choudhary, Mishra, Goswami, and
  Sarangapani]{choudhary2020comprehensive}
Tejalal Choudhary, Vipul Mishra, Anurag Goswami, and Jagannathan Sarangapani.
\newblock A comprehensive survey on model compression and acceleration.
\newblock \emph{Artificial Intelligence Review}, 53\penalty0 (7):\penalty0
  5113--5155, 2020.

\bibitem[Hoefler et~al.(2021)Hoefler, Alistarh, Ben-Nun, Dryden, and
  Peste]{hoefler2021sparsity}
Torsten Hoefler, Dan Alistarh, Tal Ben-Nun, Nikoli Dryden, and Alexandra Peste.
\newblock Sparsity in deep learning: Pruning and growth for efficient inference
  and training in neural networks.
\newblock \emph{J. Mach. Learn. Res.}, 22\penalty0 (241):\penalty0 1--124,
  2021.

\bibitem[Louizos et~al.(2017)Louizos, Ullrich, and
  Welling]{louizos2017bayesian}
Christos Louizos, Karen Ullrich, and Max Welling.
\newblock Bayesian compression for deep learning.
\newblock \emph{Advances in neural information processing systems}, 30, 2017.

\bibitem[Murshed et~al.(2021)Murshed, Murphy, Hou, Khan, Ananthanarayanan, and
  Hussain]{murshed2021machine}
MG~Sarwar Murshed, Christopher Murphy, Daqing Hou, Nazar Khan, Ganesh
  Ananthanarayanan, and Faraz Hussain.
\newblock Machine learning at the network edge: A survey.
\newblock \emph{ACM Computing Surveys (CSUR)}, 54\penalty0 (8):\penalty0 1--37,
  2021.

\bibitem[Howard et~al.(2017)Howard, Zhu, Chen, Kalenichenko, Wang, Weyand,
  Andreetto, and Adam]{howard2017mobilenets}
Andrew~G Howard, Menglong Zhu, Bo~Chen, Dmitry Kalenichenko, Weijun Wang,
  Tobias Weyand, Marco Andreetto, and Hartwig Adam.
\newblock Mobilenets: Efficient convolutional neural networks for mobile vision
  applications.
\newblock \emph{arXiv preprint arXiv:1704.04861}, 2017.

\bibitem[Zhang et~al.(2018{\natexlab{b}})Zhang, Zhou, Lin, and
  Sun]{zhang2018shufflenet}
Xiangyu Zhang, Xinyu Zhou, Mengxiao Lin, and Jian Sun.
\newblock Shufflenet: An extremely efficient convolutional neural network for
  mobile devices.
\newblock In \emph{Proceedings of the IEEE conference on computer vision and
  pattern recognition}, pages 6848--6856, 2018{\natexlab{b}}.

\bibitem[Wu et~al.(2018)Wu, He, Sun, and Tan]{wu2018light}
Xiang Wu, Ran He, Zhenan Sun, and Tieniu Tan.
\newblock A light cnn for deep face representation with noisy labels.
\newblock \emph{IEEE Transactions on Information Forensics and Security},
  13\penalty0 (11):\penalty0 2884--2896, 2018.

\end{thebibliography}

\end{document}